\documentclass[mnsc,nonblindrev]{informs3} 
\OneAndAHalfSpacedXI 

\usepackage{endnotes}
\usepackage[ruled]{algorithm}
\usepackage{algorithmicx,algpseudocode,amsfonts,subfigure}
\usepackage{graphicx,hyperref,comment,appendix}
\usepackage[capitalize,noabbrev]{cleveref}
\usepackage{tikz}
\usepackage{subfigure,caption,microtype,booktabs} 

\usepackage{amsthm,amsmath,amssymb,soul}
\usepackage{comment,enumitem}
\setlist{leftmargin=*}

\newtheorem{condition}{Condition}
\newtheorem{theorem}{Theorem}
\newtheorem{corollary}{Corollary}
\newtheorem{lemma}{Lemma}
\newtheorem{proposition}{Proposition}
\newtheorem{remark}{Remark}

\newcommand{\ie}{\textit{i.e.}}

\newcommand{\satexltp}{\texttt{SELECT-LITE+}}
\newcommand{\eg}{\textit{e.g.}}
\newcommand{\E}{\mathbb{E}}
\renewcommand{\Pr}{\text{Pr}}
\newcommand{\regret}{\texttt{Regret}}

\newcommand{\satex}{\texttt{SELECT}}
\newcommand{\satexlt}{\texttt{SELECT-LITE}}

\usepackage{natbib}
\bibpunct[, ]{(}{)}{,}{a}{}{,}%
\def\bibfont{\small}%
\usepackage{bm,comment}
\usepackage{soul}

\ECRepeatTheorems

\EquationsNumberedThrough    

\allowdisplaybreaks
\begin{document}


\RUNAUTHOR{}

\RUNTITLE{Satisficing Regret Minimization in Bandits}



\TITLE{Satisficing Regret Minimization in Bandits: Constant Bound and Light-Tailed Distribution}

\ARTICLEAUTHORS{

\AUTHOR{Qing Feng}
\AFF{School of Operations Research and Information Engineering, Cornell University\\
	\EMAIL{qf48@cornell.edu}}

\AUTHOR{Tianyi Ma}
\AFF{School of Operations Research and Information Engineering, Cornell University\\
	 \EMAIL{tm693@cornell.edu}} 

\AUTHOR{Ruihao Zhu}
\AFF{SC Johnson College of Business, Cornell University\\
	\EMAIL{ruihao.zhu@cornell.edu}} 

} 

\ABSTRACT{%
Motivated by the concept of \emph{satisficing} in decision-making, we consider the problem of satisficing regret minimization in bandit optimization. In this setting, the learner aims at selecting satisficing arms (arms with mean reward exceeding a certain threshold value) as frequently as possible. The performance is measured by \emph{satisficing regret}, which is the cumulative deficit of the chosen arm's mean reward compared to the threshold. We propose \satex, a general algorithmic template for \underline{S}atisficing R\underline{E}gret Minimization via Samp\underline{L}ing and Low\underline{E}r \underline{C}onfidence bound \underline{T}esting, that attains constant expected satisficing regret for a wide variety of bandit optimization problems in the \emph{realizable} case (\ie, a satisficing arm exists). Specifically, given a class of bandit optimization problems and a corresponding learning oracle with sub-linear expected (standard) regret upper bound, \satex~iteratively makes use of the oracle to identify a potential satisficing arm with low regret. Then, it collects data samples from this arm, and continuously compares the LCB of the identified arm's mean reward against the threshold value to determine if it is a satisficing arm. As a complement, \satex~also enjoys the same (standard) regret guarantee as the oracle in the non-realizable case. {To further ensure stability of the algorithm, we introduce~\satexlt~that achieves a light-tailed satisficing regret distribution plus a constant expected satisficing regret in the realizable case and a sub-linear expected (standard) regret in the non-realizable case. Notably, \satexlt~can operate on learning oracles with heavy-tailed (standard) regret distribution. More importantly, our results reveal the surprising compatibility between constant expected satisficing regret and light-tailed satisficing regret distribution, which is in sharp contrast to the case of (standard) regret.} Finally, we conduct numerical experiments to validate the performance of \satex~and \satexlt~on both synthetic datasets and a real-world dynamic pricing case study.
}%



\maketitle

%

\section{Introduction}\label{sec:intro}

Multi-armed bandit (MAB) is a classic framework for decision-making under uncertainty. In MAB, a learner is given a set of arms with initially unknown mean rewards. She repeatedly picks an arm to collect its noisy reward and her goal is to maximize the cumulative rewards. The performance is usually measured by the notion of \emph{regret}, which is the difference between the maximum possible total mean rewards and the total mean rewards collected by the learner. With its elegant form, MAB has found numerous applications in recommendation systems, ads display, and beyond, and many (near-)optimal algorithms have been developed for various classes of bandit optimization problems \citep{LS20}. 

While making the optimal decision is desired in certain high-stakes situations, it is perhaps not surprising that what we often need is just a good-enough or \emph{satisficing} alternative \citep{Lufkin21}. Coined by Nobel laureate Herbert Simon \citep{Simon56}, the term ``satisficing" refers to the decision-making strategy that settles for an acceptable solution rather than exhaustively hunting for the best possible. 

The concept of satisficing finds its place in many real-world decision-making under uncertainty settings \citep{caplin2011satisficing} as shown below. 
\begin{itemize}
\item { \textbf{Cloud Service:} Consider an operator tuning the configuration of a user-facing cloud service. In each period, the operator selects a configuration, such as a resource allocation, caching rule, batching level, routing policy, or timeout setting, and deploys it for a short evaluation window. Then, the operator observes a noisy measure of service quality, \eg, the fraction of requests that are completed correctly within a prescribed latency bound. The operator uses this feedback to choose the configuration in the next period. In this setting, the operator typically does not need to identify the configuration with the highest possible service quality. Instead, the goal is to meet a prespecified service-level objective, such as ensuring that at least 99\% of requests satisfy the latency requirement (see, \eg, Chapter 4 of \citealt{Sloss03} and \citealt{Google26}). Configurations that reliably meet this target are acceptable, whereas configurations that fall short expose users to degraded service and should be avoided. Therefore, in this setting, the objective is to hit the configurations that meet the reliability requirements as frequently as possible rather than searching for the configuration with the maximum service level.}  
\item \textbf{Inventory Management:} Studies show that product stockouts can result in customer disengagement and impair a firm's profitability \citep{fitzsimons2000stockout,jing2011stockout}. As such, retailers usually replenish their products periodically to improve service level (\ie, probability of no stockout for any product). { Notably, due to the complexity of inventory decisions as well as various constraints (\eg, capacity and inventory cost), it is often very challenging and costly to identify the optimal inventory decision that maximizes service level. Therefore, instead of blindly maximizing service level, retailers typically set a target service level, \eg, the ideal service level in retailing is $90\%\sim 95\%$ \citep{levy2023retailing}.} Consequently, in the face of the unknown demand distributions, one can cast the problem into a MAB setting with a \emph{different} objective. In this setting, a retailer starts without complete knowledge of the demand distributions and proceeds by trying various stock level combinations (\ie, arms) of the products over time. Here, the objective is to hit the target service level as frequently as possible rather than maximizing the service level every time.
\item \textbf{Product Design:} In fashion product design, designers usually need to design a sequence of products for a group of customers with initially unknown preferences. { Since the design of a product is often a highly complex decision involving deciding a combination of many features of a product, identifying the optimal design often involves extensive exploration over many feature combinations, which could be both costly and time-consuming for designers. Therefore, instead of demanding the best design every time, designers only need to pick a design that can hopefully meet the expectation of a certain portion of the customers for most of the time \citep{Whitenton14,Mabey22}.}
\end{itemize}  
The above examples give rise to the study of satisficing bandits \citep{TamatsukuriT19,Michel23} where the objective is to hit a satisficing arm (\ie, an arm whose mean reward exceeds a certain threshold value) as frequently as possible. The corresponding performance measure is \emph{satisficing regret}, which is the cumulative deficit of the chosen arm's mean reward compared to the threshold. In particular, \cite{Michel23}~develops SAT-UCB\footnote{ Algorithm 2, Section 3.1 in \cite{Michel23}}, an algorithm for finite-armed satisficing bandits: In the \emph{realizable case} (\ie, a satisficing arm exists), the algorithm attains a constant expected satisficing regret, \ie, independent of the length of the time horizon; otherwise, in the non-realizable case (\ie, the threshold value is out of reach), it preserves the optimal logarithmic expected (standard) regret.

Despite these, existing algorithms usually impose that there is a clear separation between the threshold value and the largest mean reward among all \emph{non-satisficing arms}, which is referred to as the \emph{satisficing gap}. Moreover, the expected satisficing regret bounds usually scale \emph{inverse proportionally} to the satisficing gap. On the other hand, it is evident that in many bandit optimization problem classes (\eg, Lipschitz bandits and concave bandits), the satisficing gap can simply be 0, making the bounds vacuous. Therefore, it is not immediately clear if SAT-UCB \citep{Michel23} or other prior results can go beyond finite-armed bandits and if a constant expected satisficing regret bound is possible for general classes of bandit optimization problems. 

\subsection{Main Contributions}

In this work, we propose \satex, a novel algorithmic template for \underline{S}atisficing R\underline{E}gret Minimization via Samp\underline{L}ing and Low\underline{E}r \underline{C}onfidence Bound \underline{T}esting. More specifically:
\begin{itemize}[leftmargin=*]
\item We describe the design of \satex~in Section \ref{sec:alg}. For a given bandit optimization problem class and an associated learning oracle with sub-linear (but not necessarily optimal) expected standard regret guarantee, \satex~runs in rounds with the help of the oracle. At the beginning of a round, \satex~first applies the oracle for a number of time steps and randomly samples an arm from its trajectory as a candidate satisficing arm. Then, it conducts forced sampling by pulling this arm for a pre-specified number of times. Finally, it continues to pull this arm and starts to monitor the resulting lower confidence bound (LCB) of its mean reward. \satex~terminates the current round and starts the next once the LCB falls below the threshold value;

\item In Section \ref{sec:reg_analysis}, we establish that in the realizable case, \ie, when a satisficing arm exists, \satex~is able to achieve a constant expected satisficing regret. Notably, the expected satisficing regret bound has no dependence on the satisficing gap (see the forthcoming Remark \ref{remark:lcb} for a detailed discussion). Instead, it scales inverse proportionally to the \emph{exceeding gap}, which measures the difference between the optimal reward and the threshold value, and is positive in general; otherwise (\ie, when no satisficing arm exists), \satex~enjoys the same expected standard regret bound as the learning oracle. { In \cref{sec:best_of_both_world} of the appendix, we further propose \texttt{SELECT-BoBW}, an adapted version of \satex~in finite-arm bandits that attains a ``best of both worlds" expected satisficing regret bound, which dominates both the expected satisficing regret bound of \satex~w.r.t. the exceeding gap, and the bound of existing algorithms w.r.t. on the satisficing gap.}

\item In Section \ref{sec:examples}, we instantiate \satex~to finite-armed bandits, concave bandits and Lipschitz bandits, and demonstrate the corresponding expected satisficing and standard regret bounds. { We also establish lower bounds of the expected satisficing regret for finite-arm bandits, concave bandits and Lipschitz bandits, showing that the expected satisficing regret bounds established by \satex~are near-optimal for all three bandit problem classes mentioned above};

\item While \satex~attains constant satisficing regret in expectation, its performance in individual realizations may not be stable, \ie, the probability of incurring a large satisficing regret is not sufficiently small. To address this, in Section \ref{sec:satisficing_tail}, we consider the tail risk control in satisficing regret minimization. To confirm the necessity of new algorithms for light-tailed satisficing regret distribution\footnote{ The formal definition of light-tailed and heavy-tailed distributions are provided at the beginning of \cref{sec:satisficing_tail}.}, we first prove that the satisficing regret distribution of \satex~can be heavy-tailed even if we use a learning oracle with light-tailed standard regret distribution. Then, we introduce~\satexlt, a modified version of \satex~to obtain a light-tailed satisficing regret distribution. Specifically, with an arbitrary learning oracle with sub-linear expected standard regret bound, \satexlt~enjoys the following desirable properties: 
\begin{itemize}
    \item In the realizable case, \satexlt~attains a constant expected satisficing regret bound, and its satisficing regret distribution is light-tailed; and
    \item In the non-realizable case, \satexlt~still maintains a sub-linear in $T$ expected standard regret bound.
\end{itemize} 
Notably, our results do not rely on any assumptions on the tail distribution of the learning oracle's standard regret. In fact, \satexlt~is able to make use of a learning oracle with heavy-tailed standard regret to obtain a light-tailed satisficing regret distribution in the realizable case. Moreover, in \cite{simchi2024simple}, it is shown that any algorithm with a light-tailed standard regret distribution has to suffer a $\Omega(\text{poly}(T))$ instance-dependent expected standard regret\footnote{ Instance-dependent regret means it is analyzed for a fixed problem instance, \eg, with instance-specific parameters such as $\Delta = \min_{A\neq A^*} r(A^*) - r(A)$, and only its dependence on the time horizon is considered.}, rather than a near-optimal $O(\log(T))$ instance-dependent expected standard regret. Because the constant expected satisficing regret bounds are typically instance-dependent, it becomes unclear if a light-tailed satisficing regret distribution will inevitably cause any dependence on $T$ in the expected satisficing regret. Nevertheless, our results show that achieving a light-tailed satisficing regret distribution only worsens the expected satisficing regret by a constant. This thus reveals the compatibility between constant expected satisficing regret and light-tailed satisficing regret distribution. It also indicates a fundamental difference between the notions of standard regret and satisficing regret.

\item In Section \ref{sec:numerical}, we conduct numerical experiments to demonstrate the performance of \satex~on synthetic instances of finite-armed bandits, concave bandits, and Lipschitz bandits. We use SAT-UCB, SAT-UCB+ (a heuristic with no regret guarantee proposed by \citealt{Michel23}), and the respective learning oracle as the benchmarks. Our results reveal that in the realizable case, \satex~does achieve constant expected satisficing regret. Moreover, both its expected satisficing regret (in the realizable cases) and standard regret (in the non-realizable cases) either significantly outperform the benchmarks' or are close to the best-performing ones'. We also test the performance of \satexlt~and compare it with \satex. We show that \satexlt~has better performance in terms of tail distribution control compared to \satex~while still maintaining a constant expected satisficing regret comparable to \satex. This indicates that \satexlt~can provide better protection against tail risks without incurring a significantly larger satisficing regret in expectation. { In \cref{sec:experiment_vary_S} of the appendix, we further conduct experiments on finite-arm bandits with varying pairs of satisficing gaps and exceeding gaps, where we compare the performance of \texttt{SELECT-BoBW} with Thompson Samping and existing algorithms on satisficing in finite-arm bandits. The results of the experiment show a robust empirical performance of \texttt{SELECT-BoBW}, as the algorithm outperforms other benchmark algorithms under most pairs of satisficing gaps and exceeding gaps.}

\item To test the performance of our algorithm in real-world instances, in Section~\ref{sec:numerical_real_data}, we also conduct numerical experiments using a case study of dynamic pricing calibrated from a real-world dataset. Using a real-world avocado sales dataset \cite{kiggins2018avocado}, we show that \satex~attains a constant expected satisficing regret in the realizable case. Furthermore, our results reveal that \satex~outperforms all other benchmarks both in the realizable and non-realizable cases.

\item A preliminary version of this paper \citep{anonymous2025satisficing} has appeared in the Proceedings of the 13th International Conference on Learning Representations (ICLR 2025). The current paper additionally provides the following significant contributions. 
\begin{enumerate}
    \item { We have strengthened our existing lower bound results in \cref{sec:lower_bounds}. Specifically, we have proven a $\Omega(L^d/(\Delta_S^*)^{d+1})$ lower bound on the expected satisficing regret for Lipschitz bandits (\cref{thm:lb_lipschitz}). We have also improved our existing lower bounds of $\Omega(1/\Delta_S^*)$ for finite-arm bandits and obtained a $\Omega(K/\Delta_S^*)$ lower bound for finite-arm bandits (\cref{thm:lb_two_arm}). }
    \item In \cref{sec:satisficing_tail}, we go beyond minimizing satisficing regret only in expectation and consider the tail risk control in satisficing regret minimization. Specifically, we build on \satex~to establish \satexlt, which is able to attain a light-tailed satisficing regret distribution (\cref{thm:tail_risk}) while still maintaining a constant expected satisficing regret in the realizable case (\cref{thm:regret_bound_tail}) for a wide range of bandit problems;
    \item In \cref{sec:numerical_tail}, we test the empirical performance of \satexlt~using an instance of finite-armed bandit. The results show that \satexlt~is able to maintain a smaller tail probability in satisficing regret compared to \satex~while the expected satisficing regret of \satexlt~is comparable to that of \satex;
    \item In \cref{sec:numerical_real_data}, we test our algorithm using a case study of dynamic pricing calibrated from a real-world avocado sales dataset. We observe that under this instance \satex~is able to significantly outperform all other benchmarks, demonstrating a strong empirical performance of our algorithm in practical bandit problems.
    \item { In \cref{sec:best_of_both_world} in the appendix, we propose \texttt{SELECT-BoBW}, an adapted version of \satex~in finite-arm bandits that attains a ``best of both worlds" expected satisficing regret bound, which dominates both the expected satisficing regret bound of \satex~w.r.t. the exceeding gap, and the bound of existing algorithms w.r.t. on the satisficing gap. We further conduct numerical experiments in \cref{sec:experiment_vary_S}, we further use numerical experiments to show that \texttt{SELECT-BoBW} achieves good and robust empirical performance and consistently outperforms other benchmark algorithms in a wide range of satisficing gap and exceeding gap pairs.}
\end{enumerate} 
\end{itemize}

\subsection{Related Works}

Besides \cite{Michel23}, the most relevant works to ours are \cite{BubeckPR13, GarivierM19, TamatsukuriT19}, all of which focus on finite-armed bandits. \cite{BubeckPR13} and \cite{GarivierM19} introduce constant regret algorithms with knowledge of the optimal reward. This can be viewed as a special case of satisficing where the threshold value is exactly the optimal mean reward. \cite{HuyukT21} further extend the algorithm in \cite{GarivierM19} to multi-objective multi-armed bandit settings. \cite{TamatsukuriT19}~also establishes constant regret for finite-armed satisficing bandits when there exists exactly one satisficing arm. As a follow-up paper of~\cite{Michel23}, \cite{HajiabolhassanO23} considers satisficing in finite-state MDP.

Among others, \cite{RussoV22}~is one of the first to introduce the notion of satisficing regret, but they focus on a time-discounted setting. Furthermore, in \cite{RussoV22}, the learner is assumed to know the difference between the values of the threshold and the optimal reward.

More broadly, the concept of satisficing has also been studied by \cite{ReverdyL14,ReverdySL16,AbernethyAZ16} in bandit learning. Both \cite{ReverdyL14} and \cite{AbernethyAZ16} focus on maximizing the number of times that the selected arm's \emph{realized reward} exceeds a threshold value. This is equivalent to a MAB problem that sets each arm's mean reward as its probability of generating a value that exceeds the threshold. In view of this, the problems investigated in \cite{ReverdyL14} and \cite{AbernethyAZ16} are closer to conventional MABs. They are also different from \cite{ReverdySL16,Michel23} and ours, which use the mean reward of the selected arms to compute the satisficing regret. We note that in \cite{ReverdySL16}, a more general Bayesian setting is studied. In this setting, the satisficing regret also takes the learner’s belief that some arm is satisficing into account.

The notion of satisficing is also studied in the pure exploration setting \citep{LocatelliGC16, MukherjeeNSR17, KanoHSM19} as \emph{thresholding bandits}. Their objective is to identify \emph{all} arms with mean reward above a certain satisficing level with limited exploration budget. Unlike the satisficing regret we consider, the performance of their algorithm is measured by simple regret, \ie, the expected number of misidentification made by the final output. Note that any pure exploration algorithm \citep{AudibertB10} is able to identify the optimal arm or a satisficing arm with high probability after a number of steps. However, due to a small but positive error probability, subsequent exploitation of the identified arm will always incur linear satisficing regret, thus a simple explore-then-exploit approach is unable to obtain a constant satisficing regret bound. Similar observations have also been made in \cite{Michel23}.

Another line of research relevant to our work is tail risk control in bandits. The general goal of this line of research is to design bandit algorithms that achieve light-tailed regret distributions besides minimizing expected regret. Some early works studying the tail distribution of bandit algorithms include \cite{audibert2009exploration} and \cite{salomon2011deviations}, who show that for UCB algorithms, the probability of regret greater than $c(\log(T))^p$ (where $p>1$ is fixed) decays only polynomially in $T$. \cite{fan2024fragility} are among the first to establish bandit algorithms with a light-tailed (standard) regret distribution. They focus on the finite-armed bandit setting and show that for informationally-optimized bandit algorithms such as the UCB algorithm and Thompson sampling, the probability of incurring a linear in $T$ regret is at least $\Omega(1/T)$. Some follow-up works include \cite{simchi2023stochastic} and \cite{simchi2024simple}. They show that any finite-armed bandit algorithm with an $O(\log(T))$ instance-dependent expected regret bound has a heavy-tailed regret distribution. They also establish an optimal tradeoff between tail risk, worst-case expected regret bound and instance-dependent expected regret bound. In our paper, we consider tail risk control for satisficing regret minimization. We show that unlike tail risk control in standard regret minimization, where $O(\log(T))$ instance-dependent expected regret bound cannot coexist with light-tailed regret distribution, in satisficing regret minimization there exists an algorithm that attains a constant expected satisficing regret and a light-tailed satisficing regret distribution.

\section{Problem Formulation}
In this section, we present the setup of our problem. Consider a class of bandit optimization problems $(\mathcal{A},\mathcal{R})$, where $\mathcal{A}$ is the arm set and $\mathcal{R}\subseteq\{r:\mathcal{A}\to[0,1]\}$ is the class of admissible reward functions. The learner is given a set of feasible arms $\mathcal{A}$ and a class of admissible reward functions $\mathcal{R}$, pulling arm $A\in\mathcal{A}$ in a time step brings a random reward with mean $r(A)$. The learner knows the underlying reward function $r(\cdot)$ belongs to the function class $\mathcal{R}$, but the exact $r(\cdot)$ is initially unknown to the learner. In each time step $t=1,2,\ldots,T$, the learner pulls an arm $A_t\in\mathcal{A}$, and then observes a noisy reward $Y_t$ { supported on $[0,1]$ with mean $\mathbb{E}[Y_t|A_1,Y_1\ldots,A_{t-1},Y_{t-1},A_t]=r(A_t)$.}

\begin{remark}
The notion of problem class $(\mathcal{A},\mathcal{R})$ captures most of the bandit optimization problems in the stationary setting. For example, if $\mathcal{A}=[K]$ and $\mathcal{R}=\{r:[K]\to[0,1]\}$, then the problem class $(\mathcal{A},\mathcal{R})$ corresponds to finite-armed bandits with $K$ arms; if $\mathcal{A}$ is any convex set in $\mathbb{R}^d$ and \mbox{$\mathcal{R}=\{r:\mathcal{A}\to[0,1],\,r\ \text{is concave and 1-Lipschitz}\}$}, then the problem class $(\mathcal{A},\mathcal{R})$ corresponds to concave bandits.
\end{remark}

In this work, we define two notions of regret. First, with the presence of satisficing level $S$, we consider the satisficing regret introduced in \cite{ReverdySL16,Michel23}, which is the cumulative deficit of the chosen arm's mean reward when compared to $S$, \ie
\begin{equation*}
\regret_S=\sum_{t=1}^T\max\{S-r(A_t),0\}.
\end{equation*}
Because it is possible that no arm achieves the satisficing level, we also follow the conventional bandit literature to define the (standard) regret, which measures the cumulative difference between the chosen arm's mean reward and the optimal mean reward. Specifically, let $A^*=\arg\max_{A\in\mathcal{A}}r(A),$ the standard regret is defined as
\begin{equation*}
\regret=T\cdot r(A^*)-\sum_{t=1}^T r(A_t).
\end{equation*}
We say that the setting is \emph{realizable} if $r(A^*)\geq S$; otherwise, it is \emph{non-realizable}. We recall that an arm $A$ is satisficing if $r(A)\geq S$; otherwise, it is non-satisficing.

\begin{remark} In the definition of satisficing regret, we consider the cumulative deficit of mean reward compared to the satisficing threshold because ideally we would like the arm pull in every time step to be satisficing, thus we penalize every arm pull of non-satisficing arms. Even though in some time steps the decision maker may pull an arm with mean reward over the satisficing threshold, the excessive mean reward cannot compensate for the satisficing regret incurred in other time steps. With this definition of satisficing regret, the problem cannot be trivially solved by running a traditional bandit algorithm. Although any bandit algorithm will gradually converge to the optimal arm and achieve a satisficing total reward with long enough time horizon, as the time horizon $T$ increases, the number of arm pulls of non-satisficing arms will also increase, incurring a satisficing regret scaling in $T$. In the forthcoming \cref{sec:numerical} and \cref{sec:numerical_real_data}, we also numerically show that directly running bandit algorithms can incur substantial satisficing regret.
\end{remark}

Existing works (\eg, \citealt{Michel23}) are mostly focused on satisficing regret minimization in finite-armed bandit settings, and usually derive expected satisficing regret bound that scales inversely proportional to the \emph{satisficing gap} 
$$\Delta_S=\min\{S-r(A):r(A)<S\}.$$ However, for bandit optimization problems with large and even infinitely many arms, $\Delta_S$ can approach 0 quickly. To address this issue, we define the notion of \emph{exceeding gap} that captures the difference between the optimal reward and $S$ { in the realizable case}, \ie,
\begin{equation*}
    \Delta_S^*=r(A^*)-S.
\end{equation*}

{

\begin{remark}[{\bf Shortfalls of Existing Algorithms}]

In what follows, we discuss how existing algorithms such as the ones in \cite{GarivierM19} and \cite{Michel23} are unable to extend to more general bandit settings. At a high level, both the algorithm in \cite{GarivierM19} and the one in \cite{Michel23} use the following steps: 
\begin{itemize}
    \item \textbf{Step 1:} Exploit a large candidate arm set that contains \textit{all} arms that are potentially satisficing;
    \item \textbf{Step 2:} While exploiting the candidate arm set, remove any arm that no longer meets the criteria of a candidate satisficing arm;
    \item \textbf{Step 3:} Whenever the candidate arm set becomes empty, uniformly explore the entire arm set until the candidate arm set becomes non-empty again. 
\end{itemize}
Here, in \cite{GarivierM19}, the criterion for a candidate satisficing arm is that its UCB should exceed $S$, while in \cite{Michel23}, the criterion is that its empirical mean reward should exceed $S$.

While the aforementioned idea works well for finite-arm bandits, it can easily run into problems in many other bandit problem classes, particularly those with continuous arm sets (\eg, concave bandits and Lipschitz bandits). Notably, continuous arm sets present at least two challenges:

\begin{itemize}

\item \underline{\it Costly Exploration and Arm Removal under Large Number of Arms:} With a continuous arm set, the number of arms becomes infinite. Even after applying some discretization, the algorithm still needs to explore many more arms compared to finite-arm bandits when doing uniform exploration (\ie, Step 3), and many of those arms could be non-satisficing. For this reason, uniform exploration  could mean a large number of arm pulls among non-satisficing arms, further incurring significant satisficing regret. Similarly, removing non-satisficing arms (\ie, Step 2) could also be costly, as the candidate satisficing arm set could potentially contain a large number of non-satisficing arms and significant satisficing regret may be incurred when removing these non-satisficing arms one by one.

\item \underline{\it No Separation between Non-Satisficing Arms' Mean Rewards and Satisficing Threshold:} The issue is further worsened by the lack of separation between non-satisficing arms' mean rewards and the satisficing threshold. Specifically, with a continuous arm set, mean rewards of non-satisficing arms can be arbitrarily close to the satisficing threshold $S$. This brings significant challenges in removal of non-satisficing arms. To see this more concretely, consider a non-satisficing arm whose mean reward is $S-\delta$. Under the criteria for candidate satisficing arms in either \cite{GarivierM19} or \cite{Michel23}, removal of this arm is guaranteed only after $\tilde{\Omega}(1/\delta^2)$ pulls of the arm, and an expected satisficing regret of $\tilde{\Omega}(1/\delta)$ is incurred during the removal process. With the arm set being continuous, $\delta$ could be arbitrarily small (even after discretization), making the expected satisficing regret incurred when removing even one single non-satisficing arm arbitrarily large. We would also like to remark that this issue cannot be solved by simply using a stricter criterion for candidate satisficing arms, \eg, requiring that the arm's LCB should exceed $S$. With a stricter critera for candidate satisficing arms, the candidate arm set can easily become empty, which forces the algorithm to run the already costly uniform exploration over and over again, thus incurring even larger satisficing regret.

\end{itemize}

For these reasons, existing algorithms in \cite{GarivierM19} and \cite{Michel23} are not applicable to satisficing regret minimization in bandit classes beyond finite-arm bandits, and addressing these challenges requires significant algorithmic departure from existing approaches. We have also added the discussions above at the end of Section 2 of our revised paper (see Remark 3 of the revised paper for details).

\end{remark}

}

\section{\satex: Satisficing Regret Minimization via Sampling and Lower Confidence Bound Testing}\label{sec:alg}

In this section, we present our algorithmic template \satex~for general classes of bandit optimization problems. When the arm set $\mathcal{A}$ is not too large, one can collect enough data samples for every arm to estimate its mean reward and gradually abandon any non-satisficing arms (see, \eg, \citealt{GarivierM19,Michel23}). In general, however, the arm set can be large and may even contain infinitely many arms (\eg, continuous arm set). Therefore, it becomes impossible to identify all non-satisficing arms. We thus take an alternative approach: The algorithm repeatedly locates a potentially satisficing arm with low (satisficing) regret. Then, it tests if this candidate arm is truly a satisficing arm or not. If not, it kicks off the next round of search.

To ensure fast detection of candidate arms with low regret for a given bandit optimization problem class $(\mathcal{A},\mathcal{R})$, \satex~makes use of a bandit algorithm for (standard) regret minimization on $(\mathcal{A},\mathcal{R})$. To this end, we assume access to a blackbox learning oracle that achieves sub-linear standard regret for the bandit optimization problem class $(\mathcal{A},\mathcal{R})$.
\begin{condition}\label{assump:learning_alg}For problem class $(\mathcal{A},\mathcal{R})$, there exists a sequence of learning algorithms $\texttt{ALG}$ such that for any reward function $r\in\mathcal{R}$ and any time horizon $t\geq 2$, the expected regret of $\texttt{ALG}(t)$ is bounded by
\begin{equation*}
t\cdot r(A^*)-\E\left[\sum_{s=1}^t r(A_s)\right]\leq C_1t^\alpha\log(t)^\beta.
\end{equation*}
where $C_1\geq 1,1/2\leq\alpha<1,\beta\geq 0$ are constants only dependent on the problem class and independent from the time horizon $t$.
\end{condition}
We remark that Condition \ref{assump:learning_alg} only asks for sub-linearity in standard regret upper bounds rather than optimality. Therefore, it is an extremely mild condition and can be easily satisfied by a wide range of bandit optimization problem classes and the corresponding bandit learning oracles, \eg, finite-armed bandits, linear bandits, concave bandits, Lipschitz bandits, etc. We will demonstrate this in the forthcoming Section \ref{sec:examples} and Section~\ref{sec:oracle_parameter} of the Appendix.

With this, for a problem class $(\mathcal{A},\mathcal{R})$ and its sub-linear regret learning algorithm $\texttt{ALG}$, \satex~runs in rounds. In round $i$, it takes the following steps (see Fig. \ref{fig:round_satex} for an illustration):
\begin{figure}[!ht]
\centering
\includegraphics[height=9cm]{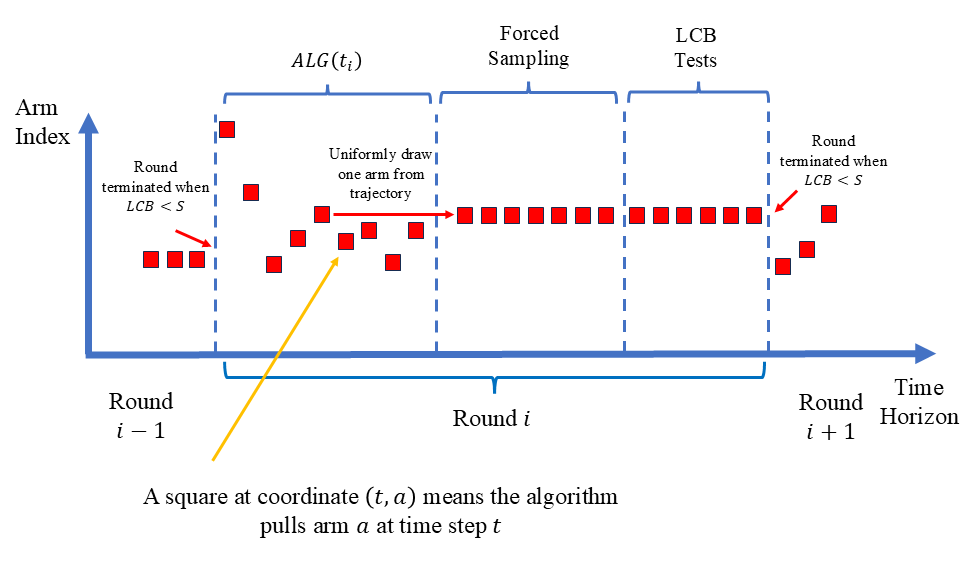}
\caption{Illustration of a single round of~\satex}
\label{fig:round_satex}
\end{figure}

\noindent\textbf{Step 1. Identify a Potentially Satisficing Arm:} Let $\gamma_i=2^{-i(1-\alpha)/\alpha}$, \satex~follows $\texttt{ALG}(t_i)$ for the first $t_i=\lceil \gamma_i^{-1/(1-\alpha)}\rceil=\lceil 2^{i/\alpha}\rceil$ steps of round $i$ and records its arm selections. Then, it samples an arm $\hat{A}_i$ uniformly at random from the trajectory of $\texttt{ALG}(t_i)$ in this round. By virtue of $\texttt{ALG}$, we find an arm whose mean reward is at most $\tilde{O}(t^{\alpha-1}_i)=\tilde{O}(\gamma_i)$ below $r(A^*)$ in expectation, and only an$\tilde{O}(t^{\alpha}_i)$ expected satisficing regret is incurred in this step in the realizable case\footnote{We adopt the asymptotic notations $O(\cdot),\Omega(\cdot),$ and $\Theta(\cdot)$ as defined in \citep{CLRS09}. When logarithmic factors are omitted, we write them as $ \tilde{O}(\cdot),\tilde{\Omega}(\cdot),$ and $\tilde{\Theta}(\cdot).$}. 
We note that in the realizable case, as $i$ increases, $\tilde{O}(\gamma_i)$ will gradually become smaller than $\Delta_S^*$, meaning that the sampled arm $\hat{A}_i$ is more likely to be a satisficing arm and enjoys no satisficing regret;

\noindent\textbf{Step 2. Forced Sampling on the Identified Arm:} To validate if $\hat{A}_i$ is a satisficing arm or not, we need to collect enough data from pulling $\hat{A}_i$ before performing any statistical test (see the forthcoming Remark \ref{remark:lcb} for the reasons). However, when the arm set $\mathcal{A}$ is large, $\texttt{ALG}(t_i)$ might only have pulled $\hat{A}_i$ for a limited number of times. To circumvent this, we perform a forced sampling on $\hat{A}_i$ by pulling it for $T_i=\lceil \gamma^{-2}_i\rceil$ times; 

\noindent\textbf{Step 3. Lower Confidence Bound Tests:} In this step, \satex~kicks off the LCB test as more data is collected from pulling $\hat{A}_i$. We use $k$ (initialized to 0) to denote the number of additional pieces of noisy reward data generated by pulling $\hat{A}_i$ in the current round and $\hat{r}^{\text{tot}}_i$ to denote the running total reward collected from Step 2 and the current step so far. At the same time, it keeps comparing the LCB of $\hat{A}_i$'s mean reward, \ie, $\hat{r}^{\text{tot}}_i/(T_i+k) - \sqrt{4\log(T_i+k)/(T_i+k)}$, against the threshold value $S$ after acquiring every piece of new data. \satex~terminates this round and enters the next whenever the LCB is less than $S$.

The pseudo-code of~\satex~is presented in Algorithm~\ref{alg:sat_exploration}.

\begin{remark}[\textbf{Key Novelties of \satex}]\label{remark:lcb}
Our algorithm brings together several key ingredients, including sampling from oracle trajectories (Step 1), forced sampling to collect data (Step 2), and the LCB test (Step 3). In what follows, we further comment on the subtleties of each step.

\noindent\textbf{Step 1.} Prior works on satisficing (\citealt{GarivierM19, Michel23}) mainly use uniform exploration to find candidates of satisficing arms. This approach could incur major satisficing regret when the arm set is large or even possibly infinite, as conducting uniform exploration on a large arm set may result in pulling a large number of non-satisficing arms. To bypass this challenge, in Step 1, we search for candidates using a bandit learning oracle with sub-linear expected standard regret. Hence, we are able to find candidates of satisficing arms without incurring large regret.

\noindent\textbf{Step 2.} One challenge from using the LCB test in Step 3 is that it is a more conservative design compared to prior works. In fact, if we directly enter Step 3 without Step 2, the LCB of $r(\hat{A_i})$ can easily fall below $S$ even if $r(\hat{A}_i)\geq S$. This is because we may not have pulled $\hat{A}_i$ for sufficiently many times in Step 1 and the corresponding confidence interval might be large. As a result, the lower confidence bound of $r(\hat{A}_i)$, which is the difference between empirical mean and confidence radius, can be significantly below $S$. This indicates, without the forced sampling in Step 2, major satisficing regret can be incurred due to frequent re-start of a new round. With the data collected from Step 2, \satex~ensures that the width of the confidence interval is of order $\tilde{O}(T_i^{-1/2})=\tilde{O}(\gamma_i)$ (\ie, same as $\E[r(A^*)-r(\hat{A}_i)]$). Consequently, whenever $\gamma_i$ shrinks to well below $\Delta_S^*$, \satex~gradually becomes less likely to terminate a round (see the forthcoming Proposition \ref{prop:exit_prob_1});

\noindent\textbf{Step 3.} In Step 3, \satex~compares the LCB of $\hat{A_i}$'s mean reward against $S$ to determine if $\hat{A}_i$ is a satisficing arm. This deviates from prior works that use UCB \citep{GarivierM19} or empirical mean \citep{Michel23}. However, it turns out that this is essential in achieving a constant expected satisficing regret that does not scale with $1/\Delta_S$, which can quickly explode in many cases {(\eg, when the satisficing gaps can be arbitrarily small)}. As we will see, once entering Step 3, \satex~can terminate a round within 1 additional time step in expectation when facing a non-satisficing arm (\ie,  \eqref{eq:real_3} in the proof of the forthcoming Proposition \ref{prop:regret_bound_round_1}). In contrast, if it follows prior works to use UCB or empirical mean instead, the number of time steps required (and hence, the satisficing regret) will unavoidably scale with $1/\Delta_S$ (see \eg, Theorem 9 of \citealt{GarivierM19} or Theorem 1 of \citealt{Michel23}). 
\end{remark}

\begin{algorithm}[!ht]
\caption{Satisficing Regret Minimization via Sampling and Lower Confidence Bound Testing (\satex)}
\label{alg:sat_exploration}
\begin{algorithmic}
\SingleSpacedXI
\State \textbf{Input:} time horizon $T$, satisficing level $S$
\State Set $\gamma_i\leftarrow 2^{-i(1-\alpha)/\alpha}$ for all $i=1,2,3\dots$
\For{round $i=1,2,\dots$}
\State Set $t_i\leftarrow\lceil \gamma_i^{-1/(1-\alpha)}\rceil$
\State Run $\texttt{ALG}(t_i)$, let $\{\bar{A}_s\}_{s\in[t_i]}$ be the trajectory of arm pulls
\State Sample $q$ uniformly at random from $\{1,2,\dots,t_i\}$ and set $\hat{A}_i\leftarrow \bar{A}_q$
\State Set $T_i\leftarrow \lceil\gamma_i^{-2}\rceil$, $k\leftarrow 0$
\State Let $t$ be the current time step, pull arm $\hat{A}_i$ for the next $T_i$ time steps, and set $\hat{r}_i^{\text{tot}}\leftarrow\sum_{s=t}^{t+T_i-1} Y_s$
\State Set $LCB(\hat{A}_i) \leftarrow \hat{r}_i^{\text{tot}}/T_i - \sqrt{4\log(T_i)/T_i}$
\While{$LCB(\hat{A}_i)\geq S$}
\State Set $k\leftarrow k+1$
\State Pull arm $\hat{A}_i$ and observe $Y_t$, where $t$ is the current time step
\State Update $\hat{r}_i^{\text{tot}}\leftarrow \hat{r}_i^{\text{tot}}+Y_t$
\State Set
\begin{equation*}
LCB(\hat{A}_i)\leftarrow\dfrac{\hat{r}_i^{\text{tot}}}{T_i+k}-\sqrt{\dfrac{4\log(T_i+k)}{T_i+k}}
\end{equation*}
\EndWhile
\EndFor
\end{algorithmic}
\end{algorithm}

\section{Regret Analysis}\label{sec:reg_analysis}

In this section, we analyze the regret bounds of \satex. We begin by providing an upper bound of expected satisficing regret in the realizable case, \ie, $r(A^*)\geq S$.

\begin{theorem}\label{thm:regret_bound_1}
If $r(A^*)\geq S$, then the expected satisficing regret of~\satex~is bounded by
\begin{equation}\label{eq:real_1}
\E[\texttt{Regret}_S]\leq\min\left\{ C_1^{\frac{1}{1-\alpha}}\left(\dfrac{1}{\Delta_S^*}\right)^{\frac{\alpha}{1-\alpha}}\cdot\textup{polylog}\left(\dfrac{C_1}{\Delta_S^*}\right),\,C_1T^{\alpha}\cdot\textup{polylog}(T)\right\}.
\end{equation}
\end{theorem}
We remark that, from the first term on the RHS of \eqref{eq:real_1}, \satex~can achieve a constant (w.r.t. $T$) expected satisficing regret in the realizable case. Moreover, even when $T$ is relatively small and the constant expected satisficing regret guarantee is loose compared to the oracle's regret bound, it is able to adapt to the oracle's performance and achieve a sub-linear in $T$ bound on expected satisficing regret. { More specifically, when the exceeding gap $\Delta_S^*$ is large, \ie, $\Delta_S^*\geq C_1T^{-(1-\alpha)}$, the regret bound w.r.t. $\Delta_S^*$ is the smaller one in the two bounds. Hence, in this case, the expected satisficing regret bound scales polynomially in $1/\Delta_S^*$ and does not scale with $T$. When the exceeding gap $\Delta_S^*$ is small, \ie, $\Delta_S^*<C_1T^{-(1-\alpha)}$, the regret bound w.r.t. $T$ is the smaller one in the two bounds. Hence, in this case, the expected satisficing regret bound scales sublinearly in $T$ and does not scale with $1/\Delta_S^*$.}
\begin{proof}[Proof Sketch.]
A complete proof for Theorem~\ref{thm:regret_bound_1}~is provided in Section~\ref{sec:thm:regret_bound_1}. { As discussed in \cref{sec:alg}, there are two key intuitions why \satex~attains low expected satisficing regret: first, by virtue of the standard regret minimization algorithm \texttt{ALG}, Step 1 and Step 2 of \satex~do not incur much satisficing regret, and Step 3 can also quickly remove non-satisficing arms without incurring much satisficing regret; second, as the rounds proceed, \satex~becomes more likely to acquire a candidate arm that is truly satisficing, and the candidate arm is also less likely to be removed by the LCB test in Step 3. In what follows, we present two key results that formalize the two intuitions.
 
To formalize the first intuition, we use the following proposition to establish an upper bound on the expected satisficing regret incurred in each round.}
\begin{proposition}\label{prop:regret_bound_round_1}
If $r(A^*)\geq S$, then the expected satisficing regret~incurred in round $i$ is bounded by
\begin{equation*}
\dfrac{{ 8}C_1}{(1-\alpha)^\beta}\gamma_i^{-\alpha/(1-\alpha)}\log({4}/\gamma_i)^{\beta}.
\end{equation*}
\end{proposition}
The proof of this proposition is provided in Section \ref{sec:prop:regret_bound_round_1} of the appendix. 

{ To formalize the second intuition, we use the following proposition to show that once \satex~runs for enough rounds so that $\gamma_i=\tilde{O}(\Delta_S^*)$, it becomes less likely to start a new round, \ie, the probability of starting a new round is upper bounded by $1/4$.} 
\begin{proposition}\label{prop:exit_prob_1}
If $r(A^*)\geq S$, then for every $i$ that satisfies
\begin{equation}\label{eq:large_enough_i_1}
\dfrac{32C_1}{(1-\alpha)^\beta}\gamma_i\log({4}/\gamma_i)^{\max\{\beta,1/2\}}\leq \Delta_S^*\,,
\end{equation}
the probability that round $i$ ends within the time horizon $T$ (conditioned on the event that round $i$ is started within the time horizon) is { upper} bounded by $1/4$.
\end{proposition}
The proof of this proposition is provided in Section \ref{sec:prop:exit_prob_1} of the appendix. Since each round of~\satex~runs independently, Proposition~\ref{prop:exit_prob_1}~indicates that the total number of rounds for~\satex~can be upper-bounded by a shifted geometric random variable with success rate $3/4$.

{With \cref{prop:regret_bound_round_1} and \cref{prop:exit_prob_1}, we are ready to finish the proof of \cref{thm:regret_bound_1}. We denote {$i_0+1$} as the minimum integer such that \eqref{eq:large_enough_i_1} holds. We prove that $2^{i_0(1-\alpha)/\alpha}=1/\gamma_{i_0}=C_1/\Delta_S^*\cdot\text{polylog}(C_1/\Delta_S^*)$. We further have $i_0=\Theta(\log(C_1/\Delta_S^*))$. Therefore by \cref{prop:regret_bound_round_1} the expected satisficing regret incurred in the first $i_0$ rounds is bounded by
\begin{equation*}
\sum_{i=1}^{i_0}\dfrac{{8}C_1}{(1-\alpha)^\beta}\gamma_i^{-\frac{\alpha}{1-\alpha}}\log({4}/\gamma_i)^\beta\leq C_1^{\frac{1}{1-\alpha}}\left(\dfrac{1}{\Delta_S^*}\right)^{\frac{\alpha}{1-\alpha}}\cdot\text{polylog}\left(\dfrac{C_1}{\Delta_S^*}\right).
\end{equation*}
Then we bound the expected satisficing regret incurred after round $i_0$. By \cref{prop:exit_prob_1}, { round $i_0+k-1$ ends within the time horizon and \satex~further} enters round $i_0+k$ with probability at most $4^{{1}-k}$. Therefore the expected satisficing regret after round $i_0$ is upper bounded by
\begin{equation*}
    \sum_{k=1}^\infty \dfrac{{8}C_1}{(1-\alpha)^\beta}\gamma_{i_0+k}^{-\frac{\alpha}{1-\alpha}}\log({4}/\gamma_{i_0+k})^\beta\cdot 4^{{1}-k} = \dfrac{{32}C_1}{(1-\alpha)^\beta}\gamma_{i_0}^{-\frac{\alpha}{1-\alpha}}\sum_{k=1}^\infty 2^{-k}\left(\dfrac{(1-\alpha)(i_0+k)}{\alpha}+1\right)^\beta\log(2)^\beta.
\end{equation*}
We have shown that $\gamma_{i_0}^{-\alpha/(1-\alpha)}=(C_1/\Delta_S^*)^{\alpha/(1-\alpha)}\cdot \text{polylog}(C_1/\Delta_S^*)$. We further show that the summation above is polynomial in $i_0$, thus polynomial in $\log(C_1/\Delta_S^*)$. Therefore the expected satisficing regret is upper bounded by $C_1^{1/(1-\alpha)}(1/\Delta_S^*)^{\alpha/(1-\alpha)}\cdot\text{polylog}(C_1/\Delta_S^*)$. Therefore the expected satisficing regret is upper bounded by the first term in the regret bound.

We further show that the expected satisficing regret is upper bounded by the second term in the regret bound. Let $i_{\max}$ be the total number of rounds for \satex. We show that $t_{i_{\max}}=\gamma_{i_{\max}}^{-1/(1-\alpha)}=O(T)$, thus $i_{\max}=O(\log(T))$. By \cref{prop:regret_bound_round_1}, the expected satisficing regret { incurred in} round { $i$} is upper bounded by {$\tilde{O}(C_1\gamma_i^{-\alpha/(1-\alpha)}) = \tilde{O}\left(C_1t_i^\alpha\right) \leq \tilde{O}\left(C_1T^\alpha\right)$}. Since the total number of rounds is upper bounded by $O(\log(T))$, the expected satisficing regret of \satex~is also upper bounded by $C_1T^{\alpha}\cdot\text{polylog}(T)$.
} 
\end{proof}

We also show that, in the non-realizable case, \satex~enjoys the same regret bound as the oracle in Condition~\ref{assump:learning_alg} (The proof is provided in Section \ref{sec:thm:regret_nonsat} of the appendix).
\begin{theorem}\label{thm:regret_nonsat}
If $r(A^*)<S$, the expected standard regret of \satex~is bounded by
\begin{equation*}
\E[\regret]\leq C_1T^{\alpha}\cdot\textup{polylog}(T).
\end{equation*}
\end{theorem}

Proof of \cref{thm:regret_nonsat} is provided in \cref{sec:thm:regret_nonsat}. \cref{thm:regret_nonsat} implies that the expected standard regret of \satex~is sub-linear in $T$ in the non-realizable case. The expected standard regret bound of \satex~matches that of the learning oracle within logarithmic factors of $T$, indicating that in the non-realizable case, \satex~attains a similar performance compared to the standard regret minimization algorithm in the corresponding class of bandit optimization problems.

With the above results, we establish that under a very general condition, \satex~does achieve constant expected satisficing regret in the realizable case without compromising the standard regret guarantee in the non-realizable case. 

\section{Examples}\label{sec:examples}
In what follows, we instantiate \satex~to several popular classes of bandit optimization problems, including finite-armed bandits, concave bandits, and Lipschitz bandits. Along the way, we showcase how our results enable constant expected satisficing regret for bandits with large and even infinite arm set, which was not achieved by existing algorithms. We remark that the examples given here are non-exhaustive and similar results can be derived for other classes of bandit optimization problems such as linear bandits. { In Section~\ref{sec:lower_bounds}, we further present lower bounds on the expected satisficing regret that match the satisficing regret upper bounds attained by \satex~in finite-arm bandits, concave bandits and Lipschitz bandits. The matching lower bound results show that \satex~attains near-optimal expected satisficing regret bound in all three bandit problem classes considered in this section.}

\vspace{2mm}
\noindent\textbf{1. Finite-Armed Bandits:} Consider any instance of finite-armed bandit with $K$ arms, \ie, $\mathcal{A}=[K]$. In this case, both the UCB algorithm (see, \eg, \citealt{BubeckC12}) and Thompson sampling (see, \eg, \citealt{AgrawalG17}) achieve a regret bound of $O(\sqrt{KT\log(T)})$. Combining them with Theorems~\ref{thm:regret_bound_1} and \ref{thm:regret_nonsat}, we have the following corollary.

\begin{corollary}\label{cor:k_arm_bandit}
By using either the UCB algorithm or Thompson sampling as $\texttt{ALG}$, if $r(A^*)\geq S$, the expected satisficing regret of \satex~is bounded by
\begin{equation*}
\E[\regret_S]\leq \min\left\{\dfrac{K}{\Delta_S^*}\cdot\textup{polylog}\left(\dfrac{K}{\Delta_S^*}\right),\,\sqrt{KT}\cdot\textup{polylog}(T)\right\};
\end{equation*}
If $r(A^*)<S$, the expected standard regret of \satex~is bounded by $\mathbb{E}[\regret]\leq \sqrt{KT}\cdot\textup{polylog}(T).$
\end{corollary}

\begin{remark}[\textbf{Comparison with \citealt{GarivierM19,Michel23}}]
{\cite{GarivierM19,Michel23}} also provide algorithms for minimizing the expected satisficing regret in finite-armed bandit settings. While our expected satisficing regret bound is incomparable to {\cite{GarivierM19}}, which attains an $O(K/\Delta_S)$ expected satisficing regret, we provide major improvement over {\cite{Michel23}}, which achieves $O(K/\Delta_S+K/(\Delta_S^*)^2)$ expected satisficing regret. Compared to {\cite{Michel23}}, our regret bound removes the dependence on $\Delta_S$ and the additional $1/\Delta_S^*$ factor. We also point out that if we want to achieve the same expected satisficing regret as {\cite{GarivierM19}}, we can simply change the LCB test in Step 3 to a UCB test. 
\end{remark}

{
\begin{remark}[\textbf{Best of Both Worlds}]
When comparing our expected regret bound in \cref{cor:k_arm_bandit} with the one in \cite{GarivierM19} (see Theorem 9 therein), one can see that our algorithm \satex~potentially has a better performance when $\Delta_S^*$ is large and $\Delta_S$ is small, while the algorithm in \cite{GarivierM19} potentially has a better performance when $\Delta_S^*$ is small and $\Delta_S$ is large. In order to establish an algorithm that attains a ``best of both worlds" performance guarantee and works well in both cases, in \cref{sec:best_of_both_world} of the appendix, we propose \texttt{SELECT-BoBW}, an adapted version of \satex~for finite-arm bandits that attains an expected satisficing regret bound of $\tilde{O}(\min\{K/\Delta_S,K/\Delta_S^*\})$. The regret bound achieves a ``best of both worlds", which improves upon the bounds in \cite{GarivierM19,Michel23} as well as the bound of \satex. Specifically, \texttt{SELECT-BoBW} is modified based on \texttt{SELECT} by adding a UCB test in the forced sampling in Step 2, which compares the UCB of candidate arm mean reward against the threshold $S$, during forced sampling. Whenever the UCB of the candidate satisficing arm falls below $S$, the forced sampling is terminated early and \texttt{SELECT-BoBW} immediately proceeds to the next round. We present a full discussion of \texttt{SELECT-BoBW} in Section~\ref{sec:best_of_both_world} in the appendix.
\end{remark}
}

\vspace{2mm}
\noindent\textbf{2. Concave Bandits:} Consider any instance of the concave bandit, \ie, $\mathcal{A}\subset\mathbb{R}^d$ is a bounded convex set, and $r(A)$ is a concave and $1$-Lipschitz continuous function defined on $\mathcal{A}$. \cite{AgarwalFH11} gives an algorithm that enjoys $\text{poly}(d)\sqrt{T}\cdot\text{polylog}(T)$ regret. Together with Theorems~\ref{thm:regret_bound_1}~and \ref{thm:regret_nonsat}, we have the following corollary.
\begin{corollary}\label{cor:concave_bandit}
By using the algorithm in~{\cite{AgarwalFH11}}~as $\texttt{ALG}$, if $r(A^*)\geq S$, then the expected satisficing regret of~\satex~is bounded by
\begin{equation*}
\E[\regret_S]\leq\min\left\{\dfrac{\textup{poly}(d)}{\Delta_S^*}\cdot\textup{polylog}\left(\dfrac{d}{\Delta_S^*}\right),\textup{poly}(d)\sqrt{T}\cdot\textup{polylog}(T)\right\};
\end{equation*}
If $r(A^*)<S$, the expected standard regret of~\satex~is bounded by
$\E[\regret]\leq \textup{poly}(d)\sqrt{T}\cdot\textup{polylog}(T).$
\end{corollary}

\vspace{2mm}
\noindent\textbf{3. Lipschitz Bandits:} Consider any instance of the Lipschitz continuous bandit in $d$ dimensions, \ie, $\mathcal{A}=[0,1]^d$, and $r(A)$ is an $L$-Lipschitz function (here Lipschitz continuous functions are defined in the sense of $\infty$-norm). \cite{BubeckSY2011} introduces a uniformly discretized UCB algorithm to achieve an $O(L^{d/(d+2)}T^{(d+1)/(d+2)}\sqrt{\log(T)})$ regret upper bound. Together with Theorem~\ref{thm:regret_bound_1}~and Theorem~\ref{thm:regret_nonsat}, we have the following corollary.

\begin{corollary}\label{cor:lipschitz_bandit}
By using the UCB algorithm with uniform discretization in {\cite{BubeckSY2011}} as $\texttt{ALG}$, if $r(A^*)\geq S$, then the expected satisficing regret of \satex~is bounded by
\begin{equation*}
\E[\regret_S]\leq \min\left\{\dfrac{L^d}{(\Delta_S^*/2)^{d+1}}\cdot\textup{polylog}\left(\dfrac{L}{\Delta_S^*}\right),\,L^{\frac{d}{d+2}}T^{\frac{d+1}{d+2}}\cdot\textup{polylog}(T)\right\};
\end{equation*}
If $r(A^*)<S$, the expected standard regret of \satex~is bounded by
$\mathbb{E}[\regret]\leq L^{\frac{d}{d+2}}T^{\frac{d+1}{d+2}}\cdot\textup{polylog}(T).$
\end{corollary}

\begin{remark}
Recall that $\Delta_S=\min\{S-r(A):r(A)<S\}$, one can easily verify that in the realizable case, $\Delta_S=0$ for both concave bandits and Lipschitz bandits. As such, one cannot directly apply the results in {\cite{GarivierM19,Michel23}} to acquire a constant expected satisficing regret. With the notion of exceeding gap $\Delta_S^*$ and \satex, we establish constant expected satisficing regret bounds for these two classes of bandit optimization problems.
\end{remark}

\subsection{Lower Bounds for Satisficing Regret Minimization}\label{sec:lower_bounds}
To complement our results, we also present the expected satisficing regret lower bounds for { finite-arm bandits, bandits with concave rewards and Lipschitz bandits. By establishing lower bounds on the expected satisficing regret, we show that \satex~attains a near-optimal expected satisficing regret bound for all three aforementioned bandit problem classes.} 

The first one is an expected satisficing regret lower bound for the finite-armed bandits.

{
\begin{theorem}[Finite-Armed Bandits]\label{thm:lb_two_arm}
For every non-anticipatory learning algorithm $\pi$, every $0<\Delta<1/20$, every $K\in\mathbb{Z}_+$, $K\geq2$, and $T\geq K/\Delta^2$, there exists an instance of finite-arm bandit with $K$ arms such that $r(A^*)-S=\Delta$, and the expected satisficing regret incurred by $\pi$ is at least $\Omega(K/\Delta)$.
\end{theorem}

Theorem~\ref{thm:lb_two_arm} establishes a $\Omega(K/\Delta_S^*)$ lower bound on the expected satisficing regret for finite-arm bandits. The lower bound established in Theorem~\ref{thm:lb_two_arm}~matches the expected satisficing regret upper bound of \satex~in~\cref{cor:k_arm_bandit}~up to a logarithmic factor of $K/\Delta_S^*$. This also suggests that the expected satisficing regret bound of \satex~is near-optimal in finite-arm bandits.

}
\begin{remark}
In {\cite{Michel23}}, the authors adapt the results from {\cite{BubeckPR13}}~(for MAB with known optimal mean reward) to establish a $\Omega(K/\Delta_S)$ expected satisficing regret lower bound for finite-armed bandits. This is different from the one in Theorem \ref{thm:lb_two_arm}, which is $\Omega(K/\Delta)$ with $\Delta$ being the exceeding gap (\ie, $\Delta_S^*$). This difference originates from the lower bound instances. The lower bound instance in {\cite{BubeckPR13}} (when adapted to satisficing bandits) sets the threshold value to the optimal mean reward, \ie, $S=r(A^*)\,,$ which makes both the expected satisficing regret and standard regret lower bounds to be $\Omega(K/\Delta_S)$. In our lower bound proof, we allow $S$ to be smaller than $r(A^*)$, which leads to a lower bound that depends on the difference between these two quantities. 
\end{remark}

Next, we provide an expected satisficing regret lower bound for bandits with concave rewards. 
\begin{theorem}[Bandits with Concave Rewards]\label{thm:lb_concave}
For every non-anticipatory learning algorithm $\pi$, every $0<\Delta<{ 1/162}$ and every $T\geq 1/\Delta^2$, there exists an instance of one-dimensional bandit with concave reward such that $r(A^*)-S=\Delta$, and the expected satisficing regret incurred by $\pi$ is at least $\Omega(1/\Delta)$.
\end{theorem}

\cref{thm:lb_concave}~establishes a $\Omega(1/\Delta_S^*)$ lower bound on the expected satisficing regret for concave bandits. The lower bound established in \cref{thm:lb_concave} matches the expected satisficing regret upper bound of \satex~in \cref{cor:concave_bandit} up to a polynomial factor of $d$ and a logarithmic factor of $d/\Delta_S^*$. If we consider the dimension $d$ as a constant, then the expected satisficing regret bound of \satex~for concave bandits is near-optimal up to a logarithmic factor of $d/\Delta_S^*$.

{

Finally, we provide an expected satisficing regret lower bound for Lipschitz bandits.

\begin{theorem}[Lipschitz Bandits]\label{thm:lb_lipschitz}
For every non-anticipatory learning algorithm $\pi$, every $0<\Delta<1/20$, every $d\in\mathbb{Z}_+$, Lipschitz constant $L\geq 1$ and time horizon $T\geq L^d/\Delta^{d+2}$, there exists an instance of $d$-dimensional Lipschitz bandit such that $r(A^*)-S=\Delta$, and the expected satisficing regret incurred by $\pi$ is at least $\Omega(L^{d}/\Delta^{d+1})$.
\end{theorem}

Theorem~\ref{thm:lb_lipschitz}~establishes a $\Omega(L^d/(\Delta_S^*)^{d+1})$ lower bound on the expected satisficing regret for Lipschitz bandits. The lower bound established in Theorem~\ref{thm:lb_lipschitz}~matches the expected satisficing regret upper bound of~\satex~in Corollary~\ref{cor:lipschitz_bandit} up to a logarithmic factor of $L/\Delta_S^*$.  This also suggests that the expected satisficing regret upper bound provided by \satex~is near-optimal in Lipschitz bandits.
}

\section{Tail Risk Control in Satisficing Regret Minimization}\label{sec:satisficing_tail}
While \satex~achieves a constant satisficing regret in expectation, it is unclear if it enjoys a light-tailed distribution on the satisficing regret (\ie, the realized satisficing regret is unlikely to be much larger than the expectation), a highly desirable property in maintaining the stability of an algorithm \citep{fan2024fragility,simchi2023stochastic}. { We say the satisficing regret of an algorithm is {\it light-tailed} if the probability of satisficing regret being greater than $x$ decays at a rate of $\exp(-x^\zeta)$ for some $\zeta>0$ as $x$ increases. In other words, the satisficing regret distribution is light-tailed if there exists $\zeta\in(0,1)$ and $\Lambda_1,\Lambda_2$, such that for any $T$ and any $0\leq x\leq T$,
\begin{equation}
    \Pr(\regret_S>x)\leq\Lambda_1\exp\left(-x^\zeta/\Lambda_2\right).
\label{eq:light_tail_def}
\end{equation}
Here $\Lambda_1,\Lambda_2>0$ are parameters independent of $x$. We remark that $\Lambda_1$ and $\Lambda_2$ can depend on the length of the time horizon $T$, but we require that $\Lambda_1$ should scale at most polynomially with $T$, and $\Lambda_2$ should scale at most poly-logarithmically with $T$. On the other hand, any algorithm whose satisficing regret distribution does not satisfy \eqref{eq:light_tail_def} is said to have a {\it heavy-tailed} satisficing regret distribution.}

A light-tailed satisficing regret distribution is critical in many practical settings of satisficing regret minimization, particularly in applications sensitive to tail risk such as healthcare and supply chain. In these applications, a light-tailed satisficing regret distribution reduces the possibility of suffering substantial losses for the decision maker, thus providing the decision maker with more protection against extreme scenarios. For instance, in the inventory management example discussed in Section \ref{sec:intro}, if an algorithm is able to achieve a light-tailed satisficing regret distribution, it indicates it is highly unlikely for the retailers to frequently fall short of the designated service level over time. 

\subsection{Key Challenges and Insufficiency of \satex}
Based on prior results (see, \eg, \citealt{fan2024fragility}, \citealt{simchi2023stochastic}, \citealt{simchi2024simple}), achieving a light-tailed satisficing regret distribution while still maintaining a constant expected satisficing regret can be extremely challenging due to the following reasons:
\begin{enumerate}
\item \textbf{Tradeoff between Instance-Dependency and Tail Risk:} Existing works show that in standard regret minimization for finite-armed bandits, a worse instance-dependent expected standard regret is inevitable for any algorithm with a light-tailed standard regret distribution. In particular, \cite{simchi2024simple} prove that any algorithm with a light-tailed standard regret distribution has to suffer a $\Omega(\text{poly}(T))$ instance-dependent expected standard regret, rather than a near-optimal $O(\log(T))$ instance-dependent expected standard regret for the UCB algorithm and Thompson sampling. Since existing constant expected satisficing regret bounds (\eg, \citealt{GarivierM19}, \citealt{Michel23}, and ours) are all instance-dependent, one may conjecture that in satisficing regret minimization, a light-tailed satisficing regret distribution could also inevitably come at a cost of worse (\eg, $\text{polylog}(T)$ or $\text{poly}(T)$) instance-dependent expected satisficing regret. Then, it becomes unclear whether a constant expected satisficing regret bound can coexist with a light-tailed satisficing regret distribution even in the finite-armed bandit setting.
\item \textbf{Heavy-Tailed Learning Oracle:} To the best of our knowledge, existing works on tail risk control for standard regret focus mostly on finite-armed bandits and linear bandits, and it remains unclear whether there exists learning oracles with light-tailed regret distribution for more general classes of bandit optimization problems. Therefore, we may have to start with learning oracles with potentially heavy-tailed standard regret distribution. This makes our algorithm design even more challenging in achieving a light-tailed satisficing regret distribution for more general classes of bandit optimization problems.
\end{enumerate}
Before proceeding, we first confirm the necessity of new algorithmic techniques in achieving a light-tailed satisficing regret distribution. We use the following proposition to show that the satisficing regret distribution of \satex~can indeed be heavy-tailed even if the learning oracle $\texttt{ALG}$ used is already light-tailed.
\begin{proposition}\label{thm:heavy_tail_2arm}
For every finite-armed bandit learning oracle with $\tilde{O}(\sqrt{T})$ expected standard regret, { if the oracle used in each round selects each arm at least once with probability at least $1/2$,} then there exists an instance of two-armed bandit such that for all $T{ \geq 8}$, the satisficing regret of \satex~satisfies
$$
\Pr\left(\texttt{Regret}_S\geq \frac{T}{{ 14}} \right)\geq { \frac{1}{4}\exp(-2\log(T)^2)} .
$$
\end{proposition}

The proof of \cref{thm:heavy_tail_2arm} is provided in Section \ref{sec:prop:heavy_tail_2arm} of the appendix. \cref{thm:heavy_tail_2arm} states that the satisficing regret distribution resulting from \satex~is heavy-tailed under mild conditions of the learning oracle. In fact, such conditions in \cref{thm:heavy_tail_2arm} hold widely in most of the existing algorithms for finite-armed bandits, such as the UCB algorithm, Thompson sampling, and even for the algorithm in \cite{simchi2024simple}, whose standard regret distribution is known to be light-tailed. This suggests that simply applying \satex~does not guarantee a light-tailed satisficing regret distribution, even if the learning oracle adopted is light-tailed. Therefore, a new algorithm is needed if we want to ensure a light-tailed satisficing regret distribution.

\subsection{\satexlt: \satex~with Light-Tailed Satisficing Regret Distribution}

In this section, we present \satexlt, an augmented algorithmic template with a light-tailed satisficing regret distribution for more general classes of bandit optimization problems.

We note that there are two factors that prevent \satex~from achieving a light-tailed satisficing regret distribution. First, the length $t_i$ of Step 1 and the length $T_i$ of Step 2 increase exponentially as rounds proceed. Since larger $t_i$ and $T_i$ often result in larger tail risk, the rapid growth of $t_i$ and $T_i$ can lead to uncontrolled tail risk, eventually resulting in a heavy-tailed satisficing regret distribution. Second, in the LCB test in Step 3 of \satex, we construct a sequence of confidence intervals for the candidate satisficing arm such that the $k$-th confidence interval contains the true mean reward with probability $1-\text{poly}(1/k)$. Although we have shown that these confidence intervals guarantee removal of a non-satisficing candidate arm in constant time steps in expectation, the distribution of the number of steps in the LCB test could also be heavy-tailed, potentially leading to a heavy-tailed satisficing regret distribution.

In light of these observations, we make two modifications to \satex~to make the satisficing regret distribution light-tailed. First, in the modified algorithm, we set the length $t_i'$ of Step 1 and the length $T_i'$ of Step 2 to scale polynomially with the number of rounds $i$ instead of exponentially. Second, we use larger confidence intervals in the LCB test such that the $k$-th confidence interval contains the mean reward with probability at least $1-\exp(-\Omega(k^\zeta))$. With this configuration of confidence intervals, if the candidate satisficing arm is non-satisficing, then the probability that the LCB test lasts for more than $k$ time steps is at most $\exp(-\Omega(k^\zeta))$, thus we are able to guarantee that the number of steps for the LCB test in each round follows a light-tailed regret distribution.

We refer to this algorithm as \satexlt, and the complete algorithm is provided in \cref{alg:sat_lite}. Here, $\Gamma(z)$ is the gamma function given by $\Gamma(z)=\int_0^\infty t^{z-1}\exp(-t)\text{d}t$. 

\begin{algorithm}[!ht]
\caption{Satisficing Regret Minimization with Light-Tailed Satisficing Regret Distribution (\satexlt)}
\label{alg:sat_lite}
\begin{algorithmic}
\SingleSpacedXI
\State \textbf{Input:} time horizon $T$, satisficing level $S$, tail distribution parameter $\zeta$
\State Set $\gamma_i\leftarrow i^{-(1/\zeta-1)(1-\alpha)}$ for all $i=1,2,3\dots$
\For{round $i=1,2,\dots$}
\State Set $t'_i\leftarrow\lceil \gamma_i^{-1/(1-\alpha)}\rceil$
\State Run $\texttt{ALG}(t'_i)$, let $\{\bar{A}_s\}_{s\in[t'_i]}$ be the trajectory of arm pulls
\State Sample $q$ uniformly at random from $\{1,2,\dots,t'_i\}$ and set $\hat{A}_i\leftarrow \bar{A}_q$
\State Set $T'_i\leftarrow \lceil\gamma_i^{-2}\rceil$, $k\leftarrow 0$
\State Let $t$ be the current time step, pull arm $\hat{A}_i$ for the next $T'_i$ time steps, and set $\hat{r}_i^{\text{tot}}\leftarrow\sum_{s=t}^{t+T'_i-1} Y_s$
\State Set $LCB(\hat{A}_i) \leftarrow \hat{r}_i^{\text{tot}}/T_i' - \sqrt{\log\left({16}\zeta^{-1}\Gamma(\zeta^{-1})\right)/T_i'}$
\While{$LCB(\hat{A}_i)\geq S$}
\State Set $k\leftarrow k+1$
\State Pull arm $\hat{A}_i$ and observe $Y_t$, where $t$ is the current time step
\State Update $\hat{r}_i^{\text{tot}}\leftarrow \hat{r}_i^{\text{tot}}+Y_t$
\State Set
\begin{equation*}
LCB(\hat{A}_i)\leftarrow\dfrac{\hat{r}_i^{\text{tot}}}{T'_i+k}-\sqrt{\frac{k^\zeta+\log\left({16}\zeta^{-1}\Gamma(\zeta^{-1})\right)}{T_i'+k}}
\end{equation*}
\EndWhile
\EndFor
\end{algorithmic}
\end{algorithm}

\subsection{Tail Risk and Regret Analysis of \satexlt}

In this section, we provide a tail risk and regret analysis of \satexlt. We show that \satexlt~has the following desirable properties: in the realizable case, \satexlt~enjoys both a light-tailed satisficing regret distribution and a constant expected satisficing regret bound at the same time; in the non-realizable case, \satexlt~still maintains a sub-linear expected standard regret bound.

We first use the following theorem to show that \satexlt~indeed has a light-tailed satisficing regret distribution in the realizable case.

\begin{theorem}\label{thm:tail_risk}
If $r(A^*)\geq S$, then for all $x>0$, the probability that satisficing regret of \satexlt~exceeds $x$ is bounded by
\begin{equation*}
    \Pr\left(\texttt{Regret}_S>x\right) \leq { \exp\left(2i_0 + 4-\left(\frac{(x-x_0)_+}{9+(2(1-\zeta)/3)^{-1}(1-\alpha)^{-1}} \right)^\zeta \right)} + \exp\left({ i_0 +2-\frac{(x-x_0)_+^\zeta}{4}} \right),
\end{equation*}
where $i_0=O\left(\left(1/\Delta_S^*\cdot\log(1/\Delta_S^*)^{\beta'}\right)^{\frac{\zeta}{(1-\zeta)^2(1-\alpha)}}\right)$ and $x_0=6(i_0+\zeta(i_0+1)^{1/\zeta})$ are constants independent of $x$ and $T$.
\end{theorem}

\begin{proof}[Proof Sketch]

The complete proof of \cref{thm:tail_risk} is provided in \cref{sec:thm:tail_risk} of the appendix. Similar to \cref{prop:exit_prob_1}, we first prove that in \satexlt, the probability of round $i$ terminating within the time horizon is at most $1/4$ for all large enough $i$.

\begin{proposition}\label{prop:exit_prob_tail}
If $r(A^*)\geq S$, then for all $i$ that satisfies
\begin{equation}
\dfrac{32C_1(1-\zeta)^{-\zeta/2}}{(1-\alpha)^\beta} {\sqrt{2 + 2\log\left(16\zeta^{-1}\Gamma(\zeta^{-1})\right)}} \gamma_i^{1-\zeta} \log({4}/\gamma_i)^{\max\{\beta,1/2\}}\leq \Delta_S^*,\label{eq:large_enough_i_tail}
\end{equation}
the probability that round $i$ ends within the time horizon $T$ (conditioned on the event that round $i$ is started within the time horizon) is { upper} bounded by $1/4$.
\end{proposition}

\cref{prop:exit_prob_tail} implies that the total number of rounds for \satexlt~can be upper-bounded by a shifted geometric random variable with success rate $3/4$. Proof of \cref{prop:exit_prob_tail} is provided in \cref{sec:prop:exit_prob_tail} of the appendix. 

The inequality in the theorem is trivial when $x<x_0$, thus we focus on the case where $x\geq x_0$. We denote $i_0$ as the smallest $i$ that satisfies \eqref{eq:large_enough_i_tail}. Since $x>x_0$, we get $x\geq 3\sum_{i=1}^{i_0}(t_i'+T_i')$. Then we divide the satisficing regret of \satexlt~into three parts: 1. the satisficing regret incurred in Step 1 and 2 by round $i_0$; 2. the satisficing regret incurred in Step 1 and 2 after round $i_0$; 3. the satisficing regret incurred in Step 3 of all rounds. Since $x\geq 3\sum_{i=1}^{i_0}(t_i'+T_i')$, the first part of satisficing regret is at most $x/3$. Therefore if the total satisficing regret exceeds $x$, then either the second or the third part of satisficing regret exceeds $x/3$.

To bound the tail distribution of the second part of satisficing regret, we first define $k_0$ as the maximum integer such that $\sum_{i=i_0+1}^{i_0+k_0}(t_i'+T_i')\leq x/3$. Then we verify that $k_0=\Omega(x^\zeta)$. If the second part of satisficing regret exceeds $x/3$, then the total number of rounds should be at least $i_0+k_0+1$, and by \cref{prop:exit_prob_tail}, \satexlt~enters round $i_0+k_0+1$ { (\ie, round $i_0+k_0$ terminates within the time horizon)} with probability at most $4^{-k_0}=\exp(-\Omega(x^\zeta))$. Therefore the probability that the second part of satisficing regret exceeds $x/3$ is at most $\exp(-\Omega(x^\zeta))$.

To bound the tail distribution of the third part of satisficing regret, by the definition of confidence intervals we have that the probability that the number of steps in the LCB test exceeds $k$ is $O(\exp(-k^\zeta))$. Furthermore, by \cref{prop:exit_prob_tail}, the total number of rounds of \satexlt~is upper bounded by a shifted geometric random variable with success probability $3/4$. Then we use the following proposition to bound the tail distribution of the third part of satisficing regret.

\begin{proposition}\label{lemma:round_exp_all}
Suppose $\{Q_i\}$ are independent nonnegative random variables and there exists $K_1>0$, $0<\zeta<1$ such that for all $x>0$ and all $i\in\mathbb{Z}^+$,
\begin{equation*}
\Pr(Q_i\geq x)\leq K_1\exp(-x^\zeta).
\end{equation*}
Let $N(\lambda)$ be a geometric distribution with success rate being $1-\lambda$, { where $\lambda<1/2$}. Then for all $x>0$, we have
{
\begin{equation*}
\Pr\left(\sum_{i=1}^{N+N(\lambda)}Q_i\geq x\right)\leq \dfrac{2^{N+1}(1-\lambda)}{1-2\lambda}\exp\left(-\dfrac{x^\zeta}{K_1+1}\right).
\end{equation*}}
\end{proposition}
\cref{lemma:round_exp_all} states that the sum of a random number of independent light-tailed random variables is still light-tailed if the number of random variables follows a shifted geometric distribution. Proof of \cref{lemma:round_exp_all} is provided in \cref{sec:lemma:round_exp_all}. By \cref{lemma:round_exp_all}, the probability that the total number of time steps for Step 3 exceeds $x/3$ is at most $\exp(-\Omega(x^\zeta))$. Therefore the probability that the satisficing regret incurred in Step 3 exceeds $x/3$ is at most $\exp(-\Omega(x^\zeta))$. Therefore we conclude that the probability that the satisficing regret of \satexlt~exceeds $x$ is at most $\exp(-\Omega(x^\zeta))$ for large enough $x$. 
\end{proof}

\begin{remark}[\textbf{Independence of $T$ for the Tail Probability Bound}]
We would like to point out that the tail probability bound of \satexlt~does not depend on $T$. In other words, for any realizable instance, there exists a fixed light-tailed distribution that dominates the distribution of satisficing regret of \satexlt~with any time horizon $T$. This is in sharp contrast with the tail probability bounds provided in prior works such as \cite{simchi2024simple}, which deteriorate as $T$ increases although the standard regret distribution for every fixed $T$ is light-tailed.
\end{remark}

We use the following theorem to show that \satexlt~enjoys a constant expected satisficing regret bound.

\begin{theorem}\label{thm:regret_bound_tail}
If $r(A^*)\geq S$, then the expected satisficing regret of \satexlt~is bounded by
\begin{equation*}
\mathbb{E}[\texttt{Regret}_S]\leq\min\left\{ C_1\left(\dfrac{C_1}{\Delta_S^*}\right)^{\frac{\zeta+\alpha(1-\zeta)}{(1-\zeta)^2(1-\alpha)}}\cdot\textup{polylog}\left(\dfrac{C_1}{\Delta_S^*}\right),\,C_1T^{\alpha(1-\zeta)+\zeta}\cdot\textup{polylog}(T)\right\}.
\end{equation*}
\end{theorem}

Proof of \cref{thm:regret_bound_tail} is similar to that of \cref{thm:regret_bound_1} and is provided in \cref{sec:thm:regret_bound_tail} of the appendix. \cref{thm:regret_bound_tail} states that for all $0<\zeta<1$, \satexlt~attains a constant expected satisficing regret bound depending on $\zeta$. The power of $1/\Delta_S^*$ in the expected satisficing regret bound is monotonically increasing in $\zeta$, suggesting that stronger guarantees on tail distribution of satisficing regret lead to larger constant satisficing regret on its expectation. When $\zeta$ approaches zero, the power of $1/\Delta_S^*$ approaches $\alpha/(1-\alpha)$, which is the same as the one in the expected satisficing regret bound of \satex. This implies that when $\zeta$ is small, \satexlt~can have a similar expected satisficing regret as \satex.

{ We would like to remark that although both \satex~and \satexlt~are able to attain a constant expected satisficing regret bound, the expected satisficing regret bound of \satexlt~is always weaker than that of \satex. Specifically, the constant bound of \satexlt~scales with $C_1/\Delta_S^*$ to the power of $(\zeta+\alpha(1-\zeta))/(1-\zeta)^2(1-\alpha)$, while the constant bound of \satex~scales with $C_1/\Delta_S^*$ to the power of $\alpha/(1-\alpha)$. Since $0<\zeta<1$ and $0<\alpha<1$, we have $\zeta+\alpha(1-\zeta)>\alpha$ and $(1-\zeta)^2(1-\alpha)<1-\alpha$. Therefore we always have $(\zeta+\alpha(1-\zeta))/(1-\zeta)^2(1-\alpha)>\alpha/(1-\alpha)$, implying that the constant bound of \satexlt~is always weaker than that of \satex. Numerical experiments also show that \satex~outperforms \satexlt~in terms of expected satisficing regret, for which we refer the readers to Section \ref{sec:numerical_tail} for details. 

To conclude, while \satexlt~is able to attain a light-tailed satisficing regret distribution, \satex~is able to attain a better constant satisficing regret bound compared to \satexlt, thus neither \satex~nor \satexlt~dominates the other. In applications that are sensitive to tail risk, it would be better to use \satexlt; on the other hand, if the decision maker values satisficing regret in expectation more, then \satex~is able to better serve the purpose of minimizing expected satisficing regret.}

Finally, we show that in the non-realizable case, the expected standard regret of \satexlt~is sub-linear in $T$.

\begin{theorem}\label{thm:regret_nonsat_tail}
If $r(A^*)<S$, the expected standard regret of \satexlt~is bounded by
\begin{equation*}
\mathbb{E}[\regret]\leq C_1T^{\alpha(1-\zeta)+\zeta}\cdot\textup{polylog}(T).
\end{equation*}
\end{theorem}

Proof of \cref{thm:regret_nonsat_tail} is provided in \cref{sec:thm:regret_nonsat_tail} of the appendix. \cref{thm:regret_nonsat_tail} states that for all $0<\zeta<1$, \satexlt~always attains a sub-linear in $T$ expected standard regret bound in the non-realizable case. When $\zeta$ approaches zero, the power of $T$ in the expected standard regret bound of \satexlt~approaches $\alpha$, which is the power of $T$ in the expected standard regret bound of \satex~and the learning oracle used in Step 1. In other words, when $\zeta$ is small, \satexlt~can have a similar expected standard regret as \satex.

\subsection{Novelty of the Results}\label{sec:novelty_tail}

In this section, we would like to highlight some major novelties of our algorithm \satexlt.

\vspace{2mm}
\noindent{\bf New Approach towards Light-Tailed Regret Distribution:} In \satexlt, a light-tailed satisficing regret distribution is attained mainly from a novel multi-round structure of the algorithm, where we gradually increase the number of time steps of running the learning oracle to search for satisficing arms. In \cref{prop:exit_prob_tail}, we prove that the number of rounds of \satexlt~is upper bounded by a shifted geometric distribution, which implies that the probability of starting a new round decays exponentially as the rounds proceed. With a polynomial growth in the length of running the oracle and an exponential decay in the probability of starting a new round, we are able to prove that the total number of time steps of running the oracle and forced sampling follows a light-tailed distribution. This design is novel in the literature of tail risk control in bandits. Our approach is also fundamentally different from existing ones in \cite{fan2024fragility} and \cite{simchi2024simple}, where a light-tailed regret distribution is obtained by using larger confidence intervals that hold with a probability of at least $1-\exp(-\Omega(T^\zeta))$.

\vspace{2mm}
\noindent{\bf Nuanced Difference between Satisficing Regret and Standard Regret:} \cite{simchi2024simple} show that for standard regret minimization in finite-armed bandits, any algorithm with a light-tailed standard regret distribution will inevitably incur a polynomial in $T$ instance-dependent expected standard regret. In other words, no algorithm can maintain both a light-tailed standard regret distribution and a near-optimal $O(\log(T))$ instance-dependent expected standard regret bound at the same time. Because the
constant expected satisficing regret bounds are typically instance-dependent, it becomes unclear if a constant expected satisficing regret is compatible with a light-tailed satisficing regret distribution. In this paper, we show that it is indeed possible to break this impossible regime. Specifically, \satexlt~ is able to attain a light-tailed satisficing regret distribution and a constant instance-dependent expected satisficing regret bound in the realizable case. Furthermore, our results can be extended beyond finite-armed bandits and are applicable to any class of bandit optimization problems that has a learning oracle with a sub-linear expected standard regret bound. This reveals a fundamental difference between the notions of standard regret and satisficing regret.

\vspace{2mm}
\noindent\textbf{No Additional Assumptions on the Learning Oracle:} We emphasize that our results do not rely on any condition on the learning oracle other than a sub-linear expected standard regret (Condition \ref{assump:learning_alg}). In particular, we do not require that the standard regret distribution of the learning oracle is light-tailed. Instead, \satexlt~is able to make use of learning oracles with a heavy-tailed standard regret distribution to attain a light-tailed satisficing regret distribution in the realizable case. For example, the satisficing regret distribution of \satexlt~is light-tailed when using the UCB algorithm or Thompson sampling as the learning oracle in Step 1, even though both algorithms are proven to have a heavy-tailed standard regret distribution (see \citealt{fan2024fragility}). The reason why the tail distribution of the learning oracle's standard regret does not change the result is that with the help of \cref{prop:exit_prob_tail}, we are able to directly prove that the total number of time steps of running the learning oracle and forced sampling follows a light-tailed distribution. Since we have assumed that the mean reward of each arm belongs to $[0,1]$, the total satisficing regret incurred when running the learning oracle and forced sampling follows a light-tailed distribution.

The ability of making use of learning oracles with potentially heavy-tailed standard regret distributions is important for \satexlt~to generalize to a broader range of bandit optimization problem classes. To the best of our knowledge, algorithms with light-tailed standard regret distribution have been established for only a very limited range of bandit optimization problems including finite-armed bandits and linear bandits. By accepting algorithms with potentially heavy-tailed standard regret distribution as learning oracle, \satexlt~is able to extend beyond this scope to any class of bandit optimization problems that admits a learning oracle with sub-linear expected standard regret bound.

\begin{remark}[\textbf{Light-Tailed Standard Regret in the Non-Realizable Case}]
We would like to clarify that although we have proven a light-tailed satisficing regret distribution for \satexlt~in the realizable case, \satexlt~may not be able to attain a light-tailed standard regret distribution in the non-realizable case. The main reason for this is that we make no assumption on the standard regret distribution of the learning oracle. Because in the non-realizable case, \satexlt~could run the learning oracle in $\Theta(T)$ time steps, it could have a heavy-tailed standard regret distribution in the non-realizable case if the standard regret distribution of the learning oracle is heavy-tailed. In \cref{sec:light-tail-KArm} of the appendix, we introduce an algorithm for finite-armed bandits that, besides achieving a constant expected satisficing regret, attains both a light-tailed satisficing regret distribution in the realizable case and a light-tailed standard regret distribution in the non-realizable case.
\end{remark}

\section{Numerical Experiments on Synthetic Data}\label{sec:numerical}

In this section, we conduct numerical experiments to test the performance of~\satex~on finite-armed bandits, concave bandits, and Lipschitz bandits. 

\subsection{Finite-Armed Bandits}\label{sec:numerical_KArm}

\textbf{Setup:} In this case, we consider an instance of $4$ arms, and the expected rewards of all arms are {$\{0.2,0.4,0.6,0.8\}$}. We vary the length of the time horizon from $500$ to {$7000$} with a stepsize of $500$. We conduct experiments for both the realizable case and the non-realizable case. For the realizable case, we set the satisficing level $S=0.7$; for the non-realizable case, we set the satisficing level $S=1.5$. In both cases, we compare \satex~against { the ``Phased Exploration with Bounded Regret" algorithm\footnote{Algorithm 1, Appendix C in \cite{GarivierM19}, with $\mu^*$ changed to $S$.} in \cite{GarivierM19} (hereafter referred to as Phased Exploration), STS algorithm in \cite{RussoV22} with adaptations to the time-undiscounted setting,} Thompson sampling in~\cite{AgrawalG17} as well as SAT-UCB and SAT-UCB+ in~\cite{Michel23}. { Note that in the setting of \cite{RussoV22}, a satisficing action (arm) $A$ is defined as a near-optimal one, \ie, $r(A)\geq r(A^*)-D$ with given $D>0$, and the regret of a single pull is defined as $r(A^*)-r(A)-D$ which can be negative, which is different from our definition on satisficing regret. Here, we run STS directly with our satisficing level $S=r(A^*)-D$. For our algorithm \satex, we use \texttt{SELECT-BoBW}, a variant of \satex~that attains a ``best of both worlds" constant expected satisficing regret bound for finite-arm bandits (we refer to Algorithm~\ref{alg:sat_KArm_bbw} in \cref{sec:best_of_both_world}~of the appendix for details), and use UCB algorithm \citep{BubeckC12} and Thompson sampling as the learning oracle used in Step 1. In order to improve empirical performance of our algorithm, we also make use of historical data from previous rounds in \satex~when updating posteriors of mean rewards of running Thompson sampling and computing upper / lower confidence bounds.} The experiment is repeated for {5000} times and we report the average satisficing regret in the realizable case and the average standard regret in the non-realizable case.

\begin{figure}[!ht]
    \centering
    \subfigure[Expected Satisficing Regret in the Realizable Case]{\includegraphics[height=4.5cm]{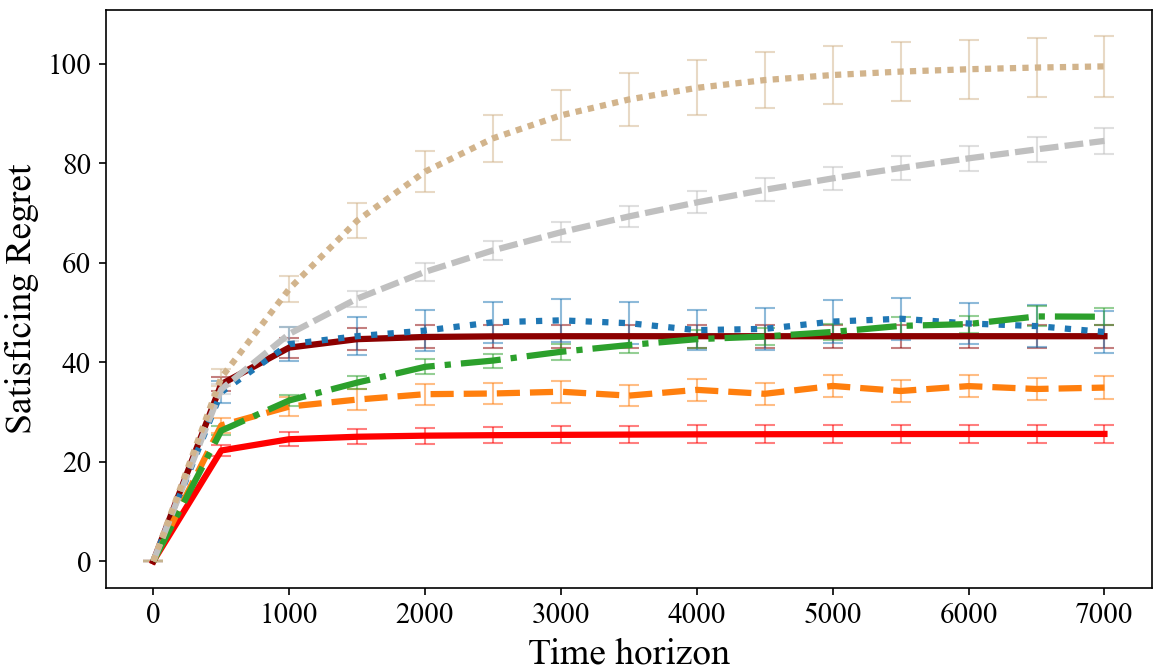}
    \label{fig:regret_k_arm_realizable}}
    \subfigure[Expected Standard Regret in the Non-Realizable Case]{\includegraphics[height=4.5cm]{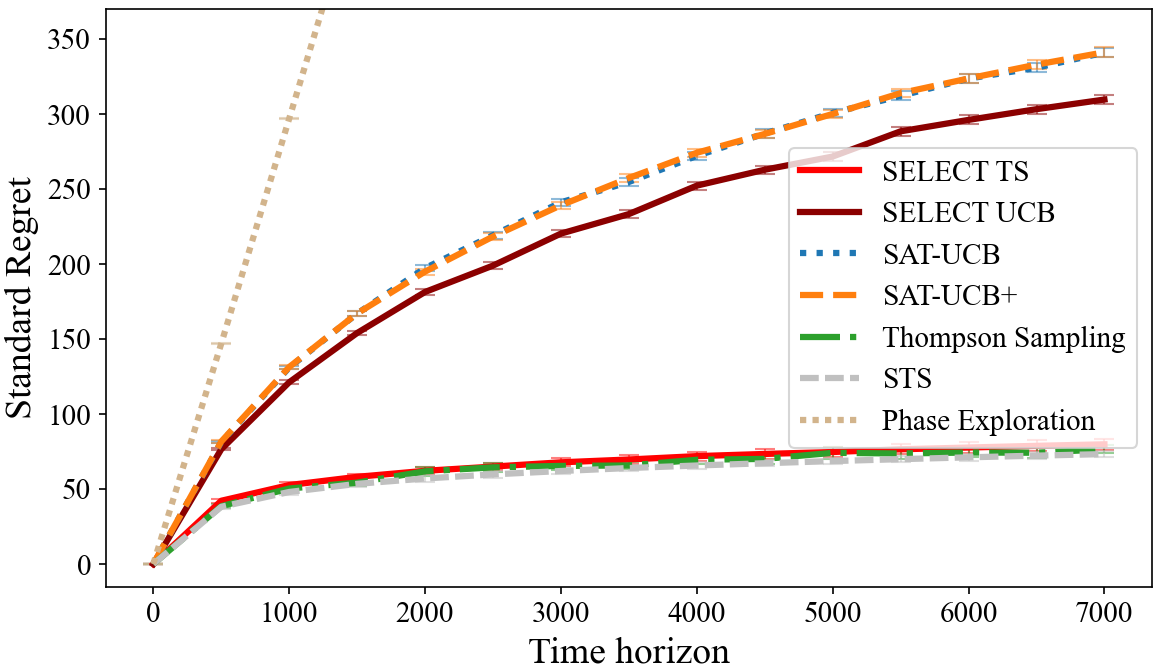}
    \label{fig:regret_k_arm_nonrealzable}}
    
    \caption{Comparison of {\texttt{SELECT-BoBW}}, Thompson sampling, STS, Phased Exploration, SAT-UCB and SAT-UCB+}
\end{figure}

\vspace{2mm}
\noindent\textbf{Results:} The results of the realizable case are presented in Figure~\ref{fig:regret_k_arm_realizable}, and the results of the non-realizable case are presented in Figure~\ref{fig:regret_k_arm_nonrealzable}. From Figure~\ref{fig:regret_k_arm_realizable}, one can see that the expected satisficing regret of {\satex~with both oracle algorithms, SAT-UCB, and SAT-UCB+ barely increases when $T$ exceeds $2500$, and the expected satisficing regret of Phased Exploration algorithm barely increases when $T$ exceeds $6000$. This suggests that \satex, SAT-UCB, SAT-UCB+ and Phased Exploration are able to attain a constant expected satisficing regret in the realizable case, consistent with the corresponding theoretical results. On the other hand, both STS and Thompson sampling fail to exhibit a constant expected satisficing regret in the realizable case. When comparing the expected satisficing regret of these algorithms, one can see that \satex~using Thompson sampling as oracle algorithm incurs the smallest expected satisficing regret among all tested algorithms, and the expected satisficing regret of \satex~using UCB as the oracle algorithm is comparable with that of SAT-UCB, slightly outperformed by the heuristic SAT-UCB+. Nevertheless, the expected satisficing regret of \satex~using UCB is still lower than Thompson sampling, STS, and the Phased Exploration. Although Phased Exploration algorithm eventually converges to a constant expected satisficing regret, the convergence to constant is slow and the algorithm incurs the highest expected satisficing regret among all tested algorithms.}

From Figure~\ref{fig:regret_k_arm_nonrealzable}~one can see that { the expected standard regret of~\satex~using Thompson sampling as oracle algorithm in the non-realizable case is comparable to that of Thompson sampling and STS. The expected standard regret of~\satex~using UCB as oracle algorithm in the non-realizable case is higher, but \satex~with both oracle algorithms incurs smaller expected standard regret than SAT-UCB and SAT-UCB+.} 
{ One can also see that Phased Exploration fails in the non-realizable setting, whose expected standard regret scales linearly as $T$ increases.}

\subsection{Concave Bandits}\label{sec:exp_concave}
\textbf{Setup:} In this case, we consider an instance with arm set $[0,1]$ and concave reward function $r(x)=1-16(x-0.25)^2$. We vary the length of the time horizon from $500$ to $5000$ with a stepsize of $500$. We consider both the realizable case and the non-realizable case. For the realizable case, we set the satisficing level $S=0.3$; for the non-realizable case, we set the satisficing level $S=1.5$. For both cases, we compare our algorithm \satex~with the convex bandit algorithm introduced in \cite{AgarwalFH11}. For \satex, we use the algorithm in \cite{AgarwalFH11} as the learning oracle. We also use SAT-UCB and SAT-UCB+ as heuristics by viewing the problem as Lipschitz bandits. That is, we first discretize the arm set uniformly with stepsize $L^{-{2}/3}T^{-1/3}$, where $L$ is the Lipschitz constant of the reward function, and then run SAT-UCB and SAT-UCB+ over the discretized arm set. { Note that for SAT-UCB and SAT-UCB+, we use a discretization stepsize that decreases as $T$ increases instead of a constant stepsize. The reason why we do not use a constant stepsize is that for any fixed constant discretization stepsize, there always exists a realizable instance such that none of the discretized arms is satisficing. This observation suggests that SAT-UCB and SAT-UCB+ with any constant discretization stepsize could incur linear in $T$ expected satisficing regret even when satisficing arms exist.} The experiment is repeated for 1000 times, and we report the average results.

\begin{figure}[!ht]
    \centering
    \subfigure[Expected Satisficing Regret in the Realizable Case]{\includegraphics[height=4.5cm]{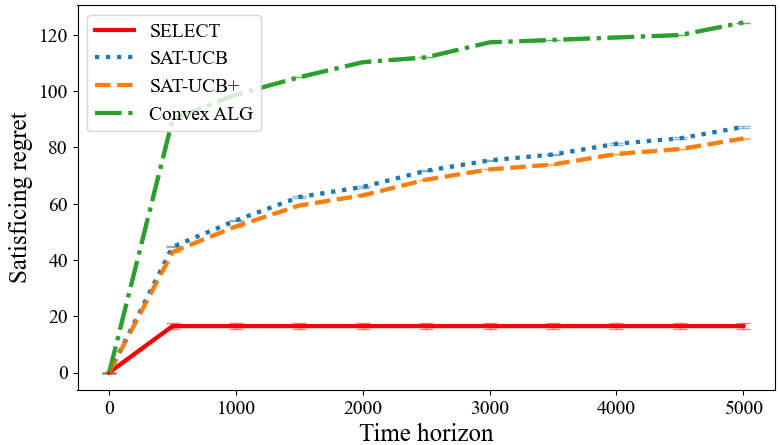}
    \label{fig:Satisficing_Regret_Concave_0}}
    \subfigure[Expected Standard Regret in the Non-Realizable Case]{\includegraphics[height=4.5cm]{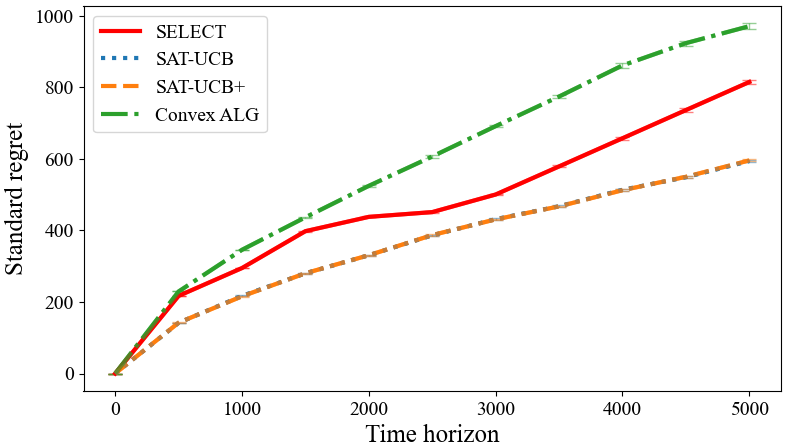}
    \label{fig:Non_Satisficing_Regret_Concave_0}}
    
    \caption{Comparison of~\satex~, Stochastic Convex, SAT-UCB and SAT-UCB+ for concave bandit}
\end{figure}

\vspace{2mm}
\noindent\textbf{Results:} The results of the realizable case are provided in Figure~\ref{fig:Satisficing_Regret_Concave_0}, and the results of the non-realizable case are provided in Figure~\ref{fig:Non_Satisficing_Regret_Concave_0}. One can see from Figure~\ref{fig:Satisficing_Regret_Concave_0} that the expected satisficing regret of \satex~becomes stable after $T$ exceeds $500$, suggesting that \satex~attains a constant expected satisficing regret in the realizable case. On the other hand, all the other three algorithms fail to exhibit a constant expected satisficing regret in the realizable case. When comparing the expected satisficing regret of the four algorithms, we observe that \satex~incurs a smaller expected satisficing regret than all three other algorithms.

One can see from \cref{fig:Non_Satisficing_Regret_Concave_0} that in the non-realizable case, \satex~incurs a smaller expected standard regret compared to the algorithm in \cite{AgarwalFH11}, { while performance of \satex~is slightly worse than both SAT-UCB and SAT-UCB+.}

In this experiment, the reason why \satex~outperforms the algorithm in \cite{AgarwalFH11} can be explained as follows. In the instance we use in this experiment, both the set of optimal arms and the set of significantly sub-optimal arms are large, and directly running the algorithm in \cite{AgarwalFH11} over the entire time horizon tends to over-explore due to larger confidence intervals, thus would potentially incur significant regret. In contrast, \satex~starts with running the algorithm in \cite{AgarwalFH11} over a small number of time steps $t_i$. This allows \satex~to have much smaller confidence intervals in early stages of the algorithm { due to a smaller log factor $\log(t_i)$ instead of $\log(T)$,} thus explores less and exploits more aggressively. Furthermore, when running forced sampling in Step 2, \satex~also exploits a reasonably good arm obtained from running the learning oracle, which means even more exploitation for \satex. Therefore, although theoretically \satex~may underperform \cite{AgarwalFH11} by a logarithmic factor of $T$, in the instance used in \cref{fig:Non_Satisficing_Regret_Concave_0}, \satex~outperforms the algorithm in \cite{AgarwalFH11}. Similar phenomena can also be observed in the numerical experiment for Lipschitz bandits in \cref{sec:experiment_lipschitz} { and for linear bandits in \cref{ssec:LinBandit_result}}.

\subsection{Lipschitz Bandits}\label{sec:experiment_lipschitz}
\textbf{Setup:} In this case, we consider a two-dimensional Lipschitz bandit in domain $(x,y)\in[0,1]^2$ with Lipschitz rewards $r(x,y) =\min\{1, 3e^{-100((x-0.5)^2 + (y-0.7)^2)}\}$. We vary the length of the time horizon from $500$ to $5000$ with a stepsize of $500$. We consider both the realizable case and the non-realizable case. For the realizable case, we set the satisficing level S = 0.5; for the non-realizable case, we set the satisficing level S = 1.5. For both cases, we compare our algorithm \satex~with the uniformly discretized UCB introduced in \cite{BubeckSY2011} (hereafter referred to as the ``Uniform UCB"). For~\satex, we use the Uniform UCB as the learning oracle. We again use SAT-UCB and SAT-UCB+ as heuristics by uniformly discretizing the arm set with stepsize $L^{-1/{2}}T^{-1/4}$, where $L$ is the Lipschitz constant of the reward function, and run the algorithms over the discretized arms. { Note that for SAT-UCB and SAT-UCB+, we use a discretization stepsize that decreases as $T$ increases instead of a constant stepsize. The reason for this is that, as we have mentioned in Section~\ref{sec:exp_concave}, SAT-UCB and SAT-UCB+ with constant stepsize could incur an expected satisficing regret that is linear in $T$.} The experiment is repeated for 1000 times and we report the average results.

\begin{figure}[!ht]
    \centering
    \subfigure[Expected Satisficing Regret in the Realizable Case]{\includegraphics[height=4.5cm]{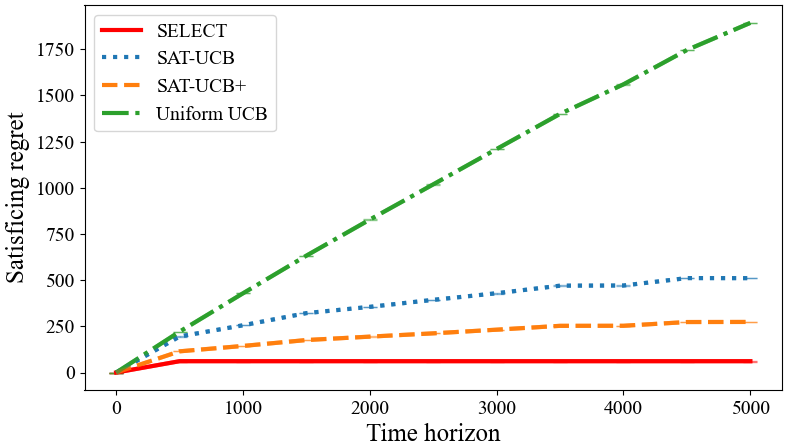}
    \label{fig:Satisficing_Regret_Lip}}
    \subfigure[Expected Standard Regret in the Non-Realizable Case]{\includegraphics[height=4.5cm]{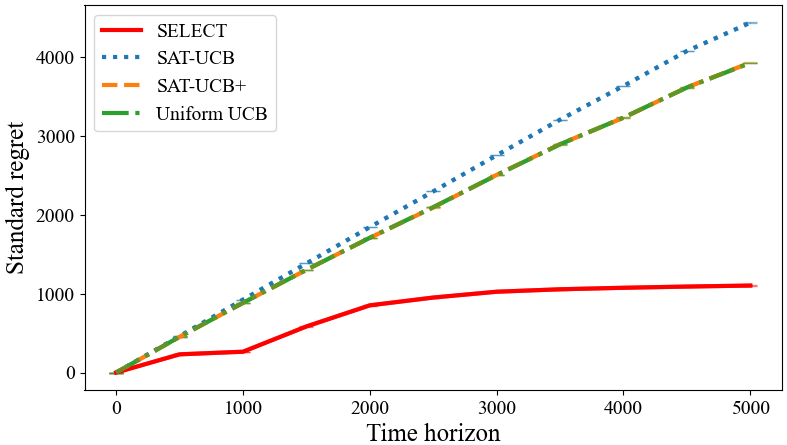}
    \label{fig:Non-Satisficing_Regret_Lip}}
    
    \caption{Comparison of~\satex, Uniform UCB, SAT-UCB and SAT-UCB+ for Lipschitz bandit}
\end{figure}

\vspace{2mm}
\noindent\textbf{Results:} The results of the realizable case are provided in Figure~\ref{fig:Satisficing_Regret_Lip}, and the results of the non-realizable case are provided in Figure~\ref{fig:Non-Satisficing_Regret_Lip}. One can see from Figure~\ref{fig:Satisficing_Regret_Lip} that the expected satisficing regret of \satex~becomes stable after $T$ exceeds $500$, suggesting that \satex~attains a constant expected satisficing regret in the realizable case. On the other hand, none of the other three algorithms are able to attain a constant expected satisficing regret. When comparing the expected satisficing regret of the four algorithms, one can observe that \satex~incurs an expected satisficing regret smaller than all three other algorithms.

As for the non-realizable case, one can see from \cref{fig:Non-Satisficing_Regret_Lip} that \satex~achieves the best empirical performance among the four algorithms, while the empirical performance of SAT-UCB and SAT-UCB+ is similar to that of Uniform UCB in the non-realizable case.

\subsection{Tail Risk Control in Finite-Armed Bandits}\label{sec:numerical_tail}
We test the performance of \satexlt~in terms of both expected satisficing regret and tail distribution of satisficing regret using an instance of finite-armed bandit.

\vspace{2mm}
\noindent\textbf{Setup:} We consider an instance of 4 arms. The expected rewards of all arms are set as $\{0.2,0.4,0.6,0.8\}$. For the realizable case, we set the satisficing level $S=0.7$; for the non-realizable case, we set the satisficing level $S=1.5$. For \satex~and \satexlt, we use Thompson sampling as the learning oracle. We vary $\zeta\in\{0.1,0.4\}$.

We first compare the expected satisficing regret of \satexlt~with that of \satex~and Thompson sampling. In this case, we vary the length of the time horizon from 1000 to 10000 with a stepsize of 1000. The experiment is repeated for 1000 times and we report the average satisficing regret. Then we compare the empirical tail distribution of satisficing regret between \satexlt~and~\satex~under time horizon $T=10000$. The experiment is repeated for 10000 times.

\begin{figure}[!ht]
    \centering
    \subfigure[\satex~Regret Distribution]{\includegraphics[height=4.8cm]{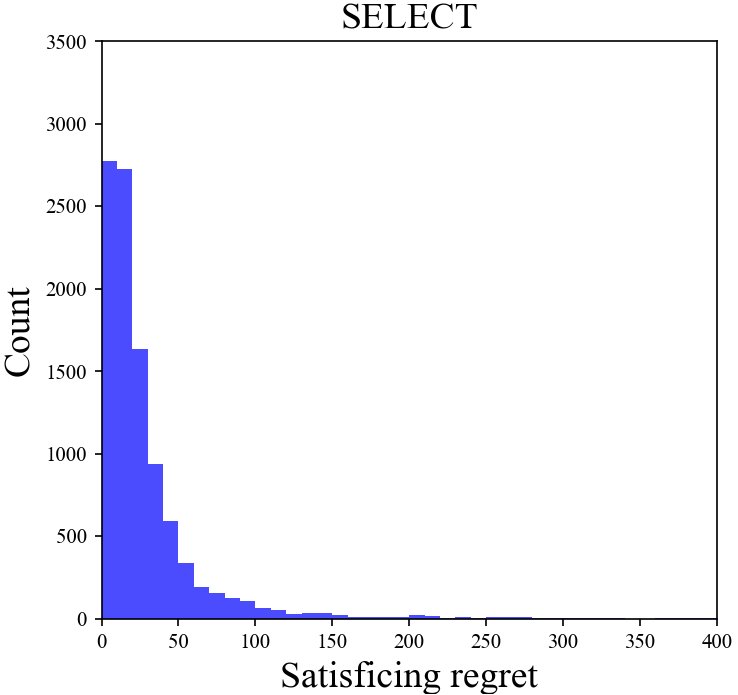}
    \label{fig:zeta=0.1-select}}
    \subfigure[\satexlt~Regret Distribution]{\includegraphics[height=4.8cm]{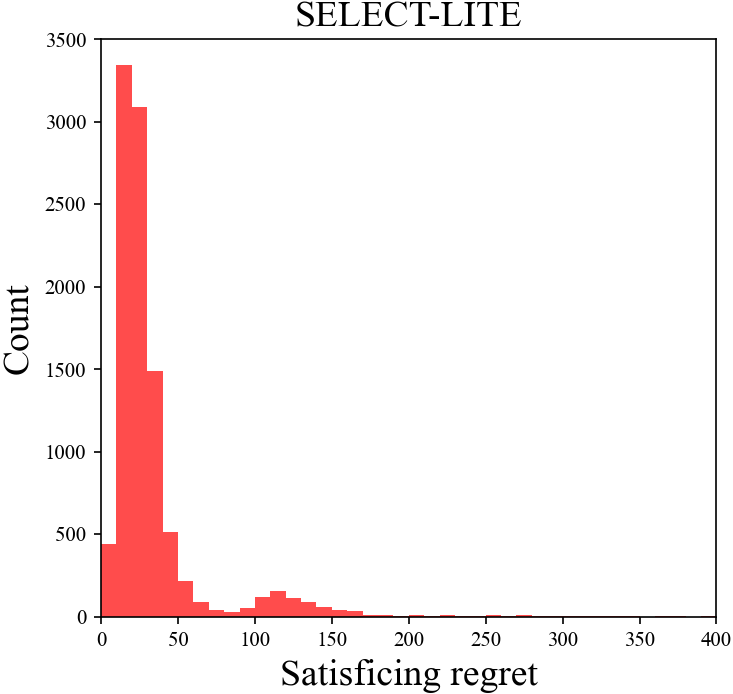}
    \label{fig:zeta=0.1-lite}}
    \subfigure[Tail Comparison]{\includegraphics[height=4.8cm]{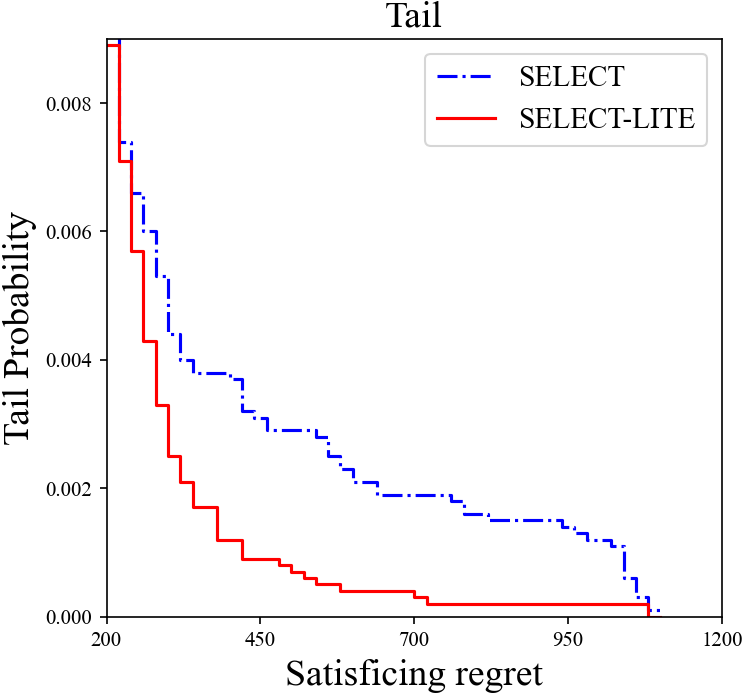}
    \label{fig:zeta=0.1-compare}}
    
    \caption{Realized satisficing regret distribution of \satex~ and \satexlt~ at $T=10000$ with $\zeta=0.1$}
    \label{fig:tail-zeta=0.1}
\end{figure}

\begin{figure}[!ht]
    \centering
    \subfigure[\satex~Regret Distribution]{\includegraphics[height=4.8cm]{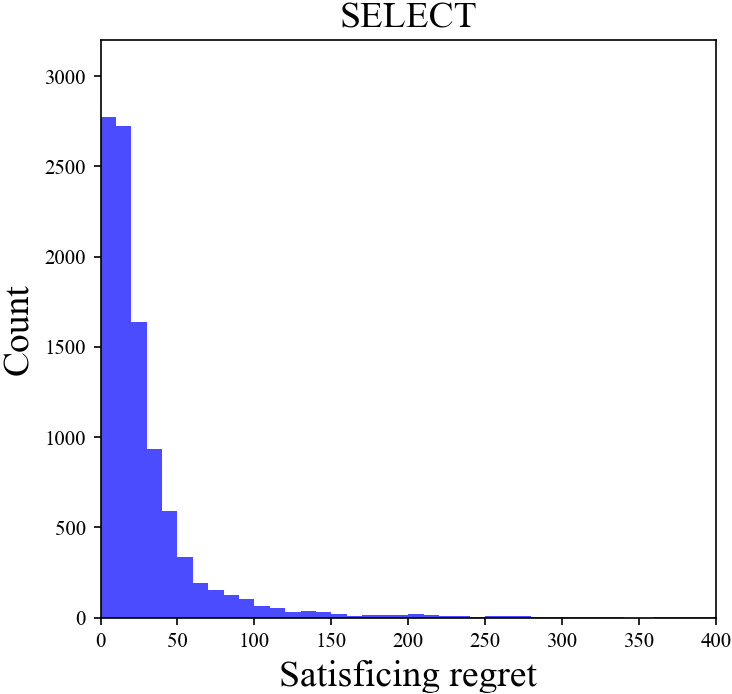}
    \label{fig:zeta=0.4-select}}
    \subfigure[\satexlt~Regret Distribution]{\includegraphics[height=4.8cm]{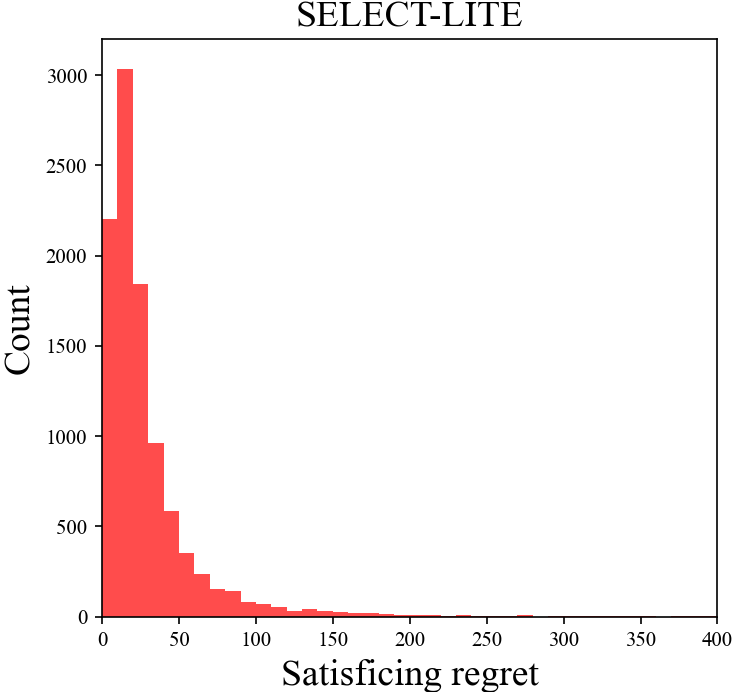}
    \label{fig:zeta=0.4-lite}}
    \subfigure[Tail Comparison]{\includegraphics[height=4.8cm]{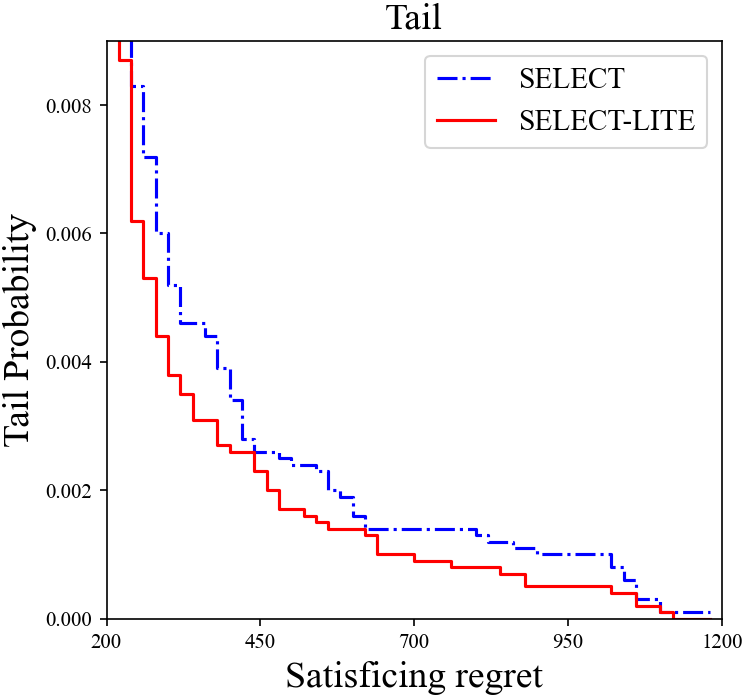}
    \label{fig:zeta=0.4-compare}}
    
    \caption{Realized satisficing regret distribution of \satex~ and \satexlt~ at $T=10000$ with $\zeta=0.4$}
    \label{fig:tail-zeta=0.4}
\end{figure}

\vspace{2mm}
\noindent\textbf{Results:} The results of realized satisficing regret distribution under $\zeta=0.1$ are presented in Figure~\ref{fig:zeta=0.1-select} for \satex~and Figure~\ref{fig:zeta=0.1-lite} for \satexlt~, and the comparison of tail probabilities is presented in Figure~\ref{fig:zeta=0.1-compare} { by zooming in on the tail region of the distributions}. The results under $\zeta=0.4$ are presented in Figure~\ref{fig:tail-zeta=0.4}. Here we report the empirical distribution of realized satisficing regret using histograms with a bin width of 10. From Figure~\ref{fig:zeta=0.1-compare} and \ref{fig:zeta=0.4-compare}, one can see that under both $\zeta=0.1$ and $\zeta=0.4$, \satexlt~exhibits a lower tail probability compared to \satex. When comparing between the case of $\zeta=0.1$ and the case of $\zeta=0.4$, \satexlt~with $\zeta=0.1$ exhibits a noticeably thinner tail, even though a larger $\zeta$ should theoretically yield the faster decay rate. This is because when $\zeta=0.1$, $t_i'$ increases rapidly in the first few rounds, thus \satexlt~can quickly find the satisficing arm within a few rounds. Consequently, the empirical tail of \satexlt~when $\zeta=0.4$ is a summation of tails of many rounds, but the one when $\zeta=0.1$ is a summation of simply one or two rounds, which potentially makes the tail probability smaller. 

\begin{figure}[!ht]
    \centering
    \subfigure[Expected Satisficing Regret in the Realizable Case]{\includegraphics[height=4.5cm]{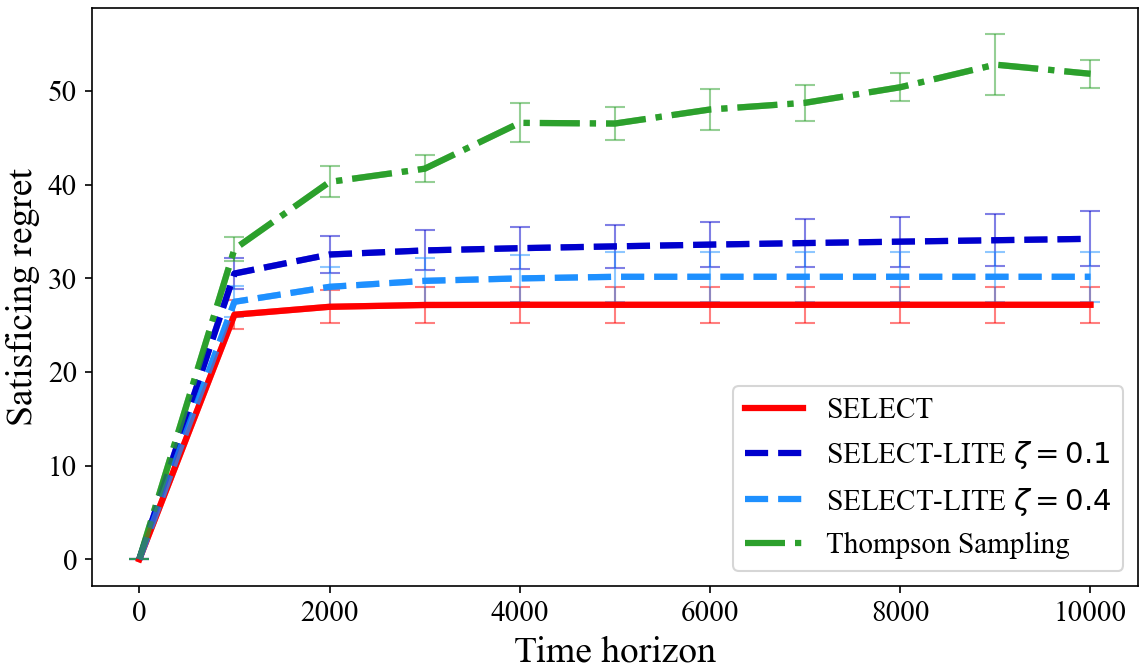}
    \label{fig:Satisficing_Regret_lite}}
    \subfigure[Expected Standard Regret in the Non-Realizable Case]{\includegraphics[height=4.5cm]{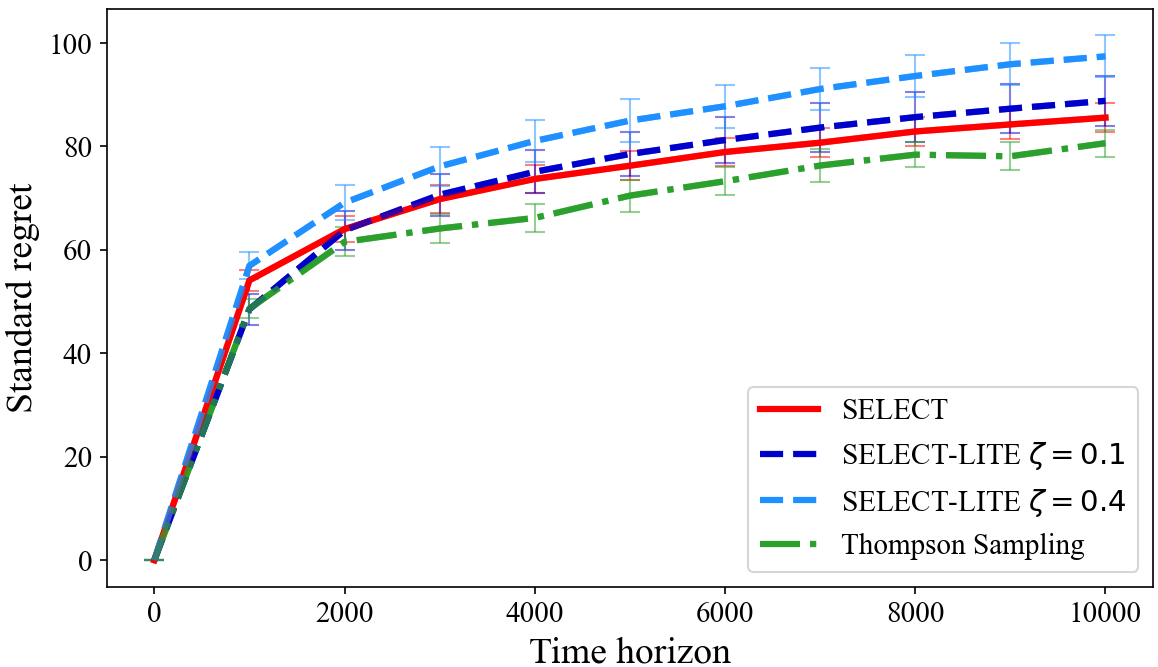}
    \label{fig:Non-Satisficing_Regret_lite}}
    
    \caption{{ Comparison of \satex~, \satexlt~ with different $\zeta$, and Thompson sampling}}
\end{figure}

The results of expected satisficing regret in the realizable case are presented in Figure~\ref{fig:Satisficing_Regret_lite}. The results of expected standard regret in the { non-realizable} case are presented in Figure~\ref{fig:Non-Satisficing_Regret_lite}. From Figure~\ref{fig:Satisficing_Regret_lite} one can observe that \satexlt~ exhibits a constant expected satisficing regret that is only slightly larger than that of \satex. Furthermore, both when $\zeta=0.1$ and when $\zeta=0.4$, \satexlt~incurs a smaller expected satisficing regret compared to Thompson sampling. From Figure~\ref{fig:Non-Satisficing_Regret_lite} one can see that \satexlt~ with different $\zeta$ incurs roughly the same expected standard regret as \satex~and is slightly { worse} than Thompson sampling. This is because each \satexlt~algorithm shares a backbone with \satex, and \satexlt~with $\zeta=0.1$ has an $O(T^{0.55})$ regret bound that is very close to that of \satex. For \satexlt~with $\zeta=0.4$, the theoretical expected standard regret bound is $\tilde{O}(T^{0.7})$, which is worse than the $\tilde{O}(\sqrt{T})$ guarantee for Thompson sampling. With these observations, we conclude that by running \satexlt~instead of \satex, the decision maker can have better protection against tail risks, both theoretically and empirically, without incurring a significantly larger satisficing regret in expectation.

\section{Numerical Experiments on Real Data: Satisficing in Dynamic Pricing}\label{sec:numerical_real_data}

To further evaluate the empirical performance of our algorithm in more practical settings, in this section, we conduct numerical experiments on an instance calibrated from the avocado dataset \cite{kiggins2018avocado}.

\subsection{Data Description, Pre-Processing, and Setup}
We use the avocado dataset \cite{kiggins2018avocado} in the New York region. The data records weekly average prices of organic avocados and the corresponding weekly total selling volume from January 4, 2015, to December 27, 2015, for a total of 52 weeks (with outliers deleted). The weekly average prices in these 52 data points range from \$1.65 to \$2.34, and the average price is \$2.03. The weekly selling volumes range from 9205 to 25675, and the average weekly selling volume is 17106.

\begin{figure}[!ht]
    \centering
    \includegraphics[width=0.5\linewidth]{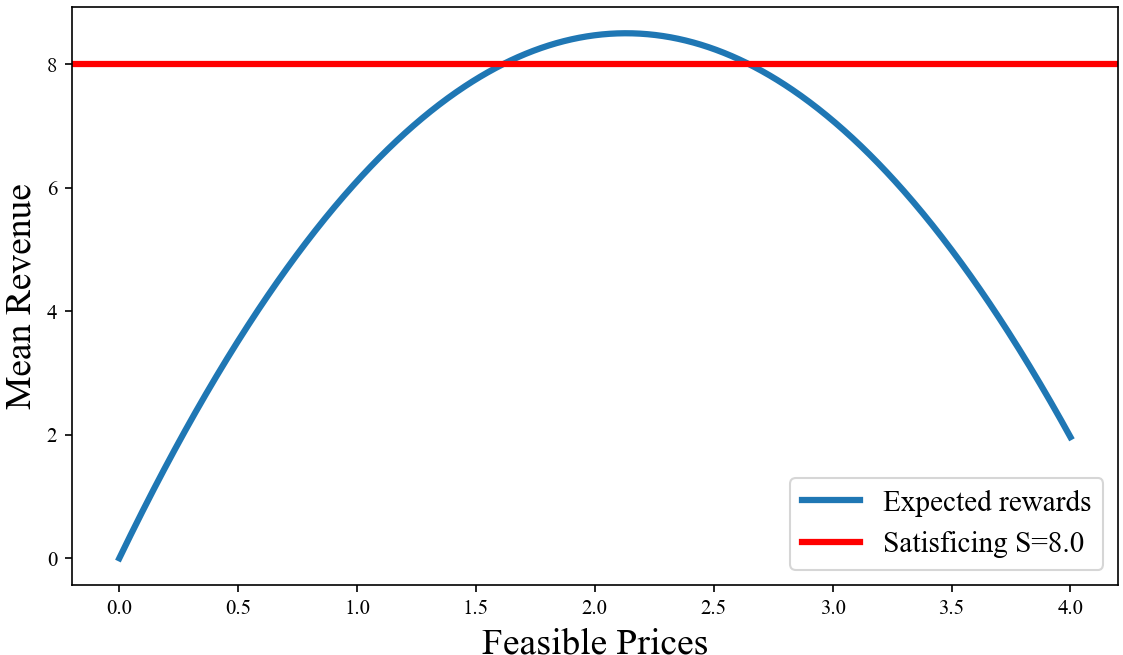}
    \caption{Expected revenue of the feasible prices $p\in[0,4]$}
    \label{fig:real-data-shape}
\end{figure}

We consider a dynamic pricing problem with linear demands to the prices, and test the performance of the algorithms on this model fitted by the avocado dataset. Assuming that the demand is equal to $D_t=g-hp_t+\epsilon_t$, where $\epsilon_t\sim N(0,\sigma^2)$, given price $p_t$ at each time step $t$, we fit the linear regression model from the real data to be $\hat{g}=32724,\;\hat{h}=7678,\;\hat{\sigma}=4100.$ By normalizing the model to ensure unit noise variance, we obtain the coefficients to be 
$$
\hat{g}_n=32724/4100,\;\hat{h}_n=7678/4100,\;\hat{\sigma}_n=1.
$$
We also extend the range of average price to be $p\in[0,4]$. The mean revenue received at each time $t$ is $p_t(\hat{g}_n-\hat{h}_np_t)$, and the mean revenue of each feasible price is visualized in Figure~\ref{fig:real-data-shape}.

We set the length of the time horizon to be 500 to 8000 for the realizable case and 5000 for the non-realizable case with a stepsize of 500. For the realizable case, we set the satisficing level $S = 8.0$; for the non-realizable case, we set the satisficing level $S = 10.0$. For both cases, we compare our algorithm \satex~with LinUCB introduced in \cite{LS20}. For \satex, we use LinUCB in \cite{LS20} as the learning oracle. We also use SAT-UCB and SAT-UCB+ as heuristics by viewing the problem as Lipschitz bandits. That is, we first discretize the arm set uniformly with a stepsize $L^{-1/3}T^{-1/3}$, where $L$ is the Lipschitz constant of the reward function, and then run SAT-UCB and SAT-UCB+ over the discretized arm set. The experiment is repeated 1000 times and we report the averaged results.

\subsection{Results}\label{ssec:LinBandit_result}
\begin{figure}[!ht]
    \centering
    \subfigure[Expected Satisficing Regret in the Realizable Case]{\includegraphics[height=4.5cm]{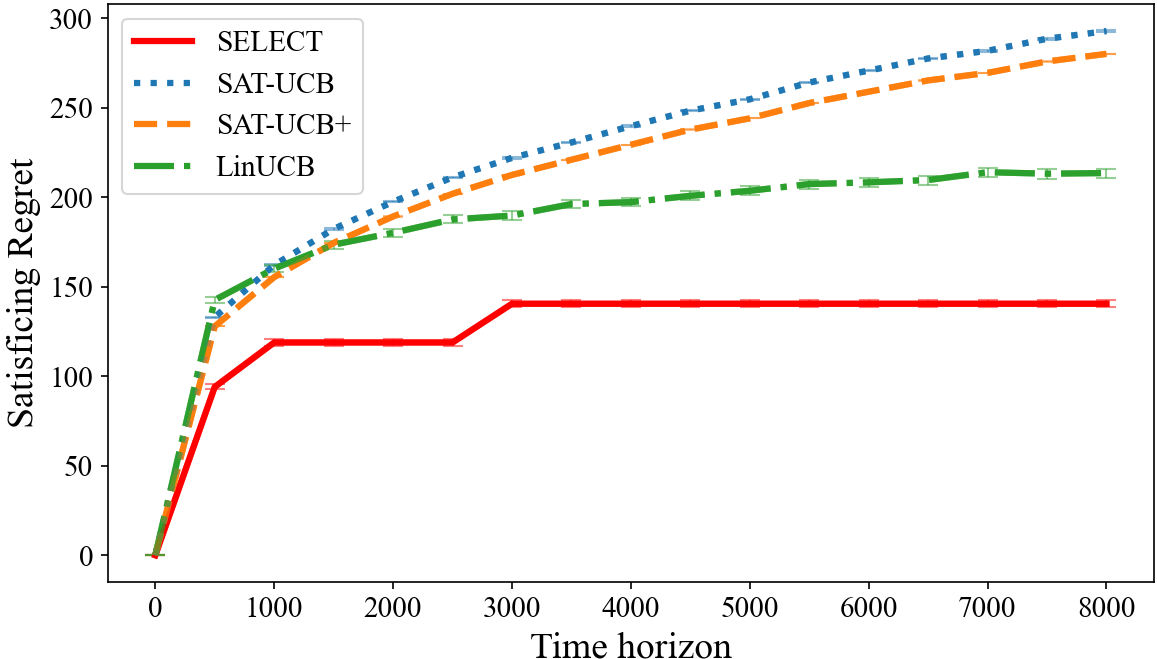}
    \label{fig:linear-sat}}
    \subfigure[Expected Standard Regret in the Non-Realizable Case]{\includegraphics[height=4.5cm]{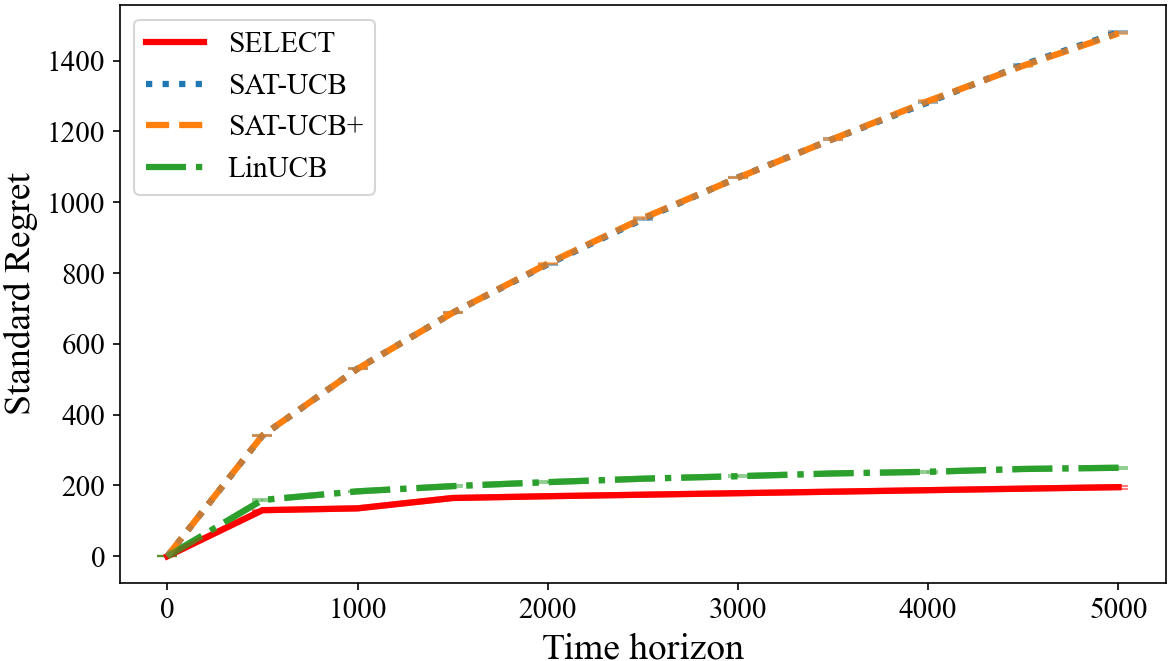}
    \label{fig:linear-non}}
    
    \caption{Comparison of~\satex~, LinUCB, SAT-UCB and SAT-UCB+ for linear bandit}
\end{figure}
The results of the realizable case are provided in Figure~\ref{fig:linear-sat}, and the results of the non-realizable case are provided in Figure~\ref{fig:linear-non}. From Figure~\ref{fig:linear-sat}, one can observe that the expected satisficing regret of \satex~becomes stable after $T$ exceeds $3000$. This suggests that in the realizable case, \satex~is able to attain constant expected satisficing regret. On the other hand, none of the other three algorithms are able to achieve constant expected satisficing regret in the realizable case. When comparing the expected satisficing regret of the four algorithms, one can see that \satex~incurs a smaller expected satisficing regret than all the other three algorithms.

From~\cref{fig:linear-non}, one can see that in the non-realizable case, \satex~slightly outperforms LinUCB and significantly outperforms both SAT-UCB and SAT-UCB+. The reason why \satex~and LinUCB significantly outperform SAT-UCB and SAT-UCB+ is that in the problem, SAT-UCB and SAT-UCB+ view the problem as a Lipschitz bandit, without making use of the linear structure of demand. { The reason why \satex~outperforms LinUCB in terms of expected standard regret is similar to the case in concave bandits in \cref{sec:exp_concave}, for which we refer to the last paragraph of \cref{sec:exp_concave} for a detailed discussion.}

\section{Conclusion}

In this paper, we propose \satex, a general algorithmic template, for satisficing regret minimization in bandit optimization. For a given class of bandit optimization problems and a sub-linear regret learning oracle, we show that \satex~attains constant expected satisficing regret in the realizable case and the same regret as the learning oracle in the non-realizable case. We instantiate \satex~to finite-armed bandits, concave bandits, and Lipschitz bandits, and also present matching lower bounds for all three bandit problem classes. We also consider tail risk control in satisficing regret minimization, where we introduce~\satexlt, which attains a light-tailed satisficing regret distribution while still maintaining a constant expected satisficing regret. Finally, we conduct numerical experiments to validate the performance of \satex~and \satexlt~on both synthetic datasets and a real-world dynamic pricing case study.

For future studies, we suggest further exploration in the following directions. First, our work mainly focuses on the stationary settings, and further research could explore satisficing in the nonstationary online learning settings \citep{wei2021non-stationary,cheung2022hedging}. Furthermore, the notion of satisficing regret indeed discourages exploration. Therefore, another direction is to study the role of satisficing regret in settings with limited adaptivity \citep{gao2019batched,feng2023fairness}.

\bibliographystyle{ormsv080}
\bibliography{ref}
\OneAndAHalfSpacedXI
\newpage

\begin{appendices}{\Large \noindent\textbf{Appendix}}

\section{Proof of Theorem~\ref{thm:regret_bound_1}}\label{sec:thm:regret_bound_1}

Denote $\beta'=\max\{\beta,1/2\}$ and $K_{\alpha,\beta}=\dfrac{32}{(1-\alpha)^\beta}$, and $i_0$ as
\begin{equation*}
i_0=\max\{i\in\mathbb{Z}_+: C_1K_{\alpha,\beta}\gamma_i\log({4}/\gamma_i)^{\max\{\beta,1/2\}}>\Delta_S^*\}.
\end{equation*}
We first prove an upper bound on $i_0$. Denote $\delta_0\in(0,1)$ the smallest solution to the following equation
\begin{equation}
\delta_0\log({4}/\delta_0)^{\beta'}=\dfrac{\Delta_S^*}{C_1K_{\alpha,\beta}}.
\label{eq:delta_Delta_equation}
\end{equation}
{ Note that $\delta_0\log({4}/\delta_0)^{\beta'}$ is continuous on $(0,1)$. When $\delta_0\to0$, $\delta_0\log({4}/\delta_0)^{\beta'}\to 0$, and its maximum is $\max_{\delta_0}\delta_0\log({4}/\delta_0)^{\beta'}\geq\delta_0\log({4}/\delta_0)^{\beta'}\big|_{\delta_0=1^-}>1$. On the other hand, since by definition $K_{\alpha,\beta}=32/(1-\alpha)^\beta\geq 32$, the right hand side is upper bounded by $\Delta_S^*(1-\alpha)^\beta/(32C_1)\leq\Delta_S^*/32\leq1/32$ with $\Delta_S^*\leq1$ and $1/2\leq\alpha<1$. Therefore there always exists $\delta_0\in(0,1)$ such that \eqref{eq:delta_Delta_equation} holds. Based on \eqref{eq:delta_Delta_equation}, we also have}
\begin{equation*}
\begin{aligned}
&\dfrac{1}{\delta_0}=\dfrac{C_1K_{\alpha,\beta}\log({ 4}/\delta_0)^{\beta'}}{\Delta_S^*}\\
=&{ \dfrac{C_1K_{\alpha,\beta}\cdot(4\beta')^{\beta'}\log((4/\delta_0)^{1/(4\beta')})^{\beta'}}{\Delta_S^*}}\\
\leq& { \dfrac{C_1K_{\alpha,\beta}\cdot(4\beta')^{\beta'}(4/\delta_0)^{1/4}}{\Delta_S^*}}\\
=& { \dfrac{C_1K_{\alpha,\beta}\cdot(4\beta')^{\beta'}2^{1/2}}{\Delta_S^*(\delta_0)^{1/4}}}\\
\leq& { \dfrac{(8\beta')^{\beta'}C_1K_{\alpha,\beta}}{\Delta_S^*(\delta_0)^{1/4}}}.
\end{aligned}
\end{equation*}
{ Here, the first inequality is because $\log(x)<x$ for all $x>0$, and the second inequality is because $\beta'=\max\{\beta,1/2\}\geq1/2$, thus $2^{1/2}\leq 2^{\beta'}$. Therefore we get}
\begin{equation*}
\dfrac{1}{\delta_0}\leq \left(\dfrac{(8\beta')^{\beta'}C_1K_{\alpha,\beta}}{\Delta_S^*}\right)^{4/3},
\end{equation*}
or equivalently,
\begin{equation}
\dfrac{4}{\delta_0}\leq {4}\left(\dfrac{(8\beta')^{\beta'}C_1K_{\alpha,\beta}}{\Delta_S^*}\right)^{4/3}\leq \left(\dfrac{4(8\beta')^{\beta'}C_1K_{\alpha,\beta}}{\Delta_S^*}\right)^{4/3}.
\label{eq:1/delta0_upperbound}
\end{equation}
{ Plugging \eqref{eq:1/delta0_upperbound} into the $\log({4}/\delta_0)$ term in \eqref{eq:delta_Delta_equation}, we have}
\begin{equation*}
\dfrac{1}{\delta_0} = \frac{C_1K_{\alpha,\beta}\log({4}/\delta_0)^{\beta'}}{\Delta_S^*} \leq {\dfrac{(4/3)^{\beta'} C_1K_{\alpha,\beta}}{\Delta_S^*}\cdot\log\left(\frac{4(8\beta')^{\beta'}C_1K_{\alpha,\beta}}{\Delta_S^*}\right)^{\beta'}}.
\end{equation*}
We define $D_{\alpha,\beta}={4}(8\beta')^{\beta'}K_{\alpha,\beta}$, $L_{\alpha,\beta}=(4/3)^{\beta'}K_{\alpha,\beta}$, then we have
\begin{equation*}
\delta_0\geq \dfrac{\Delta_S^*}{L_{\alpha,\beta}C_1\log(D_{\alpha,\beta}C_1/\Delta_S^*)^{\beta'}}.
\end{equation*}
By definition of $i_0$ we have that $i_0$ is the largest integer that satisfies $\gamma_{i_0}\geq \delta_0$. Therefore we get
\begin{equation}\label{eq:ub_gamma_i_0}
\gamma_{i_0}\geq \dfrac{\Delta_S^*}{L_{\alpha,\beta}C_1\log(D_{\alpha,\beta}C_1/\Delta_S^*)^{\beta'}}.
\end{equation}
Recall that $\gamma_i=2^{-i(1-\alpha)/\alpha}$, \eqref{eq:ub_gamma_i_0} implies that
\begin{equation*}
i_0\leq \dfrac{\alpha}{1-\alpha}\log_2\left(\dfrac{L_{\alpha,\beta}C_1\log(D_{\alpha,\beta}C_1/\Delta_S^*)^{\beta'}}{\Delta_S^*}\right),
\end{equation*}
or equivalently
\begin{equation}
2^{i_0}\leq \left(\dfrac{L_{\alpha,\beta}C_1\log(D_{\alpha,\beta}C_1/\Delta_S^*)^{\beta'}}{\Delta_S^*}\right)^{\frac{\alpha}{1-\alpha}}.
\label{eq:2^i0_upperbound}
\end{equation}

{ We divide the total satisficing regret into regret from rounds $1$ to $i_0$ and after round $i_0$.} Using the result of Proposition~\ref{prop:regret_bound_round_1}~we have that the satisficing regret by round $i_0$ is bounded by
\begin{equation}
\begin{aligned}
&\sum_{i=1}^{i_0}\dfrac{{8}C_1}{(1-\alpha)^\beta}\gamma_i^{-\alpha/(1-\alpha)}\log({4}/\gamma_i)^{\beta}\\
\leq&\sum_{i=1}^{i_0}\dfrac{{8}C_1}{(1-\alpha)^\beta}\gamma_i^{-\alpha/(1-\alpha)}\log({4}/\delta_0)^{\beta}\\
\leq&\sum_{i=1}^{i_0}\dfrac{{8}C_1}{(1-\alpha)^\beta}2^{i}\log({4}/\delta_0)^\beta\\
\leq&{ \dfrac{32C_1}{(1-\alpha)^\beta}}2^{i_0}\log({4}/\delta_0)^\beta\\
\leq &{ \dfrac{32(4/3)^\beta C_1}{(1-\alpha)^\beta}\left(\dfrac{L_{\alpha,\beta}C_1\log(D_{\alpha,\beta}C_1/\Delta_S^*)^{\beta'}}{\Delta_S^*}\right)^{\alpha/(1-\alpha)}\log\left(\dfrac{D_{\alpha,\beta}C_1}{\Delta_S^*}\right)^\beta}\\
=&\dfrac{{ 32(4/3)^\beta}L_{\alpha,\beta}^{\alpha/(1-\alpha)}C_1^{1/(1-\alpha)}}{(1-\alpha)^\beta}\left(\dfrac{1}{\Delta_S^*}\right)^{\alpha/(1-\alpha)}\log\left(\dfrac{D_{\alpha,\beta}C_1}{\Delta_S^*}\right)^{\alpha\beta'/(1-\alpha)+{\beta}}.
\label{eq:sat_regret_before_i0}
\end{aligned}
\end{equation}
{ Here the first inequality is because by definition of $\delta_0$ and $i_0$, we have $\gamma_i=2^{-i(1-\alpha)/\alpha}\geq \gamma_{i_0}\geq \delta_0$ for all $i\leq i_0$, the second inequality is by plugging into the definition of $\gamma_i=2^{-i(1-\alpha)/\alpha}$, the third inequality is due to $\sum_{i=1}^{i_0}2^i\leq 2^{i_0+1}\leq 4\cdot 2^{i_0}$, and the fourth inequality is by plugging into \eqref{eq:1/delta0_upperbound} on $\log({4}/\delta_0)$ and \eqref{eq:2^i0_upperbound} on $2^{i_0}$.}

By the results of Proposition~\ref{prop:exit_prob_1}, for every $k\geq 1$, the probability that round $i_0+k$ starts within the time horizon is at most $4^{-k+1}$. Therefore the regret incurred after round $i_0$ is bounded by
\begin{equation}
\begin{aligned}
&\sum_{k=1}^\infty \dfrac{{8}C_1}{(1-\alpha)^\beta}2^{i_0+k}\log({4}/\gamma_{i_0+k})^\beta\cdot 4^{-k+1}\\
\leq & { \sum_{k=1}^\infty \dfrac{32C_1}{(1-\alpha)^\beta}2^{i_0}\left(\log(4/\gamma_{i_0}) + \frac{1-\alpha}{\alpha}k\right)^\beta\cdot 2^{-k}}\\
\leq&{ \sum_{k=1}^\infty \dfrac{32\cdot 2^\beta C_1}{(1-\alpha)^\beta}2^{i_0}\log(4/\gamma_{i_0})^\beta \cdot  k^\beta\cdot 2^{-k} }\\
\leq &\dfrac{{ 32\cdot 2^\beta}C_1}{(1-\alpha)^\beta}2^{i_0}\log({4}/\delta_0)^\beta\cdot\sum_{k=1}^\infty k^{\beta}\cdot 2^{-k}\\
\leq &\dfrac{{ 32(8/3)^\beta}L_{\alpha,\beta}^{\alpha/(1-\alpha)}C_1^{1/(1-\alpha)}M_\beta}{(1-\alpha)^\beta}\left(\dfrac{1}{\Delta_S^*}\right)^{\alpha/(1-\alpha)}\log\left(\dfrac{D_{\alpha,\beta}C_1}{\Delta_S^*}\right)^{\alpha\beta'/(1-\alpha)+{\beta}},
\label{eq:sat_regret_after_i0}
\end{aligned}
\end{equation}
where $M_\beta=\sum_{k=1}^\infty k^\beta\cdot 2^{-k}$ is a constant only dependent on $\beta$. { The first inequality follows directly from the definition of $\gamma_i=2^{-i(1-\alpha)/\alpha}=\gamma_{i_0}\cdot2^{-k(1-\alpha)/\alpha}$ with $i=i_0+k$. The second inequality is because $\log(4/\gamma_{i_0})>\log(4)>1$ and $k\cdot(1-\alpha)/\alpha\leq k$ for $1/2\leq\alpha<1$, and for any $a,b\geq 1$, we have $2ab\geq ab+1\geq a+b$. The third inequality is because by definition $\delta_0\leq \gamma_{i_0}$. The fourth inequality is due to \eqref{eq:1/delta0_upperbound} for $\log({4}/\delta_0)$ and \eqref{eq:2^i0_upperbound} for $2^{i_0}$. Combining the satisficing regret before round $i_0$ \eqref{eq:sat_regret_before_i0} and after round $i_0$ \eqref{eq:sat_regret_after_i0},} the total satisficing regret is bounded by
\begin{equation}
\mathbb{E}[\texttt{Regret}_S]\leq\dfrac{{ 32(8/3)^\beta }L_{\alpha,\beta}^{\alpha/(1-\alpha)}C_1^{1/(1-\alpha)}(M_\beta+{2^{-\beta}})}{(1-\alpha)^\beta}\left(\dfrac{1}{\Delta_S^*}\right)^{\alpha/(1-\alpha)}\log\left(\dfrac{D_{\alpha,\beta}C_1}{\Delta_S^*}\right)^{\alpha\beta'/(1-\alpha)+{ \beta}}.
\label{eq:main_bound_const}
\end{equation}

Furthermore, since the length of the time horizon is $T$ time steps, we have $t_i\leq T$. Since by definition of $t_i=\lceil2^{i/\alpha}\rceil$, { the number of rounds that end within the time horizon is bounded by $2^{i/\alpha}\leq T$, \ie, $i\leq\lfloor\alpha\log_2(T) \rfloor$ because $i$ is an integer. Then,} the maximum number of rounds is bounded by $i_{\max}=\lceil\alpha\log_2(T)\rceil{\leq\alpha\log_2(T)+1}$. { Recall $\gamma_i=2^{-i(1-\alpha)/\alpha}$, for any $i\leq i_{\max}$, we have
\begin{equation}
\begin{aligned}
\gamma_i&\geq \gamma_{i_{\max}}=2^{-(1-\alpha)\lceil\alpha\log_2(T)\rceil/\alpha} \\
&\geq 2^{-(1-\alpha)\log_2(T)-(1-\alpha)/\alpha}\\
&\geq 2^{-\log_2(T)-1}=1/(2T),
\label{eq:gammai_1/T}
\end{aligned}
\end{equation}
where the third inequality is because $(1-\alpha)/\alpha\leq1$ for $1/2\leq\alpha<1$.} Therefore, we have that the satisficing regret is bounded by
\begin{equation}
\begin{aligned}
\mathbb{E}[\texttt{Regret}_S]\leq&\sum_{i=1}^{i_{\max}}\dfrac{{8}C_1}{(1-\alpha)^\beta}\gamma_i^{-\alpha/(1-\alpha)}\log({4}/\gamma_i)^\beta\\
\leq &\sum_{i=1}^{i_{\max}} \dfrac{{8}C_1}{(1-\alpha)^\beta}2^{i+1}\log(8T)^\beta \\
\leq& { \dfrac{32C_1}{(1-\alpha)^\beta}2^{i_{\max}}\log(8T)^\beta} \\
\leq& {  \dfrac{32C_1}{(1-\alpha)^\beta}2^{\alpha\log_2(T)+1}\log(8T)^\beta}\\
=& \dfrac{{64}C_1}{(1-\alpha)^\beta}T^\alpha\log(8T)^\beta,
\label{eq:main_bound_sublinear}
\end{aligned}
\end{equation}
{ where the first inequality is by the results of Proposition~\ref{prop:regret_bound_round_1}, the second inequality is by $4/\gamma_i\leq8T$ \eqref{eq:gammai_1/T} for all $i\leq i_{\max}$ and the definition $\gamma_i=2^{-i(1-\alpha)/\alpha}$, the third inequality is by $\sum_{i=1}^{i_{\max}}2^{i+1}\leq2\cdot2^{i_{\max}+1}=4\cdot2^{i_{\max}}$, and the fourth inequality is by $i_{\max}=\lceil\alpha\log_2(T)\rceil\leq\alpha\log_2(T)+1$.}

Combining the two parts of results, {\ie, the constant satisficing regret \eqref{eq:main_bound_const} and the sub-linear satisficing regret \eqref{eq:main_bound_sublinear},} we have completed the proof of Theorem~\ref{thm:regret_bound_1} { by putting all constant terms into the $\text{polylog}$ term}.

\subsection{Proof of Proposition~\ref{prop:regret_bound_round_1}}\label{sec:prop:regret_bound_round_1}
{ We establish a satisficing regret bound for round $i$ in three parts: Step 1 that runs the oracle $\texttt{ALG}(t_i)$, Step 2 that does forced sampling on the candidate arm, and Step 3 that pulls the candidate arm with LCB test.}

Recall that $t_i=\lceil \gamma_i^{-1/(1-\alpha)}\rceil\leq 2\gamma_i^{-1/(1-\alpha)}$, { then $\log(t_i)\leq\log(2\gamma_i^{-1/(1-\alpha)})\leq \log({4}/\gamma_i)/(1-\alpha)$ and $t_i^\alpha=\lceil2^{i/\alpha}\rceil^\alpha\leq2^{\alpha(1+i/\alpha)}\leq2\cdot2^i=2\gamma_i^{-\alpha/(1-\alpha)}$. Similarly, $t_i^{\alpha-1}\leq2^{(\alpha-1)(1+i/\alpha)}=2^{\alpha-1}\cdot2^{i(\alpha-1)/\alpha}\leq2\gamma_i$. With Condition~\ref{assump:learning_alg}} the satisficing regret incurred when running $\texttt{ALG}(t_i)$ is bounded by
\begin{equation}
\begin{aligned}
&\E\left[\sum_{s=1}^{t_i}\max\{S-r(A_s),0\}\right]\\
\leq&\E\left[t_ir(A^*)-\sum_{s=1}^{t_i}r(A_s)\right]\leq C_1t_i^\alpha\log(t_i)^\beta\\
\leq& \dfrac{2C_1}{(1-\alpha)^\beta}\gamma_i^{-\alpha/(1-\alpha)}\log({4}/\gamma_i)^\beta.
\label{eq:Step1_Regret_real}
\end{aligned}
\end{equation}
Since $\hat{A}_i$ is selected randomly according to the trajectory of $\texttt{ALG}(t_i)$ in round $i$, we further have
\begin{equation}
\begin{aligned}
&\E[\max\{S-r(\hat{A}_i),0\}]\leq\E[r(A^*)-r(\hat{A}_i)]\\
=&\frac{\E\left[t_ir(A^*)-\sum_{s=1}^{t_i}r(A_s)\right]}{t_i}\\
\leq&C_1t_i^{\alpha-1}\log(t_i)^\beta\\
\leq& \dfrac{2C_1}{(1-\alpha)^\beta}\gamma_i\log({4}/\gamma_i)^\beta.
\label{eq:single_pull_regret}
\end{aligned}
\end{equation}
Here the last inequality is because $\alpha<1$ and $t_i{=\lceil\gamma_i^{-1/(1-\alpha)}\rceil} \geq \gamma_i^{-1/(1-\alpha)}$, therefore $t_i^{\alpha-1}\leq \gamma_i${, and we also have $\log(t_i)\leq\log({4}/\gamma_i)/(1-\alpha)$ because $t_i=\lceil\gamma_i^{-1/(1-\alpha)}\rceil \leq(2/\gamma_i)^{1/(1-\alpha)}$ as $1/(1-\alpha)\geq2$}. Therefore the satisficing regret~incurred when pulling $\hat{A}_i$ for $T_i=\lceil\gamma_i^{-2}\rceil$ times { during forced sampling} is bounded by
\begin{equation}
\begin{aligned}
&\left\lceil\dfrac{1}{\gamma_i^2}\right\rceil\cdot\dfrac{2C_1}{(1-\alpha)^\beta}\gamma_i\log({4}/\gamma_i)^\beta\\
\leq&\dfrac{{4}C_1}{(1-\alpha)^\beta}\gamma_i^{-1}\log({4}/\gamma_i)^\beta\\
\leq& \dfrac{{4}C_1}{(1-\alpha)^\beta}\gamma_i^{-\alpha/(1-\alpha)}\log({4}/\gamma_i)^\beta,
\end{aligned}
\label{eq:Step2_Regret_real}
\end{equation}
{ where the first inequality is because $1/\gamma_i^2\geq 1$ and then $\lceil\gamma_i^{-2}\rceil\leq2\gamma_i^{-2}$, the second inequality is because $\alpha/(1-\alpha)\geq1$ for $0.5\leq\alpha<1$.}

{ For the LCB test, we only consider the case where the candidate arm $\hat{A}_i$ is in fact non-satisficing, \ie, $r(\hat{A}_i)<S$, because pulling satisficing ones incurs zero satisficing regret.} Then for every $k$, round $i$ is not terminated after pulling { non-satisficing} arm $\hat{A}_i$ for $k$ times (in addition to the first $T_i$ times) occurs with probability at most
\begin{equation}
\begin{aligned}
\nonumber&\Pr\left(\dfrac{\hat{r}^{\text{tot}}_i}{T_i+k}-\sqrt{\dfrac{4\log(T_i+k)}{T_i+k}}>S\right)\\
\nonumber\leq &\Pr\left(\dfrac{\hat{r}^{\text{tot}}_i}{T_i+k}-\sqrt{\dfrac{4\log(T_i+k)}{T_i+k}}>r(\hat{A}_i)\right)\\
\leq& { \exp\left(-2\cdot4\log(T_i+k)\right)} \\
=& { \dfrac{1}{(T_i+k)^{8}}\leq \dfrac{1}{(k+2)^{8}}},
\label{eq:radius_hiprob}
\end{aligned}
\end{equation}
{ where the second inequality is due to Hoeffding's inequality, and the third inequality due to for any $\alpha<1$ and $i\geq1$, $2^{2i(1-\alpha)/\alpha}>1$ then $T_i=\lceil2^{2i(1-\alpha)/\alpha} \rceil\geq2$.} Therefore expected number of arm pulls of $\hat{A}_i$ after first pulling it $T_i$ times is bounded by
\begin{equation}
\begin{aligned}
\nonumber&\E\left[\text{Number of times }\hat{A}_i\text{ is pulled in round }i\text{ after the first }T_i \text{ times}\right]\\
=&\sum_{s=1}^{\infty}\Pr(k\geq s)
={ \sum_{s=1}^\infty \dfrac{1}{(s+2)^{8}}\leq 1.}
\label{eq:real_3}
\end{aligned}
\end{equation}
Since we have assumed that the expected reward is bounded on $[0,1]$ for every arm, the expected satisficing regret incurred in step 3 is upper bounded by $1$.

Summing up { the three parts of satisficing regret, \ie, the satisficing regret from Step 1 \eqref{eq:Step1_Regret_real}, Step 2 \eqref{eq:Step2_Regret_real}, and Step 3 (combination of \eqref{eq:real_3} and \eqref{eq:single_pull_regret}),} we have that the satisficing regret~of round $i$ is bounded by
\begin{equation}
\begin{aligned}
&{ \dfrac{2C_1}{(1-\alpha)^\beta}\gamma_i^{-\alpha/(1-\alpha)}\log({4}/\gamma_i)^\beta + \dfrac{4C_1}{(1-\alpha)^\beta}\gamma_i^{-\alpha/(1-\alpha)}\log({4}/\gamma_i)^\beta + 1\cdot\dfrac{2C_1}{(1-\alpha)^\beta}\gamma_i\log({4}/\gamma_i)^\beta}\\
\leq& \dfrac{{ 8}C_1}{(1-\alpha)^\beta}\gamma_i^{-\alpha/(1-\alpha)}\log({4}/\gamma_i)^{\beta}.
\label{eq:prop_round_reg}
\end{aligned}
\end{equation}

\subsection{Proof of Proposition~\ref{prop:exit_prob_1}}\label{sec:prop:exit_prob_1}

We denote $A^*=\arg\max_{A\in\mathcal{A}}r(A)$.

By Condition~\ref{assump:learning_alg} and recall from \eqref{eq:single_pull_regret} in the proof of Proposition \ref{prop:regret_bound_round_1}, we have
\begin{equation}
\begin{aligned}
\E[r(A^*)-r(\hat{A}_i)]\leq\dfrac{2C_1}{(1-\alpha)^\beta}\gamma_i\log({4}/\gamma_i)^\beta.
\label{eq:single_pull_regret_std}
\end{aligned}
\end{equation}

According to the condition of this proposition { \ref{eq:large_enough_i_1}}, we have $i$ satisfies the condition 
\begin{equation}\label{eq:large_enough_i_cond}
    \dfrac{32C_1}{(1-\alpha)^\beta}\gamma_i\log({4}/\gamma_i)^{\max\{\beta,1/2\}}\leq \Delta_S^*.
\end{equation} 
By the Markov's inequality we have
\begin{equation}
\Pr(r(A^*)-r(\hat{A}_i)\geq \Delta_S^*/2)\leq \Pr\left(r(A^*)-r(\hat{A}_i)\geq \dfrac{16C_1}{(1-\alpha)^\beta}{\gamma_i}\log({4}/\gamma_i)^\beta\right)\leq\dfrac{1}{8},\label{eq:prob_bound_1}
\end{equation}
{ where we have $\log(4/\gamma_i)\geq\log(4)>1$ given $\gamma_i<1$ for any $i$.}

If $r(\hat{A}_i)\geq r(A^*)-\Delta_S^*/2$, which happens with probability at least 7/8 according to the above inequality, then $r(\hat{A}_i)\geq S+\Delta_S^*/2$ holds. On the other hand, if $i$ satisfies the condition stated in the proposition, { then since $C_1\geq1$ and $(1-\alpha)^\beta<1$ according to the assumption in Condition~\ref{assump:learning_alg}, we have $\gamma_i\log({4}/\gamma_i)^{1/2}\leq \Delta_S^*/32$. Plugging into the confidence radius, we have} 
\begin{equation}
\begin{aligned}
\sqrt{\dfrac{{4\log(T_i+k)}}{T_i+k}} \leq& 4\sqrt{\dfrac{\log(T_i)}{T_i}} { = 4\sqrt{\frac{\log(\lceil\gamma_i^{-2}\rceil)}{\lceil\gamma_i^{-2}\rceil}}}\\
\leq& { 4\sqrt{\gamma_i^2\log(2\gamma_i^{-2})}
\leq 4\sqrt{2}\gamma_i\log({4}/\gamma_i)^{1/2}}\\
\leq& { 4\sqrt{2}\cdot\frac{\Delta_S^*}{32}} \leq { \dfrac{\Delta_S^*}{4}},
\label{eq:radius<Delta/4}
\end{aligned}
\end{equation}
{ where the first inequality is because $T_i\geq2$, and $f(y)=\log(y)/y$ is monotonically decreasing when $y\geq 3$ and $2\cdot\log(2)/2\geq \log(3)/3$ for all $k\geq 0$, thus $\log(T_i+k)/(T_i+k)\leq 4\log(T_i)/T_i$ for all $k\geq 0$; the second inequality is because $\gamma_i^{-1}\geq1$, thus $\lceil\gamma_i^{-2}\rceil\leq2\gamma_i^{-2}$. The third inequality is because $\log(2\gamma_i^{-2})=2\log(\sqrt{2}/\gamma_i)\leq 2\log(4/\gamma_i)$. The fourth inequality follows from~\eqref{eq:large_enough_i_cond}. }Suppose round $i$ is terminated for some $k$, then
\begin{equation*}
\dfrac{\hat{r}_i^{\text{tot}}}{T_i+k}-\sqrt{\dfrac{4\log(T_i+k)}{T_i+k}}<S=r(A^*)-\Delta_S^*\leq r(\hat{A}_i)-\dfrac{\Delta_S^*}{2}\leq r(\hat{A}_i)-2\sqrt{\dfrac{4\log(T_i+k)}{T_i+k}},
\end{equation*}
or equivalently
\begin{equation*}
\dfrac{\hat{r}_i^{\text{tot}}}{T_i+k}\leq r(\hat{A}_i)-\sqrt{\dfrac{4\log(T_i+k)}{T_i+k}}.
\end{equation*}
By the Hoeffding's inequality we have
\begin{equation*}
\Pr\left(\dfrac{\hat{r}_i^{\text{tot}}}{T_i+k}\leq r(\hat{A}_i)-\sqrt{\dfrac{4\log(T_i+k)}{T_i+k}}\right)\leq { \dfrac{1}{(T_i+k)^8}\leq \dfrac{1}{(2+k)^8}},
\end{equation*}
{ where the second inequality is because $T_i\geq2$ for any $i\geq1$ and $\alpha<1$.} Therefore conditioned on $r(\hat{A}_i)\geq r(A^*)-\Delta_S^*/2$, the probability that round $i$ is terminated by the end of the horizon is bounded by
\begin{equation}
{ \sum_{k=0}^\infty \dfrac{1}{(k+2)^8}}\leq \dfrac{1}{8}.\label{eq:prob_bound_2}
\end{equation}
Combining~\eqref{eq:prob_bound_1}~and~\eqref{eq:prob_bound_2}~we conclude that the probability that round $i$ terminates by the end of the time horizon is { upper} bounded by $1/4$.

\section{Proof of Theorem~\ref{thm:regret_nonsat}}\label{sec:thm:regret_nonsat}

We first prove that the { standard} regret incurred in round $i$ is at most
\begin{equation*}
\dfrac{{8}C_1}{(1-\alpha)^\beta}\gamma_i^{-\alpha/(1-\alpha)}\log({4}/\gamma_i)^{\beta}.
\end{equation*}

The regret incurred when running $\texttt{ALG}(t_i)$ { in Step 1} is bounded by
\begin{equation*}
\E\left[t_ir(A^*)-\sum_{s=1}^{t_i}r(A_s)\right]\leq C_1t_i^\alpha\log(t_i)^\beta\leq \dfrac{2C_1}{(1-\alpha)^\beta}\gamma_i^{-\alpha/(1-\alpha)}\log({4}/\gamma_i)^\beta.
\end{equation*}
{ Recall that in \eqref{eq:single_pull_regret_std}  we have shown that}
\begin{equation*}
\E[r(A^*)-r(\hat{A}_i)]\leq \dfrac{2C_1}{(1-\alpha)^\beta}\gamma_i\log({4}/\gamma_i)^\beta.
\end{equation*}
Therefore { similar to \eqref{eq:Step2_Regret_real},} the standard regret~incurred when pulling $\hat{A}_i$ { in Step 2 forced sampling} is bounded by 
\begin{equation*}
\left\lceil\dfrac{1}{\gamma_i^2}\right\rceil\cdot\dfrac{2C_1}{(1-\alpha)^\beta}\gamma_i\log({4}/\gamma_i)^\beta \leq \dfrac{{4}C_1}{(1-\alpha)^\beta}\gamma_i^{-1}\log({4}/\gamma_i)^\beta\leq \dfrac{{4}C_1}{(1-\alpha)^\beta}\gamma_i^{-\alpha/(1-\alpha)}\log({4}/\gamma_i)^\beta.
\end{equation*}
Since $r(\hat{A}_i)\leq r(A^*)<S$, { similar to \eqref{eq:radius_hiprob},} for every $k$, round $i$ is not terminated with probability at most
\begin{equation*}
\Pr\left(\dfrac{\hat{r}^{\text{tot}}_i}{T_i+k}-\sqrt{\dfrac{4\log(T_i+k)}{T_i+k}}>S\right)\leq {\dfrac{1}{(k+2)^{8}}}.
\end{equation*}
Therefore expected number of arm pulls of $\hat{A}_i$ after pulling $T_i$ times { in Step 3} is bounded by
\begin{equation*}
\sum_{k=1}^\infty \dfrac{1}{{(k+2)^{8}}}\leq 1.
\end{equation*}
Summing up the { standard regret from Step 1, 2, and 3 in a round $i$}, we have that the standard regret~of round $i$ is bounded by
\begin{equation*}
\begin{aligned}
&\dfrac{2C_1}{(1-\alpha)^\beta}\gamma_i^{-\alpha/(1-\alpha)}\log({4}/\gamma_i)^{\beta} + \dfrac{4C_1}{(1-\alpha)^\beta}\gamma_i^{-\alpha/(1-\alpha)}\log({4}/\gamma_i)^{\beta} + {1\cdot\dfrac{2C_1}{(1-\alpha)^\beta}\gamma_i\log({4}/\gamma_i)^\beta} \\
\leq&\dfrac{{8}C_1}{(1-\alpha)^\beta}\gamma_i^{-\alpha/(1-\alpha)}\log({4}/\gamma_i)^{\beta}.
\end{aligned}
\end{equation*}
Furthermore, since the length of the time horizon is $T$ time steps, we have $t_i\leq T$. Since by definition of $t_i=\lceil2^{i/\alpha}\rceil$, { the number of rounds that end within the time horizon is bounded by $2^{i/\alpha}\leq T$, \ie, $i\leq\lfloor\alpha\log_2(T) \rfloor$ because $i$ is an integer. Then,} the maximum number of rounds is bounded by $i_{\max}=\lceil\alpha\log_2(T)\rceil$. Following the same chain of inequalities in \eqref{eq:main_bound_sublinear}, we can bound the standard regret by
\begin{align*}
\mathbb{E}[\texttt{Regret}]\leq&\sum_{i=1}^{i_{\max}}\dfrac{{8}C_1}{(1-\alpha)^\beta}\gamma_i^{-\alpha/(1-\alpha)}\log({4}/\gamma_i)^\beta\\
\leq& \dfrac{{64}C_1}{(1-\alpha)^\beta}T^\alpha\log({4}T)^\beta.
\end{align*}
Therefore, we have proven that the regret is bounded by $C_1T^\alpha\cdot\text{polylog}(T)$.

\section{Proofs for Lower Bounds}

{

\subsection{Proof of Theorem~\ref{thm:lb_two_arm}}

We consider the following $K$ instances of finite-arm bandits with $K$ arms. We set the satisficing threshold $S=1/2+\Delta$. The mean reward of each arm follows a Bernoulli distribution. For each $k\in[K]$, the mean rewards of the $k$-th instance are defined as
\begin{equation*}
r_k=\dfrac{1}{2}+2\Delta,\ r_{k'}=\dfrac{1}{2},\ \forall k\neq k'.
\end{equation*}
Similar lower bound instances are also considered in Section 2 of \cite{Slivkins2019}. By Lemma 2.8 of \cite{Slivkins2019}, there exists $0<c\leq 1$ such that for any policy that aims to identify the optimal arm within $cK/\Delta^2$ time steps, there exists $\lceil K/3\rceil$ instances such that the policy incorrectly identifies the optimal arm with probability at least $1/4$. We formally restate the lemma in \cref{lemma:aux} in \cref{sec:aux_results} of the appendix. If we consider the arm pull of policy $\pi$ at time step $t$ as a policy that commits one arm $A_t$ after $t-1$ arm pulls, then by \cref{lemma:aux} we have that for all $t\leq cK/\Delta^2$, there exists at least $\lceil K/3\rceil$ arms $k\in[K]$ such that $\Pr^{(k)}(A_t=k)\leq 3/4$. Therefore we have that $\sum_{k=1}^K\Pr^{(k)}(A_t\neq k)\geq K/3\cdot 1/4=K/12$. Since under any instance $k\in[K]$, pulling any arm $k'\neq k$ incurs a satisficing regret of $\Delta$. Therefore we have
\begin{equation*}
\sum_{k=1}^K\mathbb{E}^{(k)}[\texttt{Regret}_S]=\sum_{t=1}^{T}\sum_{k=1}^K\Delta\Pr^{(k)}(A_t\neq k)\geq \sum_{t=1}^{\lfloor cK/\Delta^2\rfloor}\sum_{k=1}^K\Delta\Pr^{(k)}(A_t\neq k)\geq \dfrac{\Delta K}{12}\cdot \left\lfloor\dfrac{cK}{\Delta^2}\right\rfloor=\Omega\left(\dfrac{K^2}{\Delta}\right).
\end{equation*}
Here the first inequality is due to the assumption that $T\geq K/\Delta^2$ and $c\leq 1$. Therefore there exists $k\in[K]$ such that the expected satisficing regret $\mathbb{E}^{(k)}[\texttt{Regret}_S]$ is at least $\Omega(K/\Delta)$.
}

{
\subsection{Proof of Theorem~\ref{thm:lb_concave}}

We define the arm set as $\mathcal{A}=[-1,1]$. We consider the following two instances of concave bandits. The reward functions of the two instances are defined as follows.
\begin{align*}
&\textbf{Instance 1:}\ r^{(1)}(x)=\begin{cases}2\sqrt{2\Delta}x+6\Delta+\dfrac{1}{2},\ &\text{if}\ -1\leq x\leq -2\sqrt{2\Delta},\\
\dfrac{1}{2}-(x+\sqrt{2\Delta})^2,\ &\text{if}\ -2\sqrt{2\Delta}<x<0\\
-2\sqrt{2\Delta}x+\dfrac{1}{2}-2\Delta,\ &\text{if}\ 0\leq x\leq 1;
\end{cases}\\
&\textbf{Instance 2:}\ r^{(2)}(x)=\begin{cases}2\sqrt{2\Delta} x+\dfrac{1}{2}-2\Delta,\ &\text{if}\ -1\leq x\leq 0,\\
\dfrac{1}{2}-(x-\sqrt{2\Delta})^2,\ &\text{if}\ 0<x<2\sqrt{2\Delta},\\
-2\sqrt{2\Delta}x+6\Delta+\dfrac{1}{2},\ &\text{if}\ 2\sqrt{2\Delta}
\leq x\leq 1.\end{cases}
\end{align*}
Since we have $\Delta<1/162$, we have $1/2\geq r^{(1)}(x)\geq 1/2-2\Delta-2\sqrt{2\Delta}\geq 1/2-1/81-2/9>1/4$ for all $x\in[-1,1]$. Similarly, we also have $1/2\geq r^{(2)}(x)\geq 1/4$ for all $x\in[-1,1]$.

We set $S=1/2-\Delta$, and the random reward of arm $x$ follows a Bernoulli distribution with mean $r^{(i)}(x)$ under instance $i$ for $i\in\{1,2\}$. Since we have proven that $r^{(1)}(x),r^{(2)}(x)\in[0,1]$ for all $x\in[-1,1]$, the definition of random reward is valid. Since for both instances, the optimal reward is $1/2$, thus both instances satisfy $r(A^*)-S=\Delta$.

By definition of $r^{(1)}$ and $r^{(2)}$ we have
\begin{align*}
&\dfrac{\text{d}r^{(1)}}{\text{d}x}(x)=\begin{cases}
2\sqrt{2\Delta},\ &\text{if}\ x<-2\sqrt{2\Delta},\\
-2x-2\sqrt{2\Delta},\ &\text{if}\ -2\sqrt{2\Delta}\leq x\leq 0,\\
-2\sqrt{2\Delta},\ &\text{if}\ x>0,
\end{cases}\\
&\dfrac{\text{d}r^{(2)}}{\text{d}x}(x)=\begin{cases}
2\sqrt{2\Delta},\ &\text{if}\ x<0,\\
-2x+2\sqrt{2\Delta},\ &\text{if}\ 0\leq x\leq 2\sqrt{2\Delta},\\
-2\sqrt{2\Delta},\ &\text{if}\ x>2\sqrt{2\Delta}.
\end{cases}
\end{align*}
Therefore we have $r^{(1)}(x)-r^{(2)}(x)$ is monotonically non-increasing, thus $r^{(1)}(x)-r^{(2)}(x)$ reaches maximum when $x=-1$ and reaches minimum when $x=1$. We further get
\begin{equation*}
r^{(1)}(-1)-r^{(2)}(-1)=8\Delta,\ r^{(1)}(1)-r^{(2)}(1)=-8\Delta.
\end{equation*}
Therefore we have that $|r^{(1)}(x)-r^{(2)}(x)|\leq 8\Delta$ for all $x\in[-1,1]$.

Next, we prove that for any $p,q\in[1/4,1/2]$, we have $\texttt{KL}(\text{Bernoulli}(p)\|\text{Bernoulli}(q))\leq 8(p-q)^2$. For any $p\in[1/4,1/2]$, we define $f_p(q)$ as
\begin{equation*}
f_p(q)=\texttt{KL}(\text{Bernoulli}(p)\|\text{Bernoulli}(q))=p\log\left(\dfrac{p}{q}\right)+(1-p)\log\left(\dfrac{1-p}{1-q}\right).
\end{equation*}
Then we have $f_p(p)=0$ and $f_p'(q)=-p/q+(1-p)/(1-q)$, thus $f'_p(p)=0$. We further have that for all $p,q\in[1/4,1/2]$,
\begin{equation*}
f''_p(q)=\dfrac{p}{q^2}+\dfrac{1-p}{(1-q)^2}\leq \dfrac{p}{(1/4)^2}+\dfrac{1-p}{(1/2)^2}=4+12p\leq10.
\end{equation*}
Here the first inequality is due to $q\in[1/4,1/2]$ and thus $1-q\in[1/2,3/4]$. Furthermore, by Taylor's theorem, there exists $\xi\in(p,q)$ such that
\begin{equation*}
f_p(q)=f_p(p)+f_p'(p)(q-p)+\dfrac{1}{2}f''_p(\xi)(q-p)^2.
\end{equation*}
Since $f_p(p)=f'_p(p)=0$, we have $f_p(q)=f''_p(\xi)(q-p)^2/2\leq5(q-p)^2 \leq 8(q-p)^2$, therefore for all $p,q\in[1/4,1/2]$, we have $\texttt{KL}(\text{Bernoulli}(p)\|\text{Bernoulli}(q))\leq 8(q-p)^2$. Since we have proven that $|r^{(1)}(x)-r^{(2)}(x)|\leq 8\Delta$, we have
\begin{equation*}
\texttt{KL}\left(\text{Bernoulli}(r^{(1)}(x)) \big\|\text{Bernoulli}(r^{(2)}(x))\right)\leq 8(r^{(1)}(x)-r^{(2)}(x))^2\leq 512\Delta^2.
\end{equation*}

We define $T_0=\lfloor 1/512\Delta^2\rfloor$, and define $\Pr^{(i)}_t$ as the probability space of a policy $\pi$ by time step $t$ under instance $i$. Then we have
\begin{equation}
\begin{aligned}
\texttt{KL}(\Pr^{(1)}_t\|\Pr^{(2)}_t)\leq& t\cdot\max_{x\in[-1,1]} \texttt{KL}\left(\text{Bernoulli}(r^{(1)}(x))\big\|\text{Bernoulli}(r^{(2)}(x))\right)\\
\leq& t\cdot 512\Delta^2\leq T_0\cdot 512\Delta^2\leq 1.
\end{aligned}\label{eq:kl_upper_bound}
\end{equation}
Here the second inequality follows from~\eqref{eq:kl_upper_bound}. By the Bretagnolle-Huber inequality (see, \eg, Theorem 14.2 of \citealt{LS20}), we have that for all $t\leq T_0$,
\begin{equation}\label{eq:huber_ineq}
\Pr^{(1)}(A_t>0)+\Pr^{(2)}(A_t\leq0)\geq \dfrac{1}{2}\exp(-\texttt{KL}(\Pr^{(1)}_t\|\Pr^{(2)}_t))\geq \dfrac{1}{2}\exp(-1).
\end{equation}
We denote $\regret_S^{(i)}$ as the satisficing regret under instance $i$. By definition of $r^{(1)}$ and $r^{(2)}$ we have $r^{(1)}(x)\leq 1/2-2\Delta$ for all $x>0$ and $r^{(2)}(x)\leq 1/2-2\Delta$ for all $x\leq0$. Since $S=1/2-\Delta$, pulling any arm $x>0$ under instance $1$ and pulling any arm $x\leq0$ under instance $2$ incurs a satisficing regret of at least $\Delta$. Therefore we have
\begin{align*}
\mathbb{E}^{(1)}[\regret_S]+\mathbb{E}^{(2)}[\regret_S]&\geq \sum_{t=1}^T \Delta(\Pr^{(1)}(A_t>0)+\Pr^{(2)}(A_t\leq0))\\
&\geq \sum_{t=1}^{T_0}\Delta(\Pr^{(1)}(A_t>0)+\Pr^{(2)}(A_t\leq0))\\
&\geq T_0\Delta\cdot \dfrac{1}{2}\exp(-1)=\dfrac{1}{2}\exp(-1)\Delta\left\lfloor\dfrac{1}{512\Delta^2}\right\rfloor=\Omega\left(\dfrac{1}{\Delta}\right).
\end{align*}
Here the first inequality is due to the assumption that $T\geq 1/\Delta^2$, and the second inequality follows from \eqref{eq:huber_ineq}. Therefore under at least one of instance $1$ and $2$, the policy incurs $\Omega(1/\Delta)$ regret.
}

{

\subsection{Proof of Theorem~\ref{thm:lb_lipschitz}}

We define $M=\lfloor L/4\Delta\rfloor$, and we define $M^d$ instances indeed by $\bm{m}\in [M]^d$. The arm set is defined as $[0,1]^d$, and the reward of each arm follows a Bernoulli distribution. For every $\bm{m}\in[M]^d$, we define $\bm{y}^{(\bm{m})}$ as $y_j^{\bm{m}}=(2m_j-1)/2M$ for every $j\in[d]$, and the mean reward function $r^{(\bm{m})}$ of instance $\bm{m}$ is defined as
\begin{equation*}
r^{(\bm{m})}(\bm{x})=\max\left\{\dfrac{1}{2},\dfrac{1}{2}+2\Delta-L\|\bm{x}-\bm{y}^{(\bm{m})}\|_\infty\right\}.
\end{equation*}

For any $\bm{x}\in[0,1]^d$ and $\bm{m}\in [M]^d$ such that $\|\bm{x}-\bm{y}^{(\bm{m})}\|_\infty\geq 1/2M$, since $M=\lfloor L/4\Delta\rfloor\leq L/4\Delta$, we have $\|\bm{x}-\bm{y}^{(\bm{m})}\|_\infty\geq 2\Delta/L$, thus $r^{(\bm{m})}(\bm{x})=1/2$. Therefore $r^{(\bm{m})}(\bm{x})>1/2$ only when $\|\bm{x}-\bm{y}^{(\bm{m})}\|_\infty \leq 1/2M$. We still set $S=1/2+\Delta$.

We also construct $M^d$ instances of finite armed bandit indexed by $[M]^d$. The reward function of instance $\bm{m}$ is denoted as $\bar{r}^{(\bm{m})}$. The arm set is set to be $[M]^d$, and for each $\bm{m}\in [M]^d$, we define instance $\bm{m}$ as
\begin{equation}\label{eq:instance_lipschitz_to_finite}
\bar{r}^{(\bm{m})}(\bm{m})=\dfrac{1}{2}+2\Delta,\ \bar{r}^{(\bm{m})}(\bm{m}')=\dfrac{1}{2},\ \forall \bm{m}'\neq \bm{m}.
\end{equation}
Namely, for instance $\bm{m}$, $\bm{m}$ is the single best arm and any other arm $\bm{m}'\neq\bm{m}$ is sub-optimal. Based on the policy $\pi$, we construct a policy $\pi'$ for finite-arm bandits with arm set being $[M]^d$. For any $\bm{x}\in[0,1]^d$, we define $\bm{m}(\bm{x})=\arg\min\{\|\bm{x}-\bm{y}^{(\bm{m})}\|_{\infty}:\bm{m}\in[M]^d\}$. For each round $t\in[T]$, suppose the policy $\pi$ chooses arm $\bm{x}$, then policy $\pi'$ pulls arm $\bm{m}(\bm{x})$ and observes a random reward $R$. Then, with probability $p_{\bm{x}}$, we feed the random reward $R$ back to policy $\pi$, and with probability $1-p_{\bm{x}}$, we generate a Bernoulli random variable with mean $1/2$ and feed the realized random variable back to policy $\pi$. Here, we define
\begin{equation*}
p_{\bm{x}}=\dfrac{1}{2\Delta}\max\{2\Delta-L\|\bm{x}-\bm{y}^{(\bm{m}(\bm{x}))}\|_\infty,0\}.
\end{equation*}
We show that under any instance $\bm{m}\in[M]^d$, the random reward fed back to $\pi$ follows a Bernoulli distribution of $r^{(\bm{m})}(\bm{x})$. For any $\bm{m}\neq\bm{m}(\bm{x})$, because the arm $\bm{m}(\bm{x})$ is sub-optimal, the random reward $R$ follows a Bernoulli distribution with mean $1/2$, thus the reward fed back to $\pi$ also follows a distribution with mean $1/2$, which is equal to $r^{(\bm{m})}(\bm{x})$ since $\bm{m}\neq \bm{m}(\bm{x})$. If $\bm{m}=\bm{m}(\bm{x})$, then the probability the reward fed back to $\pi$ is equal to $1$ is equal to
\begin{align*}
&p_{\bm{x}}\bar{r}^{(\bm{m}(\bm{x}))}(\bm{m}(\bm{x}))+(1-p_{\bm{x}})\cdot \dfrac{1}{2}=p_{\bm{x}}\left(\dfrac{1}{2}+2\Delta\right)+(1-p_{\bm{x}})\cdot \dfrac{1}{2}\\
=&\dfrac{1}{2}+2\Delta p_{\bm{x}}=\dfrac{1}{2}+\max\{2\Delta-L\|\bm{x}-\bm{y}^{(\bm{m}(\bm{x}))}\|_\infty,0\}\\
=&\max\left\{\dfrac{1}{2}+2\Delta-L\|\bm{x}-\bm{y}^{(\bm{m}(\bm{x}))}\|_\infty,\dfrac{1}{2}\right\}=r^{(\bm{m}(\bm{x}))}(\bm{x}).
\end{align*}
Therefore under any instance $\bm{m}\in [M]^d$, the mean reward fed back to $\pi$ is equal to $r^{(\bm{m})}(\bm{x})$, thus the reward observed by the policy $\pi$ in this setting follows the same distribution as running the policy on the corresponding Lipschitz bandit instance.

Next, we show that the expected satisficing regret incurred by $\pi$ is greater than or equal to that incurred by $\pi'$. It suffices to prove that $r^{(\bm{m})}(\bm{x})\leq \bar{r}^{(\bm{m})}(\bm{m}(\bm{x}))$. If $\bm{m}=\bm{m}(\bm{x})$, then by definition we have $\bar{r}^{(\bm{m})}(\bm{m}(\bm{x}))=1/2+2\Delta\geq r^{(\bm{m})}(\bm{x})$. If $\bm{m}(\bm{x})\neq \bm{m}$, we define $\bm{m}(\bm{x})=\bm{m}'$. By definition of $\{\bm{y}^{(\bm{m})}:\bm{m}\in[M]^d\}$, for every $\bm{x}\in [0,1]^d$, there exists $\bm{m}\in[M]^d$ such that $\|\bm{x}-\bm{y}^{(\bm{m})}\|_\infty\leq 1/(2M)$. Since by definition $\bm{y}^{(\bm{m}')}$ is the closest to $\bm{x}$ among $\{\bm{y}^{(\bm{m})}:\bm{m}\in[M]^d\}$ measured by infinite norm, we have $\|\bm{x}-\bm{y}^{(\bm{m}')}\|\leq 1/(2M)$. Furthermore, since $\bm{m}\neq \bm{m}'$, by definition of $\bm{y}^{(\bm{m})}$ we have $\|\bm{y}^{(\bm{m})}-\bm{y}^{(\bm{m}')}\|_\infty\geq 1/M$, therefore
\begin{equation*}
\|\bm{x}-\bm{y}^{(\bm{m})}\|_\infty\geq \|\bm{y}^{(\bm{m})}-\bm{y}^{(\bm{m}')}\|_\infty-\|\bm{x}-\bm{y}^{(\bm{m}')}\|_\infty \geq \dfrac{1}{M}-\dfrac{1}{2M}=\dfrac{1}{2M}.
\end{equation*}
We have proven that $r^{(\bm{m})}(\bm{x})=1/2$ for all $\bm{x}\in[0,1]^d$ such that $\|\bm{x}-\bm{y}^{(\bm{m})}\|_\infty \geq 1/(2M)$. Therefore in this case we also have $\bar{r}^{(\bm{m})}(\bm{m}(\bm{x}))\geq 1/2=r^{(\bm{m})}(\bm{x})$. Then we conclude that the mean reward of any arm pull of $\pi$ is smaller than or equal to that of $\pi'$, thus under any instance $\bm{m}\in [M]^d$, the expected satisficing regret of $\pi$ is greater than or equal to that of $\pi'$.

By Theorem~\ref{thm:lb_two_arm}, when $T\geq M^d/\Delta^{2}$, there exists an instance $\bm{m}\in[M]^d$ as constructed in \eqref{eq:instance_lipschitz_to_finite} such that the expected satisficing regret of $\pi'$ is at least $\Omega(M^d/\Delta)$. Since $M=\Theta(L/\Delta)$, and the expected satisficing regret of $\pi$ is greater than or equal to that of $\pi'$, we have that there exists an instance such that the expected satisficing regret of $\pi$ is at least $\Omega(L^d/\Delta^{d+1})$.

}

\section{Proof of Proposition \ref{thm:heavy_tail_2arm}}
\label{sec:prop:heavy_tail_2arm}
We consider a simple instance of two-armed bandit: $r(1)=1$ and $r(2)=0$ with $S=0.5$. 


{ Since for finite-armed bandit, there exist algorithms that attain a $\tilde{O}(\sqrt{T})$ regret bound, we have $\alpha=1/2$, thus by definition of $t_i=\lceil2^{i/\alpha}\rceil$. For ease of discussion, we assume that the algorithm \satex~runs indefinitely, but only the satisficing regret incurred during the first $T$ time periods is considered. We define $i_{\max}=\lceil\alpha\log_2(T)\rceil$, and define $\mathcal{E}$ as the event that in the infinite time horizon, there are at least $i_{\max}$ rounds that are terminated.} Recall that we have assumed that the probability that arm 2 is pulled at least once in each round $i\in[l]$ is at least $1/2$. Define event $\mathcal{E}:=\{\text{arm 2 is pulled in Step 2 for every round }i\in[l] \}$, { and $\mathcal{E}_i:=\{\text{arm 2 is pulled in Step 2 in round }i\}$. Then for each round $i$, $\mathcal{E}_i$ happens if arm 2 is pulled at least once in Step 1, and is further sampled to Step 2. According to the assumption in Proposition~\ref{thm:heavy_tail_2arm}, the probability that arm 2 is pulled at least once in Step 1 is at least $1/2$. During the random sampling that proposes a candidate arm at the end of Step 1, arm 2 is sampled as candidate with probability at least $1/t_i$ given it is pulled at least once in Step 1. Then, $\Pr(\mathcal{E}_i)\geq1/2\cdot1/t_i$ for any $i$. Recall that each round is independent,} $\mathcal{E}$ happens with probability at least
\begin{equation}
\begin{aligned}
\Pr(\mathcal{E})&=\prod_{i=1}^l{ \Pr\left(\mathcal{E}_i \right)}\geq \prod_{i=1}^l\frac{1}{2}\cdot\frac{1}{t_i} \geq \prod_{i=1}^{i_{\max}}\frac{1}{2}\cdot\frac{1}{t_i} \\ 
&\overset{ \text{(i)}}{\geq} 2^{-(1+\log_2(T){/2})}\cdot\prod_{i=1}^{1+\log_2(T){/2}}2^{-(1+{2}i)}
={ 2^{-(2+\log_2(T))} \cdot \prod_{i=1}^{1+\log_2(T)/2}2^{-2i}}  \\
&={ \frac{1}{4T}\cdot \prod_{i=1}^{1+\log_2(T)/2}2^{-2i} = \frac{1}{4T} \cdot 2^{-(2+\log_2(T)/2)(1+\log_2(T)/2)}} \\
&{ \overset{\text{(ii)}}{\geq} \frac{1}{4T} \cdot 2^{-\log_2(T)^2} = \frac{1}{4} \cdot \frac{\exp(-\log(2)\log_2(T)^2)}{\exp(\log(T))} }\\
& { = \frac{1}{4}\exp(-(\log(T)^2/\log(2) + \log(T)) )}\\
& { \overset{\text{(iii)}}{\geq} \frac{1}{4}\exp(-2\log(T)^2) }.
\label{eq:P(E)}
\end{aligned}
\end{equation}
{ where step (i) is based on $i_{\max}\leq1+\log_2(T)/2$ and the definition of $t_i=\lceil2^{i/\alpha}\rceil\leq2^{2i+1}$ for $\alpha=1/2$, step (ii) holds for $T\geq8$, and step (iii) holds with $1+\log(T)/\log(2)\leq2\log(T)$ for $T\geq8$.}

{ Next, we show that if $\mathcal{E}$ holds, then the satisficing regret of \satex~is at least $T/14$.} By definition we have $t_{i_{\max}}=2^{2i_{\max}}\geq T$, therefore within a time horizon of $T$ time steps, the total number of rounds cannot exceed $i_{\max}$. Therefore if $\mathcal{E}$ holds, then for each round $i$ that is started within a time horizon of $T$ time periods, the candidate satisficing arm $\hat{A}_i$ is arm $2$. If the total number of time steps for Step 3 is at least { $T/7$}, the total satisficing regret is at least {$T/7\cdot1/2=T/14$}. { Otherwise, the total number of time steps for Step 1 and 2 is at least $6T/7$. In this case, we claim that the total number of time steps for Step 2 is at least $T/7$. Let $i_0$ be the number of rounds that terminates within $T$ time steps. Then the total number of time steps for Steps 2 is at least $\sum_{i=1}^{i_0}4^i$, and the total number of time steps for Step 1 is at most $\sum_{i=1}^{i_0+1}4^i$. The ratio between the two is at least
\begin{equation*}
\dfrac{\sum_{i=1}^{i_0}4^i}{\sum_{i=1}^{i_0+1}4^i}= \frac{4^{i_0}-1}{4^{i_0+1}-1} \geq \dfrac{1}{5},
\end{equation*}
where the inequality is based on $i_0\geq1$. Therefore the total number of time steps for Step 1 is at most $5$ times that of Step 2, thus the total number of time steps for Step 2 is at least $T/7$. Since under $\mathcal{E}$, \satex~always pulls arm 2 in Step 2, the satisficing regret incurred during Step 2 is at least { $T/7\cdot1/2=T/14$.} Combining both cases, we know that the satisficing regret is always at least $T/14$ under $\mathcal{E}$.} Therefore, we can bound the probability of total satisficing regret exceeding $T/14$ as
\begin{equation*}
\begin{aligned}
\Pr\left(\texttt{Regret}_S\geq \frac{T}{14} \right)\geq \Pr(\mathcal{E}) \geq { \frac{1}{4}\exp(-2\log(T)^2)},
\end{aligned}
\end{equation*}
{ where the final inequality is based on \eqref{eq:P(E)}.}

\section{Proof of Theorem~\ref{thm:tail_risk}}\label{sec:thm:tail_risk}

The inequality in the theorem is trivial when $x<x_0$ { as the right-hand side is at least $\exp(2i_0+4)>1\geq\Pr(\regret_S>x)$}, thus for the rest of the proof we assume $x\geq x_0$. 

Denote $\beta'=\max\{\beta,1/2\}$ and $K_{\alpha,\beta,\zeta}=\frac{32C_1(1-\zeta)^{-\zeta/2}}{(1-\alpha)^\beta} {\sqrt{2 + 2\log\left(16\zeta^{-1}\Gamma(\zeta^{-1})\right)}}$, and $i_0$ as
\begin{equation*}
i_0=\max\{i: K_{\alpha,\beta,\zeta}\gamma_i^{1-\zeta}\log({4}/\gamma_i)^{\max\{\beta,1/2\}}>\Delta_S^*\}.
\end{equation*}
Define the satisficing regret incurred in the Step 1 and 2 in each round $i$ as $R_i$, then { the probability that cumulative satisficing regret incurred in the Step 1 and 2 after round $i_0$ exceeds $x/3$} is upper bounded by
\begin{equation*}
\begin{aligned}
    \Pr\left(\sum_{i=i_0+1}^{\infty}R_i>\frac{x}{3} \right) &= { \sum_{k=1}^{\infty}\Pr\left(\sum_{i=i_0+1}^{\infty}R_i>\frac{x}{3} \text{ and algorithm ends at round } i_0+k \right)} \\
    &= \sum_{k=1}^{\infty}\Pr\left(\sum_{i=i_0+1}^{{\infty}}R_i>\frac{x}{3} \bigg| \text{Ends at round } i_0+k \right)\cdot \Pr(\text{Ends at round } i_0+k)\\
    &\leq \sum_{k=1}^{\infty}\Pr\left(\sum_{i=i_0+1}^{i_0+k}R_i>\frac{x}{3} \right)\cdot 4^{1-k},
\end{aligned}
\end{equation*}
{ where the final inequality is based on Proposition~\ref{prop:exit_prob_tail}. }

We first prove an upper bound on $i_0$. Denote $\delta_0\in(0,1)$ the smallest solution to the following equation
\begin{equation}
\delta_0^{1-\zeta}\log({4}/\delta_0)^{\beta'}=\dfrac{\Delta_S^*}{K_{\alpha,\beta,\zeta}}.
\label{eq:delta_Delta_equation-lite}
\end{equation}
{ Here $\delta_0$ always exists because the following observations: $\delta^{1-\zeta}\log(4/\delta)^{\beta'}$ is continuous on $(0,1]$, $\delta^{1-\zeta}\log(4/\delta)^{\beta'} \to 0$ when $\delta\to 0$, and $\delta^{1-\zeta}\log(4/\delta)^{\beta'}>1$ when $\delta=1$; furthermore, $\Delta_S^*\leq 1$ and $K_{\alpha,\beta,\zeta}\geq 1$, thus the right hand side lies in $(0,1]$. Therefore such $\delta_0$ always exists.} Then we have
\begin{equation*}
\begin{aligned}
\dfrac{1}{\delta_0^{1-\zeta}} =& \dfrac{K_{\alpha,\beta,\zeta}\log({4}/\delta_0)^{\beta'}}{\Delta_S^*} \\
=&{ \dfrac{K_{\alpha,\beta,\zeta}/(1-\zeta)^{\beta'}\cdot(4\beta')^{\beta'}\log\left((4/\delta_0)^{(1-\zeta)/(4\beta')}\right)^{\beta'}}{\Delta_S^*}}\\
\leq& { \dfrac{K_{\alpha,\beta,\zeta}/(1-\zeta)^{\beta'}\cdot(4\beta')^{\beta'}\left(4/\delta_0^{1-\zeta}\right)^{1/4}}{\Delta_S^*}}\\
=& { \dfrac{K_{\alpha,\beta,\zeta}/(1-\zeta)^{\beta'}\cdot(4\beta')^{\beta'}2^{1/2}}{\Delta_S^*\left(\delta_0^{1-\zeta}\right)^{1/4}}}\\
\leq& { \dfrac{(8\beta')^{\beta'}K_{\alpha,\beta,\zeta}/(1-\zeta)^{\beta'}}{\Delta_S^*\left(\delta_0^{1-\zeta}\right)^{1/4}}},
\end{aligned}
\end{equation*}
{ where the first inequality is because $\log(x)<x$ for all $x>0$ and $2^{1-\zeta}\leq2$, and the second inequality is because $\beta'=\max\{\beta,1/2\}\geq1/2$, thus $2^{1/2}\leq 2^{\beta'}$. Therefore}
\begin{equation}
\dfrac{1}{\delta_0^{1-\zeta}}\leq \left(\dfrac{(8\beta')^{\beta'}K_{\alpha,\beta,\zeta}/(1-\zeta){^{\beta'}}}{\Delta_S^*}\right)^{4/3},
\label{eq:1/delta^{1-zeta}}
\end{equation}
{ and this further implies that 
\begin{equation*}
\begin{aligned}
\log({4}/\delta_0) &\leq \log\left(4 \left(\dfrac{(8\beta')^{\beta'}K_{\alpha,\beta,\zeta}/(1-\zeta)^{\beta'}}{\Delta_S^*}\right)^{4/(3(1-\zeta))} \right) \\
&\leq \log\left(\left(\dfrac{4^{1-\zeta}(8\beta')^{\beta'}K_{\alpha,\beta,\zeta}/(1-\zeta)^{\beta'}}{\Delta_S^*}\right)^{4/(3(1-\zeta))} \right)\\
&= \frac{4}{3(1-\zeta)}\log\left(\dfrac{4^{1-\zeta}(8\beta')^{\beta'}K_{\alpha,\beta,\zeta}/(1-\zeta)^{\beta'}}{\Delta_S^*}\right).
\end{aligned}
\end{equation*}
Plugging this inequality back to the $\log({4}/\delta_0)$ term in \eqref{eq:delta_Delta_equation-lite}, we have
\begin{equation*}
\dfrac{1}{\delta_0^{1-\zeta}} = \dfrac{K_{\alpha,\beta,\zeta}\log({4}/\delta_0)^{\beta'}}{\Delta_S^*} \leq \frac{K_{\alpha,\beta,\zeta}}{\Delta_S^*}\cdot \left(\frac{4}{3(1-\zeta)}\right)^{\beta'} \log\left(\dfrac{4^{1-\zeta}(8\beta')^{\beta'}K_{\alpha,\beta,\zeta}/(1-\zeta)^{\beta'}}{\Delta_S^*}\right)^{\beta'}.
\end{equation*}
}We write $D_{\alpha,\beta,\zeta}={4^{1-\zeta}}(8\beta')^{\beta'}K_{\alpha,\beta,\zeta}{/(1-\zeta)^{\beta'}}$, $L_{\alpha,\beta,\zeta}=(4/(3(1-\zeta)))^{\beta'}K_{\alpha,\beta,\zeta}$, then we have
\begin{equation}
\delta_0^{1-\zeta} \geq \dfrac{\Delta_S^*}{L_{\alpha,\beta,\zeta}\log(D_{\alpha,\beta,\zeta}/\Delta_S^*)^{\beta'}}.
\label{eq:delta_Delta_lite_2}
\end{equation}
By definition of $i_0$ { that it is the largest integer $i$ that satisfies $\gamma_{i}\geq\delta_0$,} we also have
\begin{equation*}
\gamma_{i_0}^{1-\zeta} \geq \dfrac{\Delta_S^*}{L_{\alpha,\beta,\zeta}\log(D_{\alpha,\beta,\zeta}/\Delta_S^*)^{\beta'}}.
\end{equation*}
{ Recall that $\gamma_i=i^{-(1/\zeta-1)(1-\alpha)}$, we have
\begin{equation*}
\gamma_{i_0}^{1-\zeta}=i_0^{-(1-\zeta)(1/\zeta-1)(1-\alpha)}=i_0^{-(1-\zeta)^2(1-\alpha)/\zeta}\geq \dfrac{\Delta_S^*}{L_{\alpha,\beta,\zeta}\log(D_{\alpha,\beta,\zeta}/\Delta_S^*)^{\beta'}}.
\end{equation*}} 
Therefore
\begin{equation}
i_0\leq \left(\dfrac{L_{\alpha,\beta,\zeta}\log(D_{\alpha,\beta,\zeta}/\Delta_S^*)^{\beta'}}{\Delta_S^*}\right)^{\frac{\zeta}{(1-\zeta)^2(1-\alpha)}}.
\label{eq:i0_lite}
\end{equation}

We set 
\begin{equation}
k_0=\left\lfloor \left(\left(\frac{x/3}{2+\zeta+(2(1/\zeta-1)(1-\alpha))^{-1}}\right)^\zeta -i_0 -1\right)_+\right\rfloor,
\label{eq:k0_def}
\end{equation}
then for any round $k\leq k_0$, { recall that $\gamma_i=i^{-(1/\zeta-1)(1-\alpha)}$, then $t_i'=\lceil \gamma_i^{-1/(1-\alpha)}\rceil\leq1+i^{1/\zeta-1}$ and $T_i'=\lceil\gamma_i^{-2}\rceil\leq1+i^{2(1/\zeta-1)(1-\alpha)}$,} the cumulative satisficing regret in Step 1 and 2 can be upper bounded by { linear in $t_i'$ and $T_i'$ as}
\begin{equation}
\begin{aligned}
    \sum_{i=i_0+1}^{i_0+k}R_i&\leq\sum_{i=i_0+1}^{i_0+k}(t_i'+T_i')\leq 2k+\sum_{i=i_0+1}^{i_0+k}(i^{1/\zeta-1}+i^{2(1/\zeta-1)(1-\alpha)})\\
    &< 2k + \int_{i_0+{1}}^{i_0+k+1}(y^{1/\zeta-1}+y^{2(1/\zeta-1)(1-\alpha)})\mathrm{d}y\\
    &= 2k + \left(\zeta y^{1/\zeta} + \frac{1}{2(1/\zeta-1)(1-\alpha)+1}y^{2(1/\zeta-1)(1-\alpha)+1}\right)\bigg|_{i_0+{1}}^{i_0+k+1}\\
    &\overset{\text{(i)}}{\leq} 2k + \left(\left(\zeta + \frac{1}{2(1/\zeta-1)(1-\alpha)}\right)y^{1/\zeta}\right)\bigg|_{i_0+{1}}^{i_0+k+1}\\
    &\overset{\text{(ii)}}{\leq} { 2k^{1/\zeta} + \left(\zeta + \frac{1}{2(1/\zeta-1)(1-\alpha)}\right)(i_0+k+1)^{1/\zeta} } \\
    &\leq \left(2 + \zeta + \frac{1}{2(1/\zeta-1)(1-\alpha)}\right)(i_0+k+1)^{1/\zeta}\\
    &\overset{\text{(iii)}}{\leq} x/3,
\label{eq:tail_i0}
\end{aligned}
\end{equation}
where { step (i) is because $2(1/\zeta-1)(1-\alpha)+1\leq 1/\zeta$ for $1/2\leq\alpha<1$, step (ii) is because $y^{1/\zeta}\big|_{i_0+1}>0$ and $k\leq k^{1/\zeta}$ for $\zeta<1$ and $k\geq1$, and step (iii)} is based on $k<k_0$, {$(i_0+k+1)^{1/\zeta}$ increasing in $k$,} and the definition \eqref{eq:k0_def} of $k_0$. Therefore if the cumulative satisficing regret in Step 1 and 2 exceeds $x/3$, the total number of rounds should be at least $i_0+k_0+1$. Furthermore, by Proposition~\ref{prop:exit_prob_tail}~we know the probability that the total number of rounds exceeds $i_0+k_0$ is at most { $4^{-k_0}$}. Therefore, plug into the definition \eqref{eq:k0_def} of $k_0$, and the probability of exceeding $x/3$ in Step 1 and 2 can be further bounded as
\begin{equation}
\begin{aligned}
&\Pr\left(\sum_{i=i_0+1}^{\infty}R_i>\frac{x}{3} \right) \leq 4^{-k_0}\\
\leq & 4^{i_0+2} \cdot 4^{\left(-\left(\frac{x/3}{2+\zeta+(2(1/\zeta-1)(1-\alpha))^{-1}}\right)^\zeta\right)}  \\
\leq& \left(e^2\right)^{i_0+2} \cdot\exp\left(-\left(\frac{x/3}{2+\zeta+(2(1/\zeta-1)(1-\alpha))^{-1}}\right)^\zeta\right)\\
\leq& \exp\left(4+2i_0-\left(\frac{x}{9+(2(1-\zeta)/3)^{-1}(1-\alpha)^{-1}} \right)^\zeta \right),
\label{eq:tail_i0+k+}
\end{aligned}
\end{equation}
{ where the final inequality is because $2+\zeta<3$ and $\zeta^{-1}-1>1-\zeta$}.

Then we bound the probability that cumulative satisficing regret incurred in the Step 3 of all rounds exceeds $x/3$. For each round $i$, the probability that round $i$ is not terminated after pulling non-satisficing arm $\hat{A}_i${, \ie, $r(\hat{A}_i)<S$,} for $k_i$ times (in addition to the first $T_i'$ times) occurs with probability at most
\begin{equation}
\begin{aligned}
&\Pr\left(\dfrac{\hat{r}^{\text{tot}}_i}{T_i'+k_i}-\sqrt{\dfrac{k_i^\zeta+\log\left({16}\zeta^{-1}\Gamma(\zeta^{-1})\right)}{T_i'+k_i}}>S\right)\\
\leq &\Pr\left(\dfrac{\hat{r}^{\text{tot}}_i}{T_i'+k_i}-\sqrt{\dfrac{k_i^\zeta+\log\left({16}\zeta^{-1}\Gamma(\zeta^{-1})\right)}{T_i'+k_i}}>r(\hat{A}_i)\right)\\
\leq &\frac{\zeta}{16\Gamma(\zeta^{-1})}\exp\left(-k_i^\zeta\right),
\label{eq:LCB_light_high_prob}
\end{aligned}
\end{equation}
{ where the second inequality is based on Hoeffding's inequality.} The probability that satisficing regret incurred in the Step 3 in round $i$ exceeds $x/3$ can be bounded by
\begin{equation*}
    \Pr\left(\max\left\{S-r(\hat{A}_i),0 \right\}\cdot k_i>x/3\right)\leq \Pr\left(k_i>x/3\right) \leq \frac{\zeta}{16\Gamma(\zeta^{-1})} \exp\left(-(x/3)^\zeta\right),
\end{equation*}
{ where the first inequality is because $\max\{S-r(\hat{A}_i),0 \}\leq1$ based on reward $r\in[0,1]$ and $0\leq S\leq r$, the second inequality is by applying \eqref{eq:LCB_light_high_prob}.} Note that $\zeta/\Gamma(\zeta^{-1})\leq3/2$, using Proposition~\ref{lemma:round_exp_all}, the probability of total regret in LCB test exceeds $x/3$ can be bounded by
\begin{equation}
\begin{aligned}
&\Pr\left(\sum_{i=1}^{\infty}\max\left\{S-r(\hat{A}_i),0 \right\}\cdot k_i>x/3 \right)\\
\leq& \Pr\left(\sum_{i=1}^{i_0+N(1/4)}\max\left\{S-r(\hat{A}_i),0 \right\}\cdot k_i>x/3 \right)\\ 
\leq& {\frac{2^{i_0+1}(1-1/4)}{1-2\cdot1/4}\cdot\exp\left(-\frac{(x/3)^\zeta}{\zeta/(16\Gamma(\zeta^{-1}))+1} \right)} \\
\leq& { 2^{i_0+2}\cdot\exp\left(-\frac{x^\zeta/3}{3/16+1}\right)}\\
\leq& { \exp\left(i_0+2-\frac{x^\zeta}{4}\right)}
\label{eq:tail_LCB}
\end{aligned}
\end{equation}
{ where the first inequality is based on Proposition~\ref{prop:exit_prob_tail}, which says that for each round $i>i_0$, it ends and starts round $i+1$ with probability at most $1/4$. The second inequality is due to \cref{lemma:round_exp_all} where the total number of rounds follow a geometric distribution with success rate $3/4=1-1/4$ in addition to the initial $i_0$ rounds. The third inequality is because $\zeta<1$ and $\Gamma(\zeta^{-1})\geq 2/3$, thus $\zeta/16\Gamma(\zeta^{-1})\leq 3/16$. The fourth inequality is due to $3(3/16+1)\leq4$ and $2^{i_0+1}\leq e^{i_0+1}$.}

As we { assume $x>x_0=6i_0+6\zeta(i_0+1)^{1/\zeta}$, recall that $\gamma_i=i^{-(1/\zeta-1)(1-\alpha)}$, then $T_i'=\lceil\gamma_i^{-2}\rceil\leq t_i'$ and $t_i'=\lceil \gamma_i^{-1/(1-\alpha)}\rceil\leq1+i^{1/\zeta-1}$, then }
\begin{equation*}
\begin{aligned}
&\sum_{i=1}^{i_0}(t_i'+T_i')<\sum_{i=1}^{i_0}2t_i'<2i_0+\sum_{i=1}^{i_0}2i^{1/\zeta-1}\\
<&2i_0+2\int_1^{i_0+1}y^{1/\zeta-1}\mathrm{d}y=2i_0+2\zeta y^{1/\zeta}\bigg|_{1}^{i_0+1}\\
<&2i_0+2\zeta(i_0+1)^{1/\zeta}<x/3.
\end{aligned}
\end{equation*}
Therefore $x/3>\sum_{i=1}^{i_0}(t_i'+T_i')$, then the satisficing regret incurred before round $i_0$ in Step 1 and 2 { is} always bounded by $x/3$. Recall $i_0$ is a constant without $x$ or $T$, we can bound the tail of total regret { with a combination of \eqref{eq:tail_i0}, \eqref{eq:tail_i0+k+}, and \eqref{eq:tail_LCB} as } 
\begin{equation*}
\begin{aligned}
    \Pr\left(\texttt{Regret}_S>x\right) \leq& \Pr\left(\sum_{i=1}^{i_0}R_i>\frac{x}{3} \right) + \Pr\left(\sum_{i=i_0+1}^{\infty}R_i>\frac{x}{3} \right) + \Pr\left(\sum_{i=1}^{\infty}\max\left\{S-r(\hat{A}_i),0 \right\}\cdot k_i>x/3 \right)\\
    \leq& {\exp\left(4+2i_0-\left(\frac{x}{9+(2(1-\zeta)/3)^{-1}(1-\alpha)^{-1}} \right)^\zeta \right)} + \exp\left({ i_0 + 2 -\frac{x^\zeta}{4}} \right).
\end{aligned}
\end{equation*}
Finally, since $(x-x_0)_+\leq x$, by replacing $x$ with $(x-x_0)_+$ we obtain the probability bound in the theorem.

\subsection{Proof of Proposition \ref{prop:exit_prob_tail}}\label{sec:prop:exit_prob_tail}
By Condition~\ref{assump:learning_alg} { and similar to~\eqref{eq:single_pull_regret_std}, because the expression of $t_i'$ in $\gamma_i$ remains unchanged compared to that in Algorithm~\ref{alg:sat_exploration},} we still have
\begin{align*}
\E[r(A^*)-r(\hat{A}_i)]=\dfrac{1}{t_i'}\E\left[t_i'r(A^*)-\sum_{s=1}^{t_i'}r(A_s)\right]\leq\dfrac{1}{t_i'}C_1t_i'^\alpha\log(t_i')^\beta \leq \dfrac{2C_1}{(1-\alpha)^\beta}\gamma_i\log({4}/\gamma_i)^\beta,
\end{align*}
where the last inequality holds because $t_i'=\lceil1/\gamma{_i}^{1/(1-\alpha)}\rceil\leq 2/\gamma{_i}^{1/(1-\alpha)} {\leq 4/\gamma{_i}^{1/(1-\alpha)}}$.

According to the condition of this proposition, we have $i$ satisfies the condition $$\dfrac{32C_1(1-\zeta)^{-\zeta/2}}{(1-\alpha)^\beta} {\sqrt{2 + 2\log\left(16\zeta^{-1}\Gamma(\zeta^{-1})\right)}} \gamma_i^{1-\zeta} \log({4}/\gamma_i)^{\max\{\beta,1/2\}} \leq \Delta_S^*.$$ 
Note that for any $0<\zeta<1$, $(1-\zeta)^{-\zeta/2}>1$ and $\gamma_i^{1-\zeta}>\gamma_i$. Therefore $i$ satisfies
$$\dfrac{32C_1}{(1-\alpha)^\beta}\gamma_i\log({4}/\gamma_i)^{\max\{\beta,1/2\}}\leq \Delta_S^*.$$
By the Markov's inequality { and $\log(4/\gamma_i)\geq\log(4)>1$ for $\gamma_i<1$,} we have
\begin{equation}
\Pr(r(A^*)-r(\hat{A}_i)\geq \Delta_S^*/2)\leq \Pr\left(r(A^*)-r(\hat{A}_i)\geq \dfrac{16C_1}{(1-\alpha)^\beta}{ \gamma_i}\log({4}/\gamma_i)^\beta\right)\leq\dfrac{1}{8}.
\label{eq:prob_bound_1_tail}
\end{equation}

Now, if $r(\hat{A}_i)\geq r(A^*)-\Delta_S^*/2$, which happens with probability at least 7/8 according to the above inequality, then $r(\hat{A}_i)\geq S+\Delta_S^*/2$ holds. On the other hand, if $i$ satisfies the condition stated in the proposition, then
\begin{equation}
(1-\zeta)^{-\zeta/2}\cdot {\sqrt{2 + 2\log\left(16\zeta^{-1}\Gamma(\zeta^{-1})\right)}} \gamma_i^{1-\zeta}\log({4}/\gamma_i)^{1/2}\leq \Delta_S^*/16.
\label{eq:i0_simple_lite}
\end{equation}
Note that for $f(y)=y^\zeta/(T_i'+y)$, it reaches its maximum when 
$$
\frac{\mathrm{d}}{\mathrm{d}y}f(y) = \frac{\zeta y^{\zeta-1}T_i'+y^\zeta(\zeta-1)}{(T_i'+y)^2}=0.
$$
Therefore the maximum of $f(y)$ is reached at $\zeta T_i'/(1-\zeta)$, { and we get
\begin{equation}\label{eq:maximum_k}
f(y)\leq \dfrac{(\frac{\zeta T_i'}{1-\zeta})^\zeta}{T_i'+\frac{\zeta T_i'}{1-\zeta}}.
\end{equation}}
For simplicity, we set $D = \log\left({16}\zeta^{-1}\Gamma(\zeta^{-1})\right)$. { For $k=0$, the confidence radius is bounded by
$$
\sqrt{\frac{D}{T_i'}} \leq \sqrt{D}\gamma_i \leq 2\sqrt{D}\gamma_i^{1-\zeta}\log({4}/\gamma_i)^{1/2} \leq \frac{\Delta_S^*}{8},
$$
where the first inequality is because $T_i'=\lceil\gamma_i^{-2}\rceil\geq\gamma_i^{-2}$, the second inequality is because $\gamma_i<1$ and $2\log({4}/\gamma_i)^{1/2}>1$, and the final inequality is due to \eqref{eq:i0_simple_lite}. For $k\geq1$, the confidence radius} is bounded by
\begin{equation}
\begin{aligned}
\sqrt{\frac{k^\zeta+D}{T_i'+k}} \overset{\text{(i)}}{\leq}& \sqrt{\frac{{(D+1)}k^\zeta}{T_i'+k}} 
\overset{\text{(ii)}}{\leq} \sqrt{\frac{{(D+1)}\left(\frac{\zeta T_i'}{1-\zeta}\right)^\zeta}{T_i'+\left(\frac{\zeta T_i'}{1-\zeta}\right)}}\\
\leq&\sqrt{\frac{{(D+1)}(1-\zeta)^{-\zeta}\cdot\zeta^\zeta T_i'^\zeta}{T_i'}} \\
\overset{\text{(iii)}}{\leq}& {\sqrt{(D+1)}}(1-\zeta)^{-\zeta/2}\gamma_i^{1-\zeta}\\
\overset{\text{(iv)}}{\leq}& {2\sqrt{(D+1)}(1-\zeta)^{-\zeta/2}\gamma_i^{1-\zeta}\log({4}/\gamma_i)^{1/2}  \leq \frac{\Delta_S^*}{8}}.
\label{eq:CI<Delta/8}
\end{aligned}
\end{equation}
{ where inequality (i) is because $k^\zeta\geq1$ as $k$ is a positive integer, inequality (ii) follows from \eqref{eq:maximum_k}, inequality (iii) is because $\zeta^\zeta<1$ as $\zeta<1$ and $T_i' = \lceil\gamma_i^{-2}\rceil\geq\gamma_i^{-2}$ as $\gamma_i^{-1}=i^{(1/\zeta-1)(1-\alpha)}\geq1$, inequality (iv) is because $2\log({4}/\gamma_i)^{1/2}>1$ for any $\gamma_i<1$, and the final inequality is due to~\eqref{eq:i0_simple_lite}.} Suppose round $i$ is terminated for some $k$, then
\begin{equation*}
\dfrac{\hat{r}_i^{\text{tot}}}{T_i'+k} - \sqrt{\frac{k^\zeta+D}{T_i'+k}} <S=r(A^*)-\Delta_S^*\leq r(\hat{A}_i)-\dfrac{\Delta_S^*}{2}\leq r(\hat{A}_i) - 2\sqrt{\frac{k^\zeta+D}{T_i'+k}},
\end{equation*}
or equivalently
\begin{equation*}
\dfrac{\hat{r}_i^{\text{tot}}}{T_i'+k}\leq r(\hat{A}_i) - \sqrt{\frac{k^\zeta+D}{T_i'+k}}.
\end{equation*}
By the Hoeffding's inequality we have
\begin{equation*}
\Pr\left(\dfrac{\hat{r}_i^{\text{tot}}}{T_i'+k}\leq r(\hat{A}_i) - \sqrt{\frac{k^\zeta+D}{T_i'+k}}\right) \leq { \exp(-k^\zeta - D) =} \frac{\zeta}{{16}\Gamma(\zeta^{-1})}\exp\left(-k^\zeta \right).
\end{equation*}
Therefore conditioned on $r(\hat{A}_i)\geq r(A^*)-\Delta_S^*/2$ the probability that round $i$ is terminated by the end of the horizon is bounded by
\begin{equation}
\begin{aligned}
    \sum_{k={0}}^\infty\frac{\zeta}{16\Gamma(\zeta^{-1})} \exp\left(-k^\zeta \right) 
    &\leq { \frac{\zeta}{16\Gamma(\zeta^{-1})} + } \frac{\zeta}{{16}\Gamma(\zeta^{-1})} \int_0^\infty \exp\left(-x^\zeta \right)\mathrm{d}x \\
    &\leq {\frac{1}{16} + }\frac{\zeta}{16\Gamma(\zeta^{-1})}\cdot\zeta^{-1}\Gamma(\zeta^{-1}) =\frac{1}{8}.
\label{eq:prob_bound_2_tail}
\end{aligned}
\end{equation} Combining \eqref{eq:prob_bound_1_tail} and \eqref{eq:prob_bound_2_tail} we conclude that the probability that round $i$ terminates by the end of the time horizon is { upper} bounded by $1/4$.

\subsection{Proof of Proposition \ref{lemma:round_exp_all}}\label{sec:lemma:round_exp_all}

We define $Y_i=Q_i^{\zeta}$ for all $i$, then for all $i\in\mathbb{Z}^+$,
\begin{equation*}
\Pr(Y_i\geq x)=\Pr(Q_i\geq x^{1/\zeta})\leq K_1\exp(-x).
\end{equation*}
For all $i\in\mathbb{Z}^+$, we have
{
\begin{equation*}
\text{Pr}\left(\exp\left(\dfrac{Y_i}{K_1+1}\right)\geq x\right)=\text{Pr}(Y_i\geq (K_1+1)\log(x))\leq K_1x^{-K_1-1}.
\end{equation*}
Since $Q_i\geq 0$, we have $\exp(Y_i/(K_1+1))\geq 1$, therefore
\begin{equation}\label{eq:bound_mgf}
\begin{aligned}
\mathbb{E}\left[\exp\left(\dfrac{Y_i}{K_1+1}\right)\right]&=1+\int_{1}^\infty \Pr\left(\exp\left(\dfrac{Y_i}{K_1+1}\right)\geq x\right)\mathrm{d}x\\
&\leq 1+\int_{1}^\infty K_1x^{-K_1-1}\text{d}x=2.
\end{aligned}
\end{equation}
Let $N(\lambda)$ be a geometric random variable with success probability $1-\lambda$. Then we have
\begin{equation*}
\begin{aligned}
\mathbb{E}\left[\exp\left(\sum_{i=1}^{N(\lambda)}\dfrac{Y_i}{K_1+1}\right)\right]&\leq \sum_{n=1}^{\infty}\Pr(N(\lambda)=n) \cdot\E\left[\exp\left(\sum_{i=1}^{n}\dfrac{Y_i}{K_1+1}\right)\right]  \\
&\leq \sum_{n=1}^{\infty}\Pr(N(\lambda)=n) \cdot \prod_{i=1}^{n}\E\left[\exp\left(\dfrac{Y_i}{K_1+1}\right)\right]  \\
&\leq\sum_{n=1}^{\infty}\Pr(N(\lambda)=n)\cdot2^n\\
&=\sum_{n=1}^{\infty}\lambda^{n-1}(1-\lambda)\cdot2^n = \frac{1-\lambda}{\lambda}\sum_{n=1}^{\infty}(2\lambda)^n\\
&=\frac{2-2\lambda}{1-2\lambda}.
\end{aligned}
\end{equation*}
where the first inequality is by conditioning on all $n\geq1$, the second inequality is by the independence of $Y_i$, the third inequality is by \eqref{eq:bound_mgf}, and the final equality is based on the assumption that $\lambda<1/2$. We define $W=\sum_{i=1}^{N+N(\lambda)}Y_i$, then combining the above inequality with \eqref{eq:bound_mgf} and the independence of $y_i$, we have
\begin{align*}
\mathbb{E}\left[\exp\left(\dfrac{W}{K_1+1}\right)\right]=&\mathbb{E}\left[\exp\left(\sum_{i=1}^{N+N(\lambda)}\dfrac{Y_i}{K_1+1}\right)\right]\\
=&\left(\mathbb{E}\left[\exp\left(\dfrac{Y_1}{K_1+1}\right)\right]\right)^N\cdot \mathbb{E}\left[\exp\left(\sum_{i=1}^{N(\lambda)}\dfrac{Y_i}{K_1+1}\right)\right]\\
\leq&\dfrac{2^{N+1}(1-\lambda)}{1-2\lambda}.
\end{align*}
By Markov's inequality and the above inequality on $\E[\exp(W/(K_1+1))]$, we have
\begin{equation}
\begin{aligned}
\Pr(W\geq x)&=\Pr\left(\exp\left(\dfrac{W}{K_1+1}\right)\geq \exp\left(\dfrac{x}{K_1+1}\right)\right)\\
&\leq\mathbb{E}\left[\exp\left(\dfrac{W}{K_1+1}\right)\right]/\exp\left(\dfrac{x}{K_1+1}\right)\\
&=\dfrac{2^{N+1}(1-\lambda)}{1-2\lambda}\cdot \exp\left(-\dfrac{x}{K_1+1}\right).
\label{eq:P(W>x)}
\end{aligned}
\end{equation}
Since $0<\zeta<1$, we have
\begin{equation*}
W=\sum_{i=1}^{N+N(\lambda)}Y_i=\sum_{i=1}^{N+N(\lambda)} Q_i^{\zeta}\geq \left(\sum_{i=1}^{N+N(\lambda)}Q_i\right)^{\zeta}.
\end{equation*}
Therefore, plugging into \eqref{eq:P(W>x)}, we have
\begin{equation*}
\begin{aligned}
\Pr\left(\sum_{i=1}^{N+N(\lambda)}Q_i\geq x\right)&=\Pr\left(\left(\sum_{i=1}^{N+N(\lambda)}Q_i\right)^\zeta\geq x^\zeta\right)\\
&\leq \Pr(W\geq x^\zeta) \\
&\leq \dfrac{2^{N+1}(1-\lambda)}{1-2\lambda}\cdot \exp\left(-\dfrac{x^\zeta}{K_1+1}\right).
\end{aligned}
\end{equation*}
}

\section{Proof of Theorem \ref{thm:regret_bound_tail}}\label{sec:thm:regret_bound_tail}

We first use the following proposition to bound the expected satisficing regret of each round.

\begin{proposition}\label{prop:regret_bound_round_tail}
If $r(A^*)\geq S$, then the expected satisficing regret~incurred in round $i$ is bounded by
\begin{equation*}
\dfrac{{ 8}C_1}{(1-\alpha)^\beta}\gamma_i^{-\alpha/(1-\alpha)}\log({4}/\gamma_i)^{\beta}.
\end{equation*}
\end{proposition}

Proof of \cref{prop:regret_bound_round_tail} is provided in \cref{sec:prop:regret_bound_round_tail}. { We also recall the bound on $i_0$ is defined as the largest $i$ such that $K_{\alpha,\beta,\zeta}\gamma_i^{1-\zeta}\log({4}/\gamma_i)^{\max\{\beta,1/2\}}>\Delta_S^*$ where $K_{\alpha,\beta,\zeta}:=\frac{32C_1(1-\zeta)^{-\zeta/2}}{(1-\alpha)^\beta} {\sqrt{2 + 2\log\left(16\zeta^{-1}\Gamma(\zeta^{-1})\right)}}$. $i_0$ is upper bounded in \eqref{eq:i0_lite} that}
\begin{equation*}
 i_0\leq \left(\dfrac{L_{\alpha,\beta,\zeta}\log(D_{\alpha,\beta,\zeta}/\Delta_S^*)^{\beta'}}{\Delta_S^*}\right)^{\frac{\zeta}{(1-\zeta)^2(1-\alpha)}},
\end{equation*}
{ where $\beta'=\max\{\beta,1/2\}$, $D_{\alpha,\beta,\zeta}={4^{1-\zeta}}(8\beta')^{\beta'}K_{\alpha,\beta,\zeta}{/(1-\zeta)^{-\beta'}}$, and $L_{\alpha,\beta,\zeta}=(4/(3(1-\zeta)))^{\beta'}K_{\alpha,\beta,\zeta}$. Furthermore, recall that $\delta_0\in(0,1)$ is the smallest solution to Equation~\eqref{eq:delta_Delta_equation-lite} as}
\begin{equation*}
\delta_0^{1-\zeta}\log({4}/\delta_0)^{\beta'}=\dfrac{\Delta_S^*}{K_{\alpha,\beta,\zeta}},
\end{equation*}
{ then we know $\delta_0\leq \gamma_{i_0}=i_0^{-(1/\zeta-1)(1-\alpha)}$.}

Using the result of Proposition~\ref{prop:regret_bound_round_tail}, we have that the satisficing regret by round $i_0$ is bounded by
\begin{equation}
\begin{aligned}
&\sum_{i=1}^{i_0}\dfrac{{8}C_1}{(1-\alpha)^\beta}\gamma_i^{-\alpha/(1-\alpha)}\log({4/\gamma_i})^{\beta} \\
\leq& \sum_{i=1}^{i_0}\dfrac{{8}C_1}{(1-\alpha)^\beta} i^{\alpha(1-\zeta)/\zeta} \log({4}/\delta_0)^\beta\\
\leq&\dfrac{{8}C_1}{(1-\alpha)^\beta}i_0^{1+\alpha(1-\zeta)/\zeta}\log({4}i_0^{(1/\zeta-1)(1-\alpha)})^\beta \\
\leq& \frac{{8}C_1}{(1-\alpha)^\beta(1-\zeta){^\beta}} \left(\dfrac{L_{\alpha,\beta,\zeta}\log(D_{\alpha,\beta,\zeta}/\Delta_S^*)^{\beta'}}{\Delta_S^*}\right)^{\frac{\zeta+\alpha(1-\zeta)}{(1-\zeta)^2(1-\alpha)}} \log\left(\frac{{4}L_{\alpha,\beta,\zeta}\log(D_{\alpha,\beta,\zeta}/\Delta_S^*)^{\beta'}}{\Delta_S^*} \right)^\beta,
\label{eq:sat_regret_before_i0_lite}
\end{aligned}
\end{equation}
{ where the first inequality is based on the definition $\gamma_{i}=i^{-(1/\zeta-1)(1-\alpha)}$ and that $\gamma_i\geq \gamma_{i_0}\geq\delta_0$ for all $i\leq i_0$, the second inequality is because $i^{\alpha(1-\zeta)/\zeta}\leq i_0^{\alpha(1-\zeta)/\zeta}$ for any $i\in[i_0]$, and the third inequality is simply plugging into the upper bound on $i_0$ in \eqref{eq:i0_lite}.}

By the results of Proposition~\ref{prop:exit_prob_tail}, for every $k\geq 1$, the probability that round $i=i_0+k$ starts within the time horizon is at most $4^{-k+1}$. Therefore the regret incurred after round $i_0$ is bounded by
\begin{equation}
\begin{aligned}
&\sum_{k=1}^\infty \dfrac{{8}C_1}{(1-\alpha)^\beta} (i_0+k)^{\alpha(1-\zeta)/\zeta} \log({4}/\gamma_i)^\beta\cdot 4^{-k+1}\\
\leq & \dfrac{{8}C_1}{(1-\alpha)^\beta}\left(\sum_{k=1}^{i_0}(2i_0)^{\alpha(1-\zeta)/\zeta} \log({4}/\gamma_i)^\beta\cdot 4^{-k+1} + \sum_{k=i_0+1}^\infty(2k)^{\alpha(1-\zeta)/\zeta} \log({4}/\gamma_i)^\beta\cdot 4^{-k+1} \right)\\
\leq& \dfrac{{8}C_1}{(1-\alpha)^\beta}\left(i_0\cdot(2i_0)^{\alpha(1-\zeta)/\zeta} \log({4}i_0^{(1/\zeta-1)(1-\alpha)})^\beta + \sum_{k=1}^\infty(2k)^{\alpha(1-\zeta)/\zeta} \log({4}k^{(1/\zeta-1)(1-\alpha)})^\beta\cdot 4^{-k+1} \right)\\
\leq& \dfrac{{8}C_1\cdot 2^{\alpha(1-\zeta)/\zeta}}{(1-\alpha)^\beta{(1-\zeta)^\beta}} \left(\dfrac{L_{\alpha,\beta,\zeta}\log(D_{\alpha,\beta,\zeta}/\Delta_S^*)^{\beta'}}{\Delta_S^*}\right)^{\frac{\zeta+\alpha(1-\zeta)}{(1-\zeta)^2(1-\alpha)}} \log\left(\frac{{4}L_{\alpha,\beta,\zeta}\log(D_{\alpha,\beta,\zeta}/\Delta_S^*)^{\beta'}}{\Delta_S^*} \right)^\beta + M,
\label{eq:sat_regret_after_i0_lite}
\end{aligned}
\end{equation}
where $M={\frac{8C_1}{(1-\alpha)^\beta}}\sum_{k=1}^\infty(2k)^{\alpha(1-\zeta)/\zeta} \log({4}k^{(1-\alpha)(1-\zeta)/\zeta})^\beta\cdot 4^{-k+1}$ is a constant only dependent on $\alpha,\beta,\zeta$ because $4^{-k+1}$ term decreases exponentially although $k^{\alpha(1-\zeta)/\zeta}$ grows polynomially, and the final inequality is again simply plugging into the upper bound on $i_0$. Then, { combining the satisficing regret before round $i_0$ \eqref{eq:sat_regret_before_i0_lite} and after round $i_0$ \eqref{eq:sat_regret_after_i0_lite},} the total satisficing regret is bounded by
\begin{equation}
\begin{aligned}
    \mathbb{E}[\texttt{Regret}_S]\leq M + &{\dfrac{C_1}{(1-\alpha)^\beta(1-\zeta)^\beta} \left(2^{\alpha(1-\zeta)/\zeta} + 2 \right)} \left(\dfrac{L_{\alpha,\beta,\zeta}}{\Delta_S^*}\right)^{\frac{\zeta+\alpha(1-\zeta)}{(1-\zeta)^2(1-\alpha)}} \\
    &\cdot\log(D_{\alpha,\beta,\zeta}/\Delta_S^*)^{\frac{\beta'(\zeta+\alpha(1-\zeta))}{(1-\zeta)^2(1-\alpha)}} \log\left(\frac{{4}L_{\alpha,\beta,\zeta}\log(D_{\alpha,\beta,\zeta}/\Delta_S^*)^{\beta'}}{\Delta_S^*} \right)^\beta.
\label{eq:main_bound_const_lite}
\end{aligned}
\end{equation}

Furthermore, since the length of the time horizon is $T$ time steps, we have 
$$
T \geq \sum_{i=1}^{i_{\max}{-1}}t_i' \geq \sum_{i=1}^{i_{\max}{-1}}i^{1/\zeta-1}\geq \int_0^{i_{\max}{-1}}x^{1/\zeta-1}\mathrm{d}x = \zeta (i_{\max}{-1})^{1/\zeta},
$$
thus the maximum number of rounds is bounded by {$i_{\max}\leq1+\lfloor(T/\zeta)^\zeta\rfloor=\lceil(T/\zeta)^\zeta\rceil$ because $i_{\max}$ is an integer. Recall that $\gamma_{i}=i^{-(1/\zeta-1)(1-\alpha)}$ is decreasing in $i$, for any $i\leq i_{\max}$, we have
\begin{equation}
\begin{aligned}
\gamma_i\geq&\gamma_{i_{\max}} = i_{\max}^{-(1/\zeta-1)(1-\alpha)}\\
\geq&(2^{1/\zeta}T/\zeta)^{-(1-\zeta)(1-\alpha)}\\
\geq&(2^{1/\zeta}T/\zeta)^{-1},
\label{eq:gammai_1/T_lite}
\end{aligned}
\end{equation}
where the second inequality is because $i_{\max}\leq\lceil(T/\zeta)^\zeta\rceil\leq(T/\zeta)^\zeta+1\leq2(T/\zeta)^\zeta$ based on $T/\zeta\geq1$ for $T\geq1$ and $\zeta<1$, and third inequality is because $\zeta<1$ and $1/2\leq\alpha<1$. So, $4/\gamma_i\leq2^{2+1/\zeta}T/\zeta$.} By the results of Proposition \ref{prop:regret_bound_round_tail}, we have that the satisficing regret is bounded by
\begin{equation}
\begin{aligned}
\mathbb{E}[\texttt{Regret}_S]\leq&\sum_{i=1}^{i_{\max}}\dfrac{{8}C_1}{(1-\alpha)^\beta}\gamma_i^{-\alpha/(1-\alpha)}\log({4}/\gamma_i)^\beta \\
\overset{\text{(i)}}{\leq}& \sum_{i=1}^{i_{\max}} \dfrac{{8}C_1}{(1-\alpha)^\beta}i^{\alpha(1-\zeta)/\zeta}\log({2^{2+1/\zeta}T/\zeta})^\beta\\
\leq & \dfrac{{8}C_1}{(1-\alpha)^\beta}\log({2^{2+1/\zeta}T/\zeta})^\beta \int_{1}^{i_{\max}+1} y^{\alpha(1-\zeta)/\zeta}\mathrm{d}y \\
= & {\dfrac{8C_1}{(1-\alpha)^\beta(\alpha(1-\zeta)/\zeta+1)}y^{\alpha(1-\zeta)/\zeta+1}\bigg|^{i_{\max}+1}_1 \log({2^{2+1/\zeta}T/\zeta})^\beta}\\
\leq & \dfrac{{8}C_1}{(1-\alpha)^\beta(\alpha(1-\zeta)/\zeta+1)}(i_{\max}+1)^{\alpha(1-\zeta)/\zeta+1} \log({2^{2+1/\zeta}T/\zeta})^\beta\\
\overset{\text{(ii)}}{\leq} & \dfrac{{8}C_1\zeta^{-\alpha(1-\zeta)-\zeta}\cdot{4}^{\alpha(\zeta^{-1}-1)+1}}{(1-\alpha)^\beta(\alpha(\zeta^{-1}-1)+1)}\,T^{\alpha(1-\zeta)+\zeta}\log({2^{2+1/\zeta}T/\zeta})^\beta.
\label{eq:main_bound_sublinear_lite}
\end{aligned}
\end{equation}
{ where step (i) is based on \eqref{eq:gammai_1/T_lite} and step (ii) is because $i_{\max}+1\leq2i_{\max}\leq2(T/\zeta)^\zeta+2\leq4(T/\zeta)^\zeta$ because $i_{\max}\geq1$ and $(T/\zeta)^\zeta\geq1$.

Combining the two parts of results, \ie, the constant satisficing regret \eqref{eq:main_bound_const_lite} and the sub-linear satisficing regret \eqref{eq:main_bound_sublinear_lite}, we have completed the proof of Theorem~\ref{thm:regret_bound_tail} by putting all constant terms into the $\text{polylog}$ term}.

\subsection{Proof of Proposition \ref{prop:regret_bound_round_tail}}\label{sec:prop:regret_bound_round_tail}
Following the proof pattern of Proposition \ref{prop:regret_bound_round_1}, as the expression of $t_i'$ in $\gamma_i$ remains $t_i'=\lceil \gamma_i^{-1/(1-\alpha)}\rceil$, { similar to \eqref{eq:Step1_Regret_real},} the satisficing regret incurred when running $\texttt{ALG}(t_i')$, {\ie, satisficing regret incurred in Step 1,} is still bounded { with Condition~\ref{assump:learning_alg}} by
\begin{align*}
\E\left[\sum_{s=1}^{t_i'}\max\{S-r(A_s),0\}\right]\leq\E\left[t_i'r(A^*)-\sum_{s=1}^{t_i'}r(A_s)\right]\leq \dfrac{2C_1}{(1-\alpha)^\beta}\gamma_i^{-\alpha/(1-\alpha)}\log({4}/\gamma_i)^\beta.
\end{align*}
Since $\hat{A}_i$ is selected randomly according to the trajectory of $\texttt{ALG}(t_i')$ in round $i$, { then similar to \eqref{eq:single_pull_regret},} we still have
\begin{align*}
\E[\max\{S-r(\hat{A}_i),0\}]\leq\E[r(A^*)-r(\hat{A}_i)] \leq \dfrac{2C_1}{(1-\alpha)^\beta}\gamma_i\log({4}/\gamma_i)^\beta.
\end{align*}
Therefore the satisficing regret incurred when pulling $\hat{A}_i$ for $T_i'=\lceil\gamma_i^{-2}\rceil$ times { in Step 2, similar to \eqref{eq:Step2_Regret_real},} is still bounded by
\begin{equation*}
\left\lceil\dfrac{1}{\gamma_i^2}\right\rceil\cdot\dfrac{2C_1}{(1-\alpha)^\beta}\gamma_i\log({4}/\gamma_i)^\beta\leq\dfrac{{4}C_1}{(1-\alpha)^\beta}\gamma_i^{-1}\log({4}/\gamma_i)^\beta\leq \dfrac{{4}C_1}{(1-\alpha)^\beta}\gamma_i^{-\alpha/(1-\alpha)}\log({4}/\gamma_i)^\beta.
\end{equation*}
{ Recall \eqref{eq:LCB_light_high_prob},} if $r(\hat{A}_i)<S$, then for every $k$, round $i$ is not terminated after pulling arm $\hat{A}_i$ for $k$ times (in addition to the first $T_i'$ times) occurs with probability at most
\begin{equation}
\begin{aligned}
\nonumber&\Pr\left(\dfrac{\hat{r}^{\text{tot}}_i}{T_i'+k} -\sqrt{\frac{k^\zeta+\log\left({16}\zeta^{-1}\Gamma(\zeta^{-1})\right)}{T_i'+k}} > S\right)\\
\nonumber\leq &\Pr\left(\dfrac{\hat{r}^{\text{tot}}_i}{T_i'+k} - \sqrt{\frac{k^\zeta+\log\left({16}\zeta^{-1}\Gamma(\zeta^{-1})\right)}{T_i'+k}} > r(\hat{A}_i)\right)\\
\leq&\exp\left(-k^\zeta + \log\left(\frac{\zeta}{{16}\Gamma(\zeta^{-1})} \right) \right) = \frac{\zeta}{{16}\Gamma(\zeta^{-1})}\exp\left(-k^\zeta \right).
\label{eq:radius_bound}
\end{aligned}
\end{equation}
Therefore, { similar to \eqref{eq:prob_bound_2_tail},} expected number of arm pulls of $\hat{A}_i$ after first pulling it $T_i'$ times, { \ie, the expected number of pulls in Step 3,} is bounded by
\begin{equation*}
\begin{aligned}
    &\E\left[\text{Number of times }\hat{A}_i\text{ is pulled in round }i\text{ after the first }T_i' \text{ times}\right]\\
    =&\sum_{s=1}^{\infty}\Pr(k\geq s) =\frac{\zeta}{{16}\Gamma(\zeta^{-1})}\sum_{s=1}^\infty \exp\left(-s^\zeta \right) \\
    \leq& \frac{\zeta}{{16}\Gamma(\zeta^{-1})} \int_0^\infty \exp\left(-x^\zeta \right)\mathrm{d}x\\
    \leq& \frac{\zeta}{{16}\Gamma(\zeta^{-1})}\cdot\zeta^{-1}\Gamma(\zeta^{-1}) =\frac{1}{16}\leq 1.
\end{aligned}
\end{equation*}
Summing up the satisficing regret { from Step 1, Step 2, and Step 3,} we have that the satisficing regret of round $i$ is still bounded by
\begin{equation*}
\dfrac{{8}C_1}{(1-\alpha)^\beta}\gamma_i^{-\alpha/(1-\alpha)}\log({4}/\gamma_i)^{\beta}.
\end{equation*}
{ Note that, this is exactly the same as \eqref{eq:prop_round_reg} in Proposition~\ref{prop:regret_bound_round_1}}

\section{Proof of Theorem \ref{thm:regret_nonsat_tail}}\label{sec:thm:regret_nonsat_tail}
We first prove that the regret incurred in round $i$ is at most
\begin{equation*}
\dfrac{{8}C_1}{(1-\alpha)^\beta}\gamma_i^{-\alpha/(1-\alpha)}\log({4}/\gamma_i)^{\beta}.
\end{equation*}

{ For Step 1,} the regret incurred when running $\texttt{ALG}(t_i')$ is bounded by
\begin{equation}
\E\left[t_i'r(A^*)-\sum_{s=1}^{t_i'}r(A_s)\right]\leq C_1t_i'^\alpha\log(t_i')^\beta\leq \dfrac{2C_1}{(1-\alpha)^\beta}\gamma_i^{-\alpha/(1-\alpha)}\log({4}/\gamma_i)^\beta.
\label{eq:Step1_std_lite}
\end{equation}
{ The same as \eqref{eq:single_pull_regret_std}}, we further have
\begin{equation}
\E[r(A^*)-r(\hat{A}_i)]\leq \dfrac{2C_1}{(1-\alpha)^\beta}\gamma_i\log({4}/\gamma_i)^\beta.
\label{eq:single_pull_regret_std_lite}
\end{equation}
Therefore, { for Step 2, the same as \eqref{eq:Step2_Regret_real}}, the expected standard regret incurred when pulling $\hat{A}_i$ is bounded by
\begin{equation}
\left\lceil\dfrac{1}{\gamma_i^2}\right\rceil\cdot\dfrac{2C_1}{(1-\alpha)^\beta}\gamma_i\log({4}/\gamma_i)^\beta {\leq} \dfrac{{4}C_1}{(1-\alpha)^\beta}\gamma_i^{-1}\log({4}/\gamma_i)^\beta \leq \dfrac{{4}C_1}{(1-\alpha)^\beta}\gamma_i^{-\alpha/(1-\alpha)}\log({4}/\gamma_i)^\beta.
\label{eq:Step2_std_lite}
\end{equation}
Since $r(\hat{A}_i)\leq r(A^*)<S$, for every $k$, round $i$ is not terminated with probability at most
\begin{equation*}
\begin{aligned}
\nonumber&\Pr\left(\dfrac{\hat{r}^{\text{tot}}_i}{T_i'+k} -\sqrt{\frac{k^\zeta+\log\left({16}\zeta^{-1}\Gamma(\zeta^{-1})\right)}{T_i'+k}} > S\right)\\
\nonumber\leq &\Pr\left(\dfrac{\hat{r}^{\text{tot}}_i}{T_i'+k} - \sqrt{\frac{k^\zeta+\log\left({16}\zeta^{-1}\Gamma(\zeta^{-1})\right)}{T_i'+k}} > r(\hat{A}_i)\right)\\
\leq&\exp\left(-k^\zeta + \log\left(\frac{\zeta}{{16}\Gamma(\zeta^{-1})} \right) \right) = \frac{\zeta}{{16}\Gamma(\zeta^{-1})}\exp\left(-k^\zeta \right).
\end{aligned}
\end{equation*}
Therefore, expected number of arm pulls of $\hat{A}_i$ after pulling $T_i'$ times is bounded by
\begin{equation*}
\begin{aligned}
    &\E\left[\text{Number of times }\hat{A}_i\text{ is pulled in round }i\right]\\
    =&\sum_{s=1}^{\infty}\Pr(k\geq s) =\frac{\zeta}{{16}\Gamma(\zeta^{-1})}\sum_{s=1}^\infty \exp\left(-s^\zeta \right) \leq \frac{\zeta}{{16}\Gamma(\zeta^{-1})} \int_0^\infty \exp\left(-x^\zeta \right)\mathrm{d}x\\
    \leq& \frac{\zeta}{{16}\Gamma(\zeta^{-1})}\cdot\zeta^{-1}\Gamma(\zeta^{-1}) =\frac{1}{16}\leq 1.
\end{aligned}
\end{equation*}
{ Then, the expected standard regret incurred in Step 3 is bounded by
\begin{equation}
1\cdot \E[r(A^*)-r(\hat{A}_i)]\leq \dfrac{2C_1}{(1-\alpha)^\beta}\gamma_i\log({4}/\gamma_i)^\beta\leq\dfrac{2C_1}{(1-\alpha)^\beta}\gamma_i^{-\alpha/(1-\alpha)}\log({4}/\gamma_i)^\beta,
\label{eq:Step3_std_lite}
\end{equation}
where the first inequality is based on \eqref{eq:single_pull_regret_std_lite} and the second inequality is based on $\gamma_i<1$.}

Summing up the { standard regret in Step 1 \eqref{eq:Step1_std_lite}, 2 \eqref{eq:Step2_std_lite}, and 3 \eqref{eq:Step3_std_lite}}, we have that the standard regret~of round $i$ is bounded by
\begin{equation*}
\dfrac{{8}C_1}{(1-\alpha)^\beta}\gamma_i^{-\alpha/(1-\alpha)}\log({4}/\gamma_i)^{\beta}.
\end{equation*}
Since the length of the time horizon is $T$ time steps, we have 
$$
T \geq \sum_{i}^{i_{\max}}t_i' \geq \sum_{i}^{i_{\max}}i^{1/\zeta-1}\geq \int_0^{i_{\max}}x^{1/\zeta-1}\mathrm{d}x = \zeta i_{\max}^{1/\zeta},
$$
thus the maximum number of rounds is bounded by $i_{\max}{\leq\lfloor(T/\zeta)^\zeta\rfloor}$, { which implies that $i_{\max}+1\leq2i_{\max}\leq2(T/\zeta)^\zeta$ because $i_{\max}\geq1$. Recall from \eqref{eq:gammai_1/T_lite} that $4/\gamma_i\leq2^{2+1/\zeta}T/\zeta$ and follow the same chain of inequalities in \eqref{eq:main_bound_sublinear_lite},} we have that the regret is bounded by
\begin{align*}
\mathbb{E}[\texttt{Regret}]\leq&\sum_{i=1}^{i_{\max}}\dfrac{{8}C_1}{(1-\alpha)^\beta}\gamma_i^{-\alpha/(1-\alpha)}\log({4}/\gamma_i)^\beta\\
\leq& \sum_{i=1}^{i_{\max}} \dfrac{{8}C_1}{(1-\alpha)^\beta}i^{\alpha(1-\zeta)/\zeta}\log({2^{2+1/\zeta}T/\zeta})^\beta\\
\leq & \dfrac{{8}C_1}{(1-\alpha)^\beta}\log({2^{2+1/\zeta}T/\zeta})^\beta \int_{1}^{i_{\max}+1} y^{\alpha(1-\zeta)/\zeta}\mathrm{d}y \\
= & { \dfrac{{8}C_1}{(1-\alpha)^\beta(\alpha(1-\zeta)/\zeta+1)}y^{\alpha(1-\zeta)/\zeta+1}\bigg|_1^{i_{\max}+1} \log({2^{2+1/\zeta}T/\zeta})^\beta} \\
\leq& \dfrac{{8}C_1}{(1-\alpha)^\beta(\alpha(1-\zeta)/\zeta+1)}(i_{\max}+1)^{\alpha(1-\zeta)/\zeta+1} \log({2^{2+1/\zeta}T/\zeta})^\beta\\
\leq & \dfrac{{8}C_1\zeta^{-\alpha(1-\zeta)-\zeta}\cdot{4}^{\alpha(\zeta^{-1}-1)+1}}{(1-\alpha)^\beta(\alpha(\zeta^{-1}-1)+1)}\cdot { 2^{\alpha(1-\zeta)/\zeta+1}}\cdot T^{\alpha(1-\zeta)+\zeta}\log({2^{2+1/\zeta}T/\zeta})^\beta.
\end{align*}
Therefore the expected regret is bounded by $C_1T^{\alpha(1-\zeta)+\zeta}\cdot\textup{polylog}(T)$.

\section{Tail Risk Control for Satisficing in Finite-Armed Bandits}\label{sec:light-tail-KArm}

In \cref{sec:satisficing_tail}, we established \satexlt, which is able to attain a constant expected satisficing regret and a light-tailed satisficing regret distribution in the realizable case. However, in the non-realizable case, the standard regret distribution of \satexlt~may not be light-tailed. In this section, we focus on the finite-armed bandit setting and establish an algorithm that: 1) attains a constant expected satisficing regret and a light-tailed satisficing regret distribution in the realizable case; 2) attains a sub-linear expected standard regret and a light-tailed standard regret distribution in the non-realizable case.

Similar to Condition~\ref{assump:learning_alg}, we first assume access to a learning oracle that attains a sub-linear expected standard regret and a light-tailed standard regret distribution for finite-armed bandits.
\begin{condition}\label{assump:learning_alg_tail}
There { exists} a finite-armed bandit algorithm $\texttt{ALG}(t)$, functions $\Lambda(K,t),\Lambda'(K,t)$ and $0<\zeta<1$ such that for any instance of $K$-arm bandits and any time horizon $t$,
\begin{enumerate}
\item The expected standard regret of $\texttt{ALG}(t)$ is upper bounded by $C_1K^{\alpha_1}t^{\alpha_2}\log(t)^\beta$, where $C_1\geq 1,\alpha_1\geq 0,1/2\leq\alpha_2\leq 1,\beta\geq 0$ are constants independent from $K$ and $t$;
\item The probability that the standard regret of $\texttt{ALG}(t)$ exceeds $x$ is upper bounded by $\Lambda(K,t)\exp(-x^\zeta/\Lambda'(K,t))$;
\item $\Lambda(K,t)$ can be at most polynomially dependent on $t$ and $\Lambda'(K,t)$ can be at most poly-logarithmically dependent on $t$.
\end{enumerate}
\end{condition}
Without loss of generality we also assume that $\Lambda(K,t)$ and $\Lambda'(K,t)$ are monotonically increasing in $t$ for every $K$, and is monotonically decreasing in $x$ for every fixed $t$ and $K$. Condition~\ref{assump:learning_alg_tail} can be satisfied by existing algorithms, such as the one established in \cite{simchi2024simple}.

For finite-armed bandits, we notice that the oracle algorithm is able to output the candidate arm that asymptotically converges to the optimal arm by choosing the one that is the most frequently pulled. Because the arm is already been pulled by at least $t_i/K$ times, leveraging this history already yields a tightened confidence interval for it, making forced sampling unnecessary. This holds for both realizable and non-realizable cases.

Therefore, we make three modifications to \satex~to make both satisficing regret distribution and standard regret distribution light-tailed for finite-armed bandits with the help of a light-tailed oracle algorithm. First, we get rid of forced sampling and directly start LCB test after running the oracle. Second, we use the arm that is pulled most frequently during $\texttt{ALG}(t_i)$ as the candidate arm, and the historical data of this arm being pulled when running $\texttt{ALG}(t_i)$ can be used in LCB test. Third, we use larger confidence intervals in the LCB test the same as the one in \satexlt. The complete algorithm is provided in \cref{alg:sat_lite_KArm}.

\begin{algorithm}[!ht]
\caption{Light-Tailed Satisficing Regret Minimization for Finite-Armed Bandits (\satexltp)}
\label{alg:sat_lite_KArm}
\begin{algorithmic}
\SingleSpacedXI
\State \textbf{Input:} time horizon $T$, satisficing level $S$, tail distribution parameter $\zeta$
\State Set $\gamma_i\leftarrow 2^{-i(1-\alpha_2)/\alpha_2}$ for all $i=1,2,3\dots$
\For{round $i=1,2,\dots$}
\State Set $t_i\leftarrow\lceil \gamma_i^{-1/(1-\alpha_2)}\rceil$
\State Run $\texttt{ALG}(t_i)$, let $\hat{A}_i$ denote the arm that is pulled most frequently along the trajectory during $\texttt{ALG}(t_i)$.
\State Set $\tau_i$ be the number of times $\hat{A}_i$ is pulled during $\texttt{ALG}(t_i)$, $k\leftarrow 0$.
\State Set $\hat{r}_i^{\text{tot}}$ be the sum of observed reward of pulling $\hat{A}_i$ for $\tau_i$ times during $\texttt{ALG}(t_i)$.
\State Set $LCB(\hat{A}_i)\leftarrow \hat{r}_i^{\text{tot}}/\tau_i-\sqrt{\log\left({16}\zeta^{-1}\Gamma(\zeta^{-1})\right)/\tau_i}$
\While{$LCB(\hat{A}_i)\geq S$}
\State Set $k\leftarrow k+1$
\State Pull arm $\hat{A}_i$ and observe $Y_t$, where $t$ is the current time step
\State Update $\hat{r}_i^{\text{tot}}\leftarrow \hat{r}_i^{\text{tot}}+Y_t$
\State Set
\begin{equation*}
LCB(\hat{A}_i)\leftarrow\dfrac{\hat{r}_i^{\text{tot}}}{\tau_i+k}-\sqrt{\frac{2k^\zeta+\log\left({16}\zeta^{-1}\Gamma(\zeta^{-1})\right)}{\tau_i+k}}
\end{equation*}
\EndWhile
\EndFor
\end{algorithmic}
\end{algorithm}

With the help of a finite-armed bandit learning algorithm with light-tailed standard regret, we use the following theorem to show that \satexltp~can enjoy a light-tailed satisficing regret distribution in the realizable case and a light-tailed standard regret distribution in the non-realizable case.
\begin{theorem}
If $r(A^*)\geq S$, then under Condition~\ref{assump:learning_alg_tail}, for any $x>0$ the probability that satisficing regret of \satexltp~ exceeds $x$ is bounded by
\begin{align*}
    &\Pr\left(\texttt{Regret}_S>x\right)\\
    \leq&(\alpha_2\log_2(T)+1)\Lambda(K,T)\exp\left(\dfrac{-x^\zeta}{2{^\zeta}(\alpha_2\log_2(T)+1){^\zeta}\Lambda'(K,T)}\right) + \exp\left(i_0{+1-\frac{x^\zeta}{3}} \right),
\end{align*}
where $i_0=O\left(\log_2\left(1/\Delta_S^*\cdot\log(1/\Delta_S^*)^{\beta'} \right)\right)$ is a constant independent of $x$.

If $r(A^*)<S$ and the oracle algorithm $\texttt{ALG}$ is light-tailed, then the probability that standard regret of \satexltp~ exceeds $x$ is bounded by
\begin{align*}
    &\Pr\left(\texttt{Regret}>x\right)\\
    \leq&(\alpha_2\log_2(T)+1)\Lambda(K,T)\exp\left(\dfrac{-x^\zeta}{2^{ \zeta}(\alpha_2\log_2(T)+1)\Lambda'(K,T)}\right)^{\zeta} + 2T^{\alpha_2}\exp\left(-(x/2)^\zeta\right).
\end{align*}
\label{thm:light_tail_KArm}
\end{theorem}

Proof of Theorem~\ref{thm:light_tail_KArm} is provided in Section~\ref{sec:thm:light_tail_KArm}. In the next theorem, we show that similar to \satex, the satisficing regret of \satexltp~can be bounded by a constant in the realizable case. 
\begin{theorem}\label{thm:regret_bound_KArm}
If $r(A^*)\geq S$, then the satisficing regret of \satexltp~is bounded by
\begin{equation*}
\mathbb{E}[\texttt{Regret}_S]\leq\min\left\{ C_1\left(\dfrac{C_1K^{\alpha_1+1}}{\Delta_S^*}\right)^{\frac{\alpha_2}{(1-\zeta)(1-\alpha_2)}}\cdot\textup{polylog}\left(\dfrac{C_1K^{\alpha_1+1}}{\Delta_S^*}\right),\,C_1T^{\alpha_2}\cdot\textup{polylog}(T)\right\}.
\end{equation*}
\end{theorem}
Proof of \cref{thm:regret_bound_KArm} is provided in Section~\ref{sec:thm:regret_bound_KArm}. The proof relies on two critical results. In Proposition~\ref{prop:regret_bound_round_KArm}, we show an upper bound on the satisficing regret of round $i$. The proof is provided in Section~\ref{sec:prop:regret_bound_round_KArm} of the appendix. In Proposition~\ref{prop:exit_prob_KArm}, we also show that once $\satexltp~$ runs for enough rounds, it is unlikely to start a new round. The proof is provided in Section~\ref{sec:prop:exit_prob_KArm} of the appendix.
\begin{proposition}\label{prop:regret_bound_round_KArm}
If $r(A^*)\geq S$, then the expected satisficing regret~incurred in round $i$ is { upper} bounded by
\begin{equation*}
\dfrac{{4}C_1K^{\alpha_1+1}}{(1-\alpha_2)^\beta}\gamma_i^{-\alpha_2/(1-\alpha_2)}\log({4}/\gamma_i)^\beta.
\end{equation*}
\end{proposition}

\begin{proposition}\label{prop:exit_prob_KArm}
If $r(A^*)\geq S$, then round $i$ terminates within the time horizon with probability at most $1/4$ if $i$ that satisfies
\begin{equation}
\dfrac{32C_1K^{\alpha_1+1}(1-\zeta)^{-\zeta/2}}{(1-\alpha_2)^\beta} {\sqrt{2+ 4\log\left({16}\zeta^{-1}\Gamma(\zeta^{-1})\right)}} \gamma_i^{1-\zeta} \log({4}/\gamma_i)^{\max\{\beta,1/2\}}\leq \Delta_S^*\,.
\label{eq:i0_KArm_lite_def}
\end{equation}
\end{proposition}

We also show that in the non-realizable case, \satexltp~enjoys the same standard regret bound as the oracle shown in Theorem~\ref{thm:regret_nonsat_KArm}. The proof is provided in Section~\ref{sec:thm:regret_nonsat_KArm} of the appendix.
\begin{theorem}
If $r(A^*)<S$, the expected standard regret of \satexltp is bounded by
$$
C_1K^{\alpha_1+1}T^{\alpha_2}\cdot\textup{polylog}(T).
$$
\label{thm:regret_nonsat_KArm}
\end{theorem}

\subsection{Proof of Theorem \ref{thm:light_tail_KArm}}\label{sec:thm:light_tail_KArm}
For notational brevity we define $p(x;K,t)=\Lambda(K,t)\exp(-x^\zeta/\Lambda'(K,t))$. We first bound the tail probability of running the oracle. We define the satisficing regret incurred in the oracle algorithm in each round $i$ as $R_i$. { Since by definition of $t_i=\lceil2^{i/\alpha_2}\rceil$, the number of rounds that end within the time horizon is bounded by $2^{i/\alpha_2}\leq T$, \ie, $i\leq\lfloor\alpha_2\log_2(T) \rfloor$ because $i$ is an integer. Then, the maximum number of rounds is bounded by $i_{\max}=\lceil\alpha_2\log_2(T)\rceil$, and} the probability that cumulative satisficing regret incurred in oracle algorithm exceeds $x/2$ is
\begin{equation}
\begin{aligned}
\Pr\left(\sum_{i=1}^{i_{\max}}R_i > \frac{x}{2} \right) \leq& \sum_{i=1}^{i_{\max}}\Pr\left(R_i > \frac{x}{2i_{\max}} \right) \\
\leq & \sum_{i=1}^{i_{\max}}p\left(\frac{x}{2i_{\max}};\,K,t_i \right) \\
\leq & i_{\max}p\left(\frac{x}{2i_{\max}};\,K,T \right)\\
\leq & (\alpha_2\log_2(T)+1)p\left(\frac{x}{2(\alpha_2\log_2(T)+1)};\,K,T \right)
\label{eq:oracle_tail_KArm}
\end{aligned}
\end{equation}
The third inequality is because $p(x;\,K,t)$ is monotonically increasing in $t$, and the fourth inequality is because $p(x;\,K,t)$ is monotonically decreasing in $x$.

We then bound the tail probability of LCB test. This is divided into realizable and non-realizable cases. For the realizable case, the probability that satisficing regret incurred in the LCB test in round $i$ exceeds $x/2$ can be bounded by
\begin{equation*}
\begin{aligned}
&\Pr\left(\max\left\{S-r(\hat{A}_i),0 \right\}\cdot k_i>x/2\right) \\
\leq& \Pr\left(k_i>x/2\right) \\
\leq& \frac{\zeta}{{16}\Gamma(\zeta^{-1})} \exp\left(-{(x/2)}^\zeta\right),
\end{aligned}
\end{equation*}
{ where the first inequality is because $\max\{S-r(\hat{A}_i),0 \}\leq1$ based on reward $r\in[0,1]$ and $0\leq S\leq r$, and the second inequality is by applying \eqref{eq:LCB_light_high_prob}.} Note that $\zeta/\Gamma(\zeta^{-1})\leq3/2$, $\exp\left(2^\zeta\right)\leq e^2$. { Proposition~\ref{prop:exit_prob_KArm} tells that the total number of rounds follows a geometric distribution with success rate $3/4$ in addition to the initial $i_0$ rounds, where $i_0$ is the largest $i$ that fails to satisfy Inequality~\eqref{eq:i0_KArm_lite_def} in Proposition~\ref{prop:exit_prob_KArm}. Then,} using Proposition \ref{lemma:round_exp_all}, { similar to \eqref{eq:tail_LCB},} the probability of total regret in the LCB test exceeds $x/2$ can be bounded by
\begin{equation}
\begin{aligned}
&\Pr\left(\sum_{i=1}^{\infty}\max\left\{S-r(\hat{A}_i),0 \right\}\cdot k_i>x/2 \right)\\
\leq& \Pr\left(\sum_{i=1}^{i_0+N(1/4)}\max\left\{S-r(\hat{A}_i),0 \right\}\cdot k_i>x/2 \right)\\ 
\leq& {\frac{2^{i_0}(1-1/4)}{1-2\cdot1/4}\cdot\exp\left(-\frac{(x/2)^\zeta}{\zeta/({16}\Gamma(\zeta^{-1}))+1} \right)} \\
\leq& { 2^{i_0+1}\cdot\exp\left(-\frac{x^\zeta/2}{3/16+1}\right)}\\
\leq& { \exp\left(i_0+1-\frac{x^\zeta}{3}\right)}
\label{eq:lcb_tail_KArm_sat}
\end{aligned}
\end{equation}

For the non-realizable case, the probability that cumulative standard regret incurred in oracle algorithm exceeds $x/2$ is upper bounded by
\begin{equation}
\begin{aligned}
&\Pr\left(\sum_{i=1}^{\infty}(r(A^*)-r(\hat{A}_i))\cdot k_i>x/2 \right) \\
\leq& \Pr\left(\left(\sum_{i=1}^{i_{\max}}k_i\right)^\zeta>\left(\frac{x}{2}\right)^\zeta \right)\\ 
\leq& \Pr\left(\exp\left(\sum_{i=1}^{i_{\max}}k_i^\zeta\right)>\exp\left(\left(\frac{x}{2}\right)^\zeta\right) \right) \\
\leq& \frac{\E\left[\exp\left(\sum_{i=1}^{i_{\max}}k_i^\zeta\right) \right]}{\exp((x/2)^\zeta)}\\
=& \exp\left(-(x/2)^\zeta\right)\prod_{i=1}^{i_{\max}}\E\left[\exp\left(k_i^\zeta \right)\right],
\label{eq:lcb_tail_KArm_non_1}
\end{aligned}
\end{equation}
{ where where the first inequality is based on reward $0\leq r(\hat{A}_t)\leq r(A^*)\leq1$, and the second inequality is based on Markov inequality, and the independence among each round $i$ is used in the final equality.} For each term $\E\left[\exp\left(k_i^\zeta \right)\right]$, it can be bounded by
\begin{equation}
\begin{aligned}
\E\left[\exp\left(k_i^\zeta \right)\right] &= \sum_{s=0}^{\infty}\Pr\left(k_i=s \right)\cdot\exp\left(s^\zeta \right) \\
&\leq 1+\sum_{s=1}^{\infty}\Pr\left(k_i=s \right)\cdot\exp\left(s^\zeta \right)\\
&\leq 1+\frac{\zeta}{{16}\Gamma(\zeta^{-1})}\sum_{s=1}^{\infty}\exp\left(-2s^\zeta+s^\zeta \right)\\
&\leq 1+\frac{\zeta}{{16}\Gamma(\zeta^{-1})}\cdot\zeta^{-1}\Gamma(\zeta^{-1})\leq2,
\label{eq:lcb_tail_KArm_non_2}
\end{aligned}
\end{equation}
{ where the second inequality is based on Inequality~\eqref{eq:prob_bound_2_tail_KArm}. Note that, the bound on $\E[\exp(k_i^\zeta)]$} is independent of the round $i$. { With $\max\{S-r(\hat{A}_i),0\}\leq1$ based on reward $r\in[0,1] $ and $r(A^*)\geq S$, we plug \eqref{eq:lcb_tail_KArm_non_2} into \eqref{eq:lcb_tail_KArm_non_1}, and} can bound the probability that cumulative standard regret incurred in oracle algorithm exceeds $x/2$ by
\begin{equation}
    \Pr\left(\sum_{i=1}^{\infty}(r(A^*)-r(\hat{A}_i))\cdot k_i>x/2 \right) \leq \exp\left(-(x/2)^\zeta\right)\cdot 2^{1+\alpha_2\log_2(T)}=2T^{\alpha_2}\exp\left(-(x/2)^\zeta\right).
\label{eq:lcb_tail_KArm_non}
\end{equation}

Combining oracle tail probability \eqref{eq:oracle_tail_KArm} with the tail probability~\eqref{eq:lcb_tail_KArm_sat} in the LCB test for the realizable case and the tail probability~\eqref{eq:lcb_tail_KArm_non} in the LCB test for the non-realizable case, we { bound the tail probability in the realizable case as}
$$
\Pr\left(\texttt{Regret}_S>x\right) \leq (\alpha_2\log_2(T)+1)p\left(\frac{x}{2(\alpha_2\log_2(T)+1)};\,K,T \right) + \exp\left(i_0{+1-\frac{x^\zeta}{3}} \right)
$$
{ and tail probability in the non-realizable case as}
$$
\Pr\left(\texttt{Regret}>x\right) \leq (\alpha_2\log_2(T)+1)p\left(\frac{x}{2(\alpha_2\log_2(T)+1)};\,K,T \right) + 2T^{\alpha_2}\exp\left(-(x/2)^\zeta\right).
$$

\subsection{Proof of Theorem~\ref{thm:regret_bound_KArm}}\label{sec:thm:regret_bound_KArm}
Denote $\beta'=\max\{\beta,1/2\}$ and $K_{\alpha,\beta,\zeta}=\frac{32C_1K^{\alpha_1+1}(1-\zeta)^{-\zeta/2}}{(1-\alpha_2)^\beta} {\sqrt{2+ 4\log\left(16\zeta^{-1}\Gamma(\zeta^{-1})\right)}}$, and $i_0$ as
\begin{equation*}
i_0=\max\{i: K_{\alpha,\beta,\zeta}\gamma_i^{1-\zeta}\log({4}/\gamma_i)^{\max\{\beta,1/2\}}>\Delta_S^*\}.
\end{equation*}
{ We first follow the proof of Theorem~\ref{thm:tail_risk} in Section~\ref{sec:thm:tail_risk} and Theorem~\ref{thm:regret_bound_1} in Section~\ref{sec:thm:regret_bound_1} to get} an upper bound on $i_0$. Denote $\delta_0\in(0,1)$ the smallest solution to the following equation
\begin{equation*}
\delta_0^{1-\zeta}\log({4}/\delta_0)^{\beta'}=\dfrac{\Delta_S^*}{K_{\alpha,\beta,\zeta}},
\end{equation*}
{ the same as Inequality~\eqref{eq:delta_Delta_equation-lite}. Following exactly the same proof steps to Inequality~\eqref{eq:1/delta^{1-zeta}}, we get}
\begin{equation*}
\dfrac{1}{\delta_0^{1-\zeta}}\leq \left(\dfrac{(8\beta')^{\beta'}K_{\alpha,\beta,\zeta}{/(1-\zeta)^{\beta'}}}{\Delta_S^*}\right)^{4/3}.
\end{equation*}
We write $D_{\alpha,\beta,\zeta}={4^{1-\zeta}}(8\beta')^{\beta'}K_{\alpha,\beta,\zeta}$, $L_{\alpha,\beta,\zeta}=(4/(3(1-\zeta)))^{\beta'}K_{\alpha,\beta,\zeta}$, then, { the same as \eqref{eq:delta_Delta_lite_2},} we have
\begin{equation*}
\delta_0^{1-\zeta} \geq \dfrac{\Delta_S^*}{L_{\alpha,\beta,\zeta}\log(D_{\alpha,\beta,\zeta}/\Delta_S^*)^{\beta'}}.
\end{equation*}
By definition of $i_0$ we also have
\begin{equation*}
\gamma_{i_0}^{1-\zeta} \geq \dfrac{\Delta_S^*}{L_{\alpha,\beta,\zeta}\log(D_{\alpha,\beta,\zeta}/\Delta_S^*)^{\beta'}},
\end{equation*}
thus
\begin{equation*}
i_0\leq \dfrac{\alpha_2}{(1-\alpha_2)(1-\zeta)} \log_2\left(\dfrac{L_{\alpha,\beta,\zeta}\log(D_{\alpha,\beta,\zeta}/\Delta_S^*)^{\beta'}}{\Delta_S^*}\right).
\end{equation*}

Using the result of Proposition~\ref{prop:regret_bound_round_KArm} we have that the satisficing regret by round $i_0$ is bounded by
\begin{equation}
\begin{aligned}
&\sum_{i=1}^{i_0}\dfrac{{4}C_1K^{\alpha_1+1}}{(1-\alpha_2)^\beta}\gamma_i^{-\alpha_2/(1-\alpha_2)}\log(4/{\gamma_i})^\beta \\
=& \sum_{i=1}^{i_0}\dfrac{{4}C_1K^{\alpha_1+1}}{(1-\alpha_2)^\beta}2^i\log({4}/\delta_0)^\beta
=\dfrac{4C_1K^{\alpha_1+1}}{(1-\alpha_2)^\beta}(2^{i_0}{-1})\log({4}/\delta_0)^\beta\\
\leq &\dfrac{{8}C_1K^{\alpha_1+1}}{(1-\alpha_2)^\beta{ (1-\zeta)^\beta}} \left(\dfrac{L_{\alpha,\beta,\zeta}\log(D_{\alpha,\beta,\zeta}/\Delta_S^*)^{\beta'}}{\Delta_S^*}\right)^{\frac{\alpha_2}{(1-\alpha_2)(1-\zeta)}} \log\left(\dfrac{2L_{\alpha,\beta,\zeta}\log(D_{\alpha,\beta,\zeta}/\Delta_S^*)^{\beta'}}{\Delta_S^*}\right)^\beta.
\label{eq:sat_regret_before_i0_KArm_lite}
\end{aligned}
\end{equation}

By the results of Proposition~\ref{prop:exit_prob_KArm}, for every $k\geq 1$, the probability that round $i_0+k$ starts within the time horizon is at most $4^{-k+1}$. Therefore the regret incurred after round $i_0$ is bounded by
\begin{equation}
\begin{aligned}
&\sum_{k=1}^\infty \dfrac{{4}C_1K^{\alpha_1+1}}{(1-\alpha_2)^\beta} 2^{i_0+k} \log(4/\gamma_{i_0+k})^\beta \cdot 4^{-k+1}\\
\leq & { \sum_{k=1}^\infty \dfrac{16C_1K^{\alpha_1+1}}{(1-\alpha_2)^\beta} 2^{i_0} \left(\log(4/\gamma_{i_0}) + \frac{1-\alpha_2}{\alpha_2}k\right)^\beta \cdot 2^{-k}}\\
\leq&{ \sum_{k=1}^\infty \dfrac{16\cdot 2^\beta C_1K^{\alpha_1+1}}{(1-\alpha_2)^\beta}2^{i_0}\log(4/\gamma_{i_0})^\beta \cdot \dfrac{(1-\alpha_2)^\beta}{\alpha_2^\beta}k^\beta\cdot 2^{-k} }\\
\leq &\dfrac{{ 16\cdot 2^\beta C_1K^{\alpha_1+1}}}{\alpha_2^\beta} 2^{i_0} \log({4}/\delta_0)^\beta \cdot \sum_{k=1}^\infty k^{\beta}\cdot 2^{-k}\\
\leq &{\dfrac{ 16\cdot 2^\beta C_1K^{\alpha_1+1}M}{\alpha_2^\beta} 2^{i_0} \log\left(\left(4\cdot\delta_0^{-1+\zeta}\right)^{1/(1-\zeta)}\right)^\beta}\\
\leq& {\dfrac{ 16\cdot 2^\beta C_1K^{\alpha_1+1}M}{\alpha_2^\beta(1-\zeta)^\beta}} \left(\dfrac{L_{\alpha,\beta,\zeta}\log(D_{\alpha,\beta,\zeta}/\Delta_S^*)^{\beta'}}{\Delta_S^*}\right)^{\frac{\alpha_2}{(1-\alpha_2)(1-\zeta)}} \log\left(\dfrac{4L_{\alpha,\beta,\zeta}\log(D_{\alpha,\beta,\zeta}/\Delta_S^*)^{\beta'}}{\Delta_S^*}\right)^\beta,
\label{eq:sat_regret_after_i0_KArm_lite}
\end{aligned}
\end{equation}
where $M=\sum_{k=1}^\infty k^\beta\cdot2^{-k}$ is a constant only dependent on $\alpha,\beta,\zeta$, { and the second and the third inequality follows the same pattern in Inequality~\eqref{eq:sat_regret_after_i0}. Then, combining the satisficing regret incurred by round $i_0$ \eqref{eq:sat_regret_before_i0_KArm_lite} and after $i_0$ \eqref{eq:sat_regret_after_i0_KArm_lite},} the total satisficing regret is bounded by
\begin{equation*}
\begin{aligned}
\mathbb{E}[\texttt{Regret}_S]\leq& {\dfrac{8C_1K^{\alpha_1+1}(2^{\beta+1}\alpha_2^{-\beta}M+(1-\alpha_2)^{-\beta})}{(1-\zeta)^\beta}} \left(\dfrac{L_{\alpha,\beta,\zeta}}{\Delta_S^*}\right)^{\frac{\alpha_2}{(1-\alpha_2)(1-\zeta)}}\\
&\cdot \log\left(\frac{D_{\alpha,\beta,\zeta}}{\Delta_S^*}\right)^{\frac{\alpha_2\beta'}{(1-\alpha_2)(1-\zeta)}} \log\left(\dfrac{2L_{\alpha,\beta,\zeta}}{\Delta_S^*}\log\left(\frac{D_{\alpha,\beta,\zeta}}{\Delta_S^*}\right)^{\beta'}\right)^\beta.
\end{aligned}
\end{equation*}

Furthermore, { since by definition of $t_i=\lceil2^{i/\alpha_2}\rceil$, the number of rounds that end within the time horizon is bounded by $2^{i/\alpha_2}\leq T$, \ie, $i\leq\lfloor\alpha_2\log_2(T) \rfloor$ because $i$ is an integer. Then, the maximum number of rounds is bounded by $i_{\max}=\lceil\alpha_2\log_2(T)\rceil$.} Therefore, we have that the satisficing regret is bounded by
\begin{align*}
\mathbb{E}[\texttt{Regret}_S]&\leq \sum_{i=1}^{i_{\max}}\dfrac{4C_1K^{\alpha_1+1}}{(1-\alpha_2)^\beta}\gamma_i^{-\alpha_2/(1-\alpha_2)}\log({4/\gamma_i})^\beta\\
&\leq \sum_{i=1}^{i_{\max}} \dfrac{4C_1K^{\alpha_1+1}}{(1-\alpha_2)^\beta}2^{i+1}\log({8}T)^\beta\\
&\leq \dfrac{{32}C_1K^{\alpha_1+1}}{(1-\alpha_2)^\beta}T^{\alpha_2}\log({8}T)^\beta
\end{align*}
{ where the first inequality is based on Proposition \ref{prop:regret_bound_round_KArm}, the second inequality is based on the definition $\gamma_i=2^{-i(1-\alpha_2)/\alpha_2}$ and $4/\gamma_i\leq8T$ as shown in \eqref{eq:gammai_1/T}.}

\subsection{Proof of Proposition~\ref{prop:regret_bound_round_KArm}}\label{sec:prop:regret_bound_round_KArm}
{ We follow the proof of Proposition~\ref{prop:regret_bound_round_1} in Section~\ref{sec:prop:regret_bound_round_1}.} Recall that $t_i=\lceil\gamma_i^{-1/(1-\alpha_2)} \rceil$, { similar to \eqref{eq:Step1_Regret_real},} the satisficing regret incurred when running $\texttt{ALG}(t_i)$ is bounded by
\begin{equation}
\E\left[\sum_{s=1}^{t_i}\max\{S-r(A_s),0\}\right]\leq\E\left[t_ir(A^*)-\sum_{s=1}^{t_i}r(A_s)\right]\leq \dfrac{2C_1K^{\alpha_1}}{(1-\alpha_2)^\beta}\gamma_i^{-\alpha_2/(1-\alpha_2)}\log({4}/\gamma_i)^\beta.
\label{eq:Step1_Regret_real_KArm_lite}
\end{equation}
Since $\hat{A}_i$ is the most frequently selected arm, it has been pulled for at least $t_i/K$ times during $\texttt{ALG}(t_i)$, and we further have
\begin{equation}
\E[\max\{S-r(\hat{A}_i),0\}]\leq\E[r(A^*)-r(\hat{A}_i)] \leq \dfrac{\E\left[t_ir(A^*)-\sum_{s=1}^{t_i}r(A_s)\right]}{t_i/K} \leq \dfrac{2C_1K^{\alpha_1+1}}{(1-\alpha_2)^\beta}\gamma_i\log({4}/\gamma_i)^\beta.
\label{eq:single_pull_regret_KArm_lite}
\end{equation}
If $r(\hat{A}_i)<S$, then for every $k$, round $i$ is not terminated after pulling arm $\hat{A}_i$ for $k$ times (in addition to the number of pulls $\tau_i$ during $\texttt{ALG}(t_i)$) occurs with probability at most
\begin{equation}
\begin{aligned}
\nonumber&\Pr\left(\dfrac{\hat{r}^{\text{tot}}_i}{\tau_i+k} -\sqrt{\frac{2k^\zeta+\log\left({16}\zeta^{-1}\Gamma(\zeta^{-1})\right)}{\tau_i+k}} > S\right)\\
\nonumber\leq &\Pr\left(\dfrac{\hat{r}^{\text{tot}}_i}{\tau_i+k} - \sqrt{\frac{2k^\zeta+\log\left({16}\zeta^{-1}\Gamma(\zeta^{-1})\right)}{\tau_i+k}} > r(\hat{A}_i)\right)\\
\leq&\exp\left(-2k^\zeta + \log\left(\frac{\zeta}{{16}\Gamma(\zeta^{-1})} \right) \right) {\leq} \frac{\zeta}{{16}\Gamma(\zeta^{-1})}\exp\left(-k^\zeta \right).
\label{eq:radius_bound}
\end{aligned}
\end{equation}
Therefore expected number of arm pulls of $\hat{A}_i$ after first pulling it $\tau_i$ times is bounded by
\begin{equation*}
\begin{aligned}
    &\E\left[\text{Number of times }\hat{A}_i\text{ is pulled in round }i\right]\\
    =&\sum_{s=1}^{\infty}\Pr(k\geq s) =\frac{\zeta}{{16}\Gamma(\zeta^{-1})}\sum_{s=1}^\infty \exp\left(-s^\zeta \right) \leq \frac{\zeta}{{16}\Gamma(\zeta^{-1})} \int_0^\infty \exp\left(-x^\zeta \right)\mathrm{d}x\\
    \leq& \frac{\zeta}{{16}\Gamma(\zeta^{-1})}\cdot\zeta^{-1}\Gamma(\zeta^{-1}) =\frac{1}{16}\leq 1.
\end{aligned}
\end{equation*}
Summing up the satisficing regret { incurred in oracle running \eqref{eq:Step1_Regret_real_KArm_lite} and LCB tests \eqref{eq:single_pull_regret_KArm_lite},} we have that the satisficing regret of round $i$ is bounded by
\begin{equation*}
\begin{aligned}
&{ \dfrac{2C_1K^{\alpha_1}}{(1-\alpha_2)^\beta}\gamma_i^{-\alpha_2/(1-\alpha_2)}\log({4}/\gamma_i)^\beta + 1\cdot\dfrac{2C_1K^{\alpha_1+1}}{(1-\alpha_2)^\beta}\gamma_i\log({4}/\gamma_i)^\beta} \\
\leq& \dfrac{{4}C_1K^{\alpha_1+1}}{(1-\alpha_2)^\beta}\gamma_i^{-\alpha_2/(1-\alpha_2)}\log({4}/\gamma_i)^\beta.
\end{aligned}
\end{equation*}

\subsection{Proof of Proposition~\ref{prop:exit_prob_KArm}}\label{sec:prop:exit_prob_KArm}
{ We follow the proof of Proposition~\ref{prop:exit_prob_tail} in Section~\ref{sec:prop:exit_prob_tail}.} Recall from the proof of Proposition \ref{prop:regret_bound_round_KArm}, we have
\begin{align*}
\E[\max\{S-r(\hat{A}_i),0\}]\leq\E[r(A^*)-r(\hat{A}_i)] \leq \dfrac{\E\left[t_ir(A^*)-\sum_{s=1}^{t_i}r(A_s)\right]}{t_i/K} \leq \dfrac{2C_1K^{\alpha_1+1}}{(1-\alpha_2)^\beta}\gamma_i\log({4}/\gamma_i)^\beta.
\end{align*}

According to the condition of this proposition, we have $i$ satisfies the condition $$\dfrac{32C_1K^{\alpha_1+1}(1-\zeta)^{-\zeta/2}}{(1-\alpha_2)^\beta} {\sqrt{2+ 4\log\left(16\zeta^{-1}\Gamma(\zeta^{-1})\right)}} \gamma_i^{1-\zeta} \log({4}/\gamma_i)^{\max\{\beta,1/2\}} \leq \Delta_S^*.$$ 
Note that for any $0<\zeta<1$, $(1-\zeta)^{-\zeta/2}>1$ and $\gamma_i^{1-\zeta}>\gamma_i$. Therefore $i$ satisfies
$$\dfrac{32C_1K^{\alpha_1+1}}{(1-\alpha_2)^\beta}\gamma_i\log({4}/\gamma_i)^{\max\{\beta,1/2\}}\leq \Delta_S^*,$$
by the Markov's inequality we have
\begin{equation}
\Pr(r(A^*)-r(\hat{A}_i)\geq \Delta_S^*/2)\leq \Pr\left(r(A^*)-r(\hat{A}_i)\geq \dfrac{16C_1K^{\alpha_1+1}}{(1-\alpha_2)^\beta}\gamma_i\log({4}/\gamma_i)^\beta\right)\leq\dfrac{1}{8}.
\label{eq:prob_bound_1_tail_KArm}
\end{equation}

Now, if $r(\hat{A}_i)\geq r(A^*)-\Delta_S^*/2$, which happens with probability at least 7/8 according to the above inequality, then $r(\hat{A}_i)\geq S+\Delta_S^*/2$ holds. On the other hand, if $i$ satisfies the condition stated in the proposition, then
$$
(1-\zeta)^{-\zeta/2}\cdot {\sqrt{2+ 4\log\left(16\zeta^{-1}\Gamma(\zeta^{-1})\right)}} \gamma_i^{1-\zeta}\log({4}/\gamma_i)^{1/2}\leq \Delta_S^*/16.
$$
Note that for $f(y)=y^\zeta/(\tau_i+y)$, it reaches its maximum when 
$$
\frac{\mathrm{d}}{\mathrm{d}y}f(y) = \frac{\zeta y^{\zeta-1}\tau_i+y^\zeta(\zeta-1)}{(\tau_i+y)^2}=0.
$$
Therefore the maximum of $f(y)$ is reached at $\zeta\tau_i/(1-\zeta)$. For simplicity, we set $D = \log\left({16}\zeta^{-1}\Gamma(\zeta^{-1})\right)$. { Similar to Inequality~\eqref{eq:CI<Delta/8},} the confidence radius for any $k$ is bounded by
\begin{equation*}
\begin{aligned}
    &\sqrt{\frac{2k^\zeta+D}{\tau_i+k}}\leq \sqrt{\frac{{(D+2)}k^\zeta}{\tau_i+k}} \leq \sqrt{\frac{{(D+2)}\left(\frac{\zeta\tau_i}{1-\zeta}\right)^\zeta}{\tau_i+\left(\frac{\zeta\tau_i}{1-\zeta}\right)}}\\
    \leq&\sqrt{\frac{{(D+2)}(1-\zeta)^{-\zeta}\cdot\zeta^\zeta \tau_i^\zeta}{\tau_i}}\leq {\sqrt{D+2}K^{(1-\zeta)/2}} (1-\zeta)^{-\zeta/2}\gamma_i^{1-\zeta} \\
    \leq& {2\sqrt{D+2} K (1-\zeta)^{-\zeta/2} \gamma_i^{1-\zeta} \log({4}/\gamma_i)^{1/2}} \leq \frac{\Delta_S^*}{{8}},
\end{aligned}
\end{equation*}
{ where the only difference to Inequality~\eqref{eq:CI<Delta/8} is the constant term, \ie, $D+2$ instead of $D+1$, and $\tau_i\geq t_i/K\geq\gamma_i^{-1/(1-\alpha_2)}/K\geq\gamma_i^{-2}/K$ is used.} Suppose round $i$ is terminated for some $k$, then
\begin{equation*}
\dfrac{\hat{r}_i^{\text{tot}}}{\tau_i+k} - \sqrt{\frac{2k^\zeta+D}{\tau_i+k}} <S=r(A^*)-\Delta_S^*\leq r(\hat{A}_i)-\dfrac{\Delta_S^*}{2}\leq r(\hat{A}_i) - 2\sqrt{\frac{2k^\zeta+D}{\tau_i+k}},
\end{equation*}
or equivalently
\begin{equation*}
\dfrac{\hat{r}_i^{\text{tot}}}{\tau_i+k}\leq r(\hat{A}_i) - \sqrt{\frac{2k^\zeta+D}{\tau_i+k}}.
\end{equation*}
By the Hoeffding's inequality we have
\begin{equation}
\Pr\left(\dfrac{\hat{r}_i^{\text{tot}}}{\tau_i+k}\leq r(\hat{A}_i) - \sqrt{\frac{2k^\zeta+D}{\tau_i+k}}\right) \leq \frac{\zeta}{{16}\Gamma(\zeta^{-1})} \exp\left(-2k^\zeta \right) \leq \frac{\zeta}{{16}\Gamma(\zeta^{-1})} \exp\left(-k^\zeta \right).
\label{eq:prob_bound_2_tail_KArm}
\end{equation}
Therefore conditioned on $r(\hat{A}_i)\geq r(A^*)-\Delta_S^*/2$ the probability that round $i$ is terminated by the end of the horizon is bounded by
\begin{equation}
\begin{aligned}
    \frac{\zeta}{{16}\Gamma(\zeta^{-1})}\sum_{k={0}}^\infty \exp\left(-k^\zeta \right) 
    &\leq { \frac{\zeta}{16\Gamma(\zeta^{-1})} +} \frac{\zeta}{{16}\Gamma(\zeta^{-1})} \int_0^\infty \exp\left(-x^\zeta \right)\mathrm{d}x\\
    &\leq { \frac{1}{16} +} \frac{\zeta}{{16}\Gamma(\zeta^{-1})}\cdot\zeta^{-1}\Gamma(\zeta^{-1}) =\frac{1}{8}
\label{eq:prob_bound_2_tail_KArm_cumulative}
\end{aligned}
\end{equation}
Combining \eqref{eq:prob_bound_1_tail_KArm} and \eqref{eq:prob_bound_2_tail_KArm_cumulative} we conclude that the probability that round $i$ terminates by the end of the time horizon is { upper} bounded by $1/4$.

\subsection{Proof of Theorem \ref{thm:regret_nonsat_KArm}}\label{sec:thm:regret_nonsat_KArm}
We first prove that the regret incurred in round $i$ is at most
$$
\dfrac{2C_1K^{\alpha_1+1}}{(1-\alpha_2)^\beta}\gamma_i^{-\alpha_2/(1-\alpha_2)}\log({4}/\gamma_i)^\beta.
$$

{ Similar to \eqref{eq:Step1_Regret_real} with modification on constant term $C_1$ to $C_1K^{\alpha_1}$,} the standard regret incurred when running $\texttt{ALG}(t_i)$ is bounded by
\begin{equation}
\E\left[t_ir(A^*)-\sum_{s=1}^{t_i}r(A_s)\right]\leq \dfrac{2C_1K^{\alpha_1}}{(1-\alpha_2)^\beta}\gamma_i^{-\alpha_2/(1-\alpha_2)}\log({4}/\gamma_i)^\beta.
\label{eq:Step1_Regret_real_KArm_lite_std}
\end{equation}
Since $\hat{A}_i$ is the most frequently selected arm, it has been pulled for at least $t_i/K$ times during $\texttt{ALG}(t_i)$, and { similar to \eqref{eq:single_pull_regret}} we further have
\begin{equation}
\E[r(A^*)-r(\hat{A}_i)] \leq \dfrac{\E\left[t_ir(A^*)-\sum_{s=1}^{t_i}r(A_s)\right]}{t_i/K} \leq \dfrac{2C_1K^{\alpha_1+1}}{(1-\alpha_2)^\beta}\gamma_i\log({4}/\gamma_i)^\beta.
\label{eq:single_pull_regret_KArm_lite_std}
\end{equation}
Since $r(\hat{A}_i)\leq r(A^*)<S$, then for every $k$, round $i$ is not terminated after pulling arm $\hat{A}_i$ for $k$ times (in addition to the number of pulls $\tau_i$ during $\texttt{ALG}(t_i)$) occurs with probability at most
\begin{equation}
\begin{aligned}
\nonumber&\Pr\left(\dfrac{\hat{r}^{\text{tot}}_i}{\tau_i+k} -\sqrt{\frac{2k^\zeta+\log\left({16}\zeta^{-1}\Gamma(\zeta^{-1})\right)}{\tau_i+k}} > S\right)\\
\nonumber\leq &\Pr\left(\dfrac{\hat{r}^{\text{tot}}_i}{\tau_i+k} - \sqrt{\frac{2k^\zeta+\log\left({16}\zeta^{-1}\Gamma(\zeta^{-1})\right)}{\tau_i+k}} > r(\hat{A}_i)\right)\\
\leq&\exp\left(-2k^\zeta + \log\left(\frac{\zeta}{{16}\Gamma(\zeta^{-1})} \right) \right) = \frac{\zeta}{{16}\Gamma(\zeta^{-1})}\exp\left(-k^\zeta \right).
\label{eq:radius_bound}
\end{aligned}
\end{equation}
Therefore expected number of arm pulls of $\hat{A}_i$ after first pulling it $\tau_i$ times is bounded by
\begin{equation}
\begin{aligned}
    &\sum_{s=1}^\infty\frac{\zeta}{{16}\Gamma(\zeta^{-1})} \exp\left(-k^\zeta \right) \leq \frac{\zeta}{{16}\Gamma(\zeta^{-1})} \int_0^\infty \exp\left(-x^\zeta \right)\mathrm{d}x\\
    \leq& \frac{\zeta}{{16}\Gamma(\zeta^{-1})}\cdot\zeta^{-1}\Gamma(\zeta^{-1}) =\frac{1}{16}\leq 1.
\label{eq:real3_Karm_lite_std}
\end{aligned}
\end{equation}
Summing up the standard regret { in oracle running \eqref{eq:Step1_Regret_real_KArm_lite_std} and LCB tests as a combination of \eqref{eq:single_pull_regret_KArm_lite_std} and \eqref{eq:real3_Karm_lite_std},} we have that the standard regret of round $i$ is bounded by
\begin{equation*}
\dfrac{2C_1K^{\alpha_1+1}}{(1-\alpha_2)^\beta}\gamma_i^{-\alpha_2/(1-\alpha_2)}\log({4}/\gamma_i)^\beta.
\end{equation*}

Since the length of the time horizon is $T$ time steps, we have $t_i\leq T$, following  $i_{\max}=\lceil\alpha_2\log_2(T)\rceil$. Therefore, { following similar chain of inequality in \eqref{eq:main_bound_sublinear},} we have that the standard regret is bounded by
\begin{align*}
\mathbb{E}[\texttt{Regret}]\leq&\sum_{i=1}^{i_{\max}}\dfrac{4C_1K^{\alpha_1+1}}{(1-\alpha_2)^\beta}\gamma_i^{-\alpha_2/(1-\alpha_2)}\log({8}T)^\beta \\
\leq & \sum_{i=1}^{i_{\max}} \dfrac{4C_1K^{\alpha_1+1}}{(1-\alpha_2)^\beta}2^{i+1}\log({8T})^\beta\\
\leq & \dfrac{16C_1K^{\alpha_1+1}}{(1-\alpha_2)^\beta}T^{\alpha_2}\log({8T})^\beta
\end{align*}
Therefore the regret is bounded by $C_1K^{\alpha_1+1}T^{\alpha_2}\cdot\text{polylog}(T)$.

\section{Choices of $C_1,\alpha,\beta$ for Several Bandit Optimization Problem Classes}\label{sec:oracle_parameter}
In this section, we provide a detailed discussion on the choice of $C_1,\alpha,\beta$ for several bandit optimization problem classes. In three cases

\subsection{Finite-armed Bandits}\label{sec:K-arm-oracle-parameter}
Consider any instance of finite-armed bandits with mean reward bounded by $[0,1]$. By Theorem 7.2 of \cite{LS20}, for any given time horizon $t\geq 2$, the standard regret of the UCB algorithm is upper bounded by 
$$
8\sqrt{Kt\log(t)}+3\sum_{k=1}^K (r(k^*)-r(k))\leq 8\sqrt{Kt\log(t)}+3K,
$$ 
where $K\geq 2$ is the number of arms and $k^*$ is the optimal arm. 

When $t>K$, the regret is upper bounded by $8\sqrt{Kt\log(t)}+3K\leq 11\sqrt{Kt\log(t)}$, where the last inequality is due to $t>K\geq 2$, thus $t\geq 3$ and $\log(t)\geq \log(3)>1$; when $t\leq K$, the regret is upper bounded by $t\leq \sqrt{Kt}\leq 11\sqrt{Kt\log(2)}\leq 11\sqrt{Kt\log(t)}$. 

Thus for any given time horizon $t\geq 2$, the regret of the UCB algorithm is bounded by $11\sqrt{Kt\log(t)}$. 
Therefore, by setting $C_1=11\sqrt{K}$ and $\alpha=\beta=1/2$ and using UCB algorithm as the oracle, Condition~\ref{assump:learning_alg} is satisfied for finite-armed bandits.

\subsection{Concave Bandits}\label{sec:concave-oracle-parameter}
Consider any concave bandits with $\mathcal{A}\subseteq\mathbb{R}^d$ being a compact and convex set contained in a Euclidean ball with radius $1$, and reward function being concave and $1$-Lipschitz. By Theorem 2 of \cite{AgarwalFH11}, for any given time horizon $t\geq 2$, the standard regret of Algorithm 2 in \cite{AgarwalFH11} is upper bounded by 
\begin{equation*}
768d^3\sqrt{t}\log(t)^2\left(\dfrac{2d^2\log(d)}{c_2^2}+1\right)\left(\dfrac{4d^7c_1}{c_2^3}+\dfrac{d(d+1)}{c_2}\right)\left(\dfrac{4c_1d^4}{c_2^2}+11\right),
\end{equation*}
where $c_1,c_2$ are algorithm hyperparameters that satisfies $c_1\geq 64$, $c_2\leq 1/32$. Therefore Condition~\ref{assump:learning_alg}~is satisfied for concave bandits by using Algorithm 2 in \cite{AgarwalFH11} and setting $\alpha=1/2$, $\beta=2$ and
\begin{equation*}
C_1=768d^3\left(\dfrac{2d^2\log(d)}{c_2^2}+1\right)\left(\dfrac{4d^7c_1}{c_2^3}+\dfrac{d(d+1)}{c_2}\right)\left(\dfrac{4c_1d^4}{c_2^2}+11\right),
\end{equation*}

Furthermore, if the instance of concave bandit is one-dimensional, \ie, $\mathcal{A}=[0,1]$, by Theorem 1 of \cite{AgarwalFH11}, for any given time horizon $t\geq 2$, the standard regret of Algorithm 1 in \cite{AgarwalFH11} is bounded by
\begin{equation*}
108\sqrt{t\log(t)}\cdot\log_{4/3}(t)=108/\log(4/3)\sqrt{t}\cdot\log(t)^{3/2}.
\end{equation*}
Therefore Condition~\ref{assump:learning_alg}~is also satisfied for one-dimensional concave bandits by using Algorithm 1 in \cite{AgarwalFH11}~and setting $C_1=108/\log(4/3)$, $\alpha=1/2$, $\beta=3/2$.

\subsection{Lipschitz Bandits}\label{sec:lipschitz-oracle-parameter}
Consider any Lipschitz bandits with arm set being $[0,1]^d$, reward function being $L$-Lipschitz and mean reward bounded on $[0,1]$ (here we define Lipschitz continuity in the sense of $\infty$-norm). By Section 2.2 of \cite{BubeckSY2011}, if the regret of the UCB algorithm for finite-armed bandits with $K$ arms is bounded by $c_{\text{ALG}}\sqrt{Kt\log(t)}$ for any $t\geq 2$, then for any given time horizon $t\geq 2$, the standard regret of the uniformly discretized UCB algorithm introduced by \cite{BubeckSY2011} (hereafter referred to as the ``Uniform UCB") is upper bounded by 
$$
(1+c_{\text{ALG}})L^{d/(d+2)}t^{(d+1)/(d+2)}\cdot\sqrt{\log(t)}.
$$

Using the results in Section~\ref{sec:K-arm-oracle-parameter}, we have $c_{\text{ALG}}=11$, thus the regret of the Uniform UCB algorithm is bounded by 
$$
12L^{d/(d+2)}t^{(d+1)/(d+2)}\cdot\sqrt{\log(t)}.
$$
Therefore, by setting $C_1=12L^{d/(d+2)}$, $\alpha=(d+1)/(d+2)$, $\beta=1/2$ and using Uniform UCB as the oracle, Condition~\ref{assump:learning_alg} is satisfied for Lipschitz bandits.

\section{Ablation Study for the Three Steps of~\satex}

In this section, we use an ablation study to demonstrate that all three steps of~\satex~play a vital role in obtaining a constant satisficing regret. Specifically, we compare the performance of~\satex~with the following adapted versions of~\satex:

\begin{itemize}
\item {\bf \satex~without Step 1:} instead of identifying candidate satisficing arms using a learning oracle, the candidate satisficing arm is drawn uniformly at random from the arm set before we proceed to forced sampling and LCB test in each round;

\item {\bf \satex~without Step 2:} After identifying a candidate satisficing arm from Step 1 in each round, the forced sampling in Step 2 is skipped and an LCB test on the candidate satisficing arm is immediately started;

\item {\bf \satex~without Step 3:} Each round is terminated after running Step 1 and Step 2 without entering the LCB test in Step 3.
\end{itemize}

\vspace{2mm}
\noindent\textbf{Setup:} We conduct the ablation test using an instance of the Lipschitz bandit. We use the reward function defined in Section~\ref{sec:experiment_lipschitz}, and set the satisficing threshold $S=0.7$. The learning oracle we use in Step 1 is Uniform UCB introduced in \cite{BubeckSY2011}.

\begin{figure}[!ht]
\centering
\includegraphics[height=4.5cm]{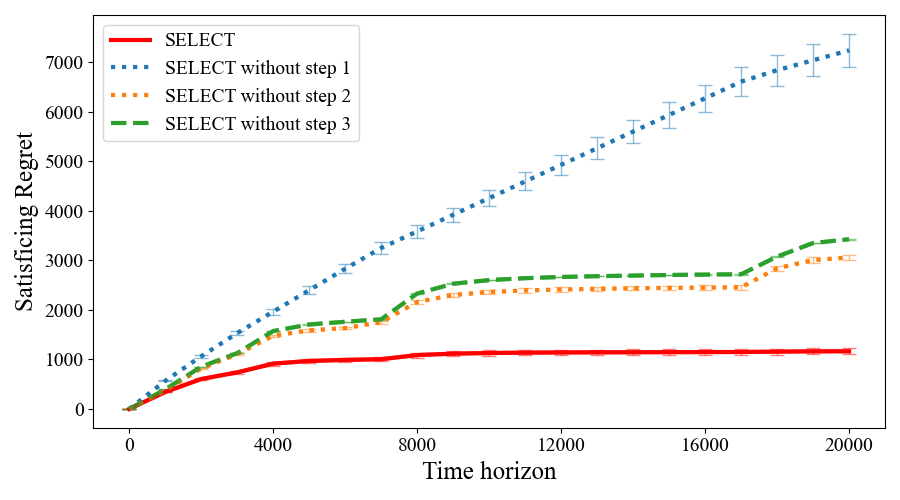}
\caption{Results of the ablation test}
\label{fig:abltion-test}
\end{figure}

\vspace{2mm}
\noindent\textbf{Results:} The results are provided in Figure~\ref{fig:abltion-test}. One can see that after removing any of the three steps, our algorithm is unable to maintain constant satisficing regret in the realizable case. Specifically, Step 1 serves the purpose of efficiently identifying a candidate satisficing arm. If Step 1 is replaced by randomly drawing candidate satisficing arms from the arm set, then the satisficing regret is almost linear in $T$, indicating that the algorithm struggles to identify satisficing arms without the help of the learning oracle. Step 2 and Step 3 combined mainly serve the purpose of exploiting the candidate satisficing arm and at the same time, testing whether it is truly satisficing. Without either Step 2 or Step 3, the satisficing regret starts a round of rapid increase whenever the time horizon $T$ exceeds certain thresholds, thus failing to maintain constant satisficing regret. This indicates that without either Step 2 or Step 3, the algorithm almost always terminates a round shortly after completing Step 1 even if the candidate satisficing arm obtained from Step 1 is indeed satisficing. We refer to Remark~\ref{remark:lcb}~for a detailed discussion on the reason for this. Therefore we conclude that all three steps of~\satex~play a vital role in obtaining a constant satisficing regret bound in the realizable case.

\section{Robustness on the choice of $\gamma_i$}
\begin{figure}[!ht]
    \centering
    \subfigure[Realizable case]{\includegraphics[height=4.4cm]{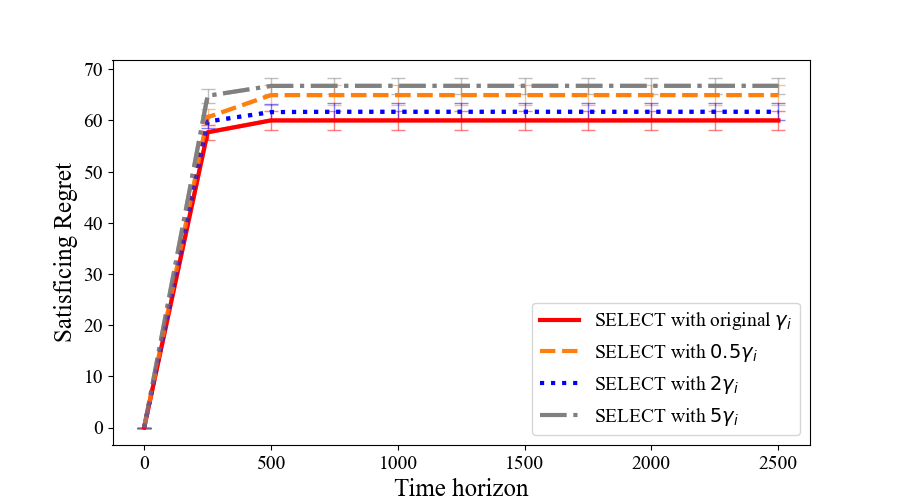}
    \label{fig:robustness_gamma_realizable}}
    \subfigure[Non-realizable case]{\includegraphics[height=4.4cm]{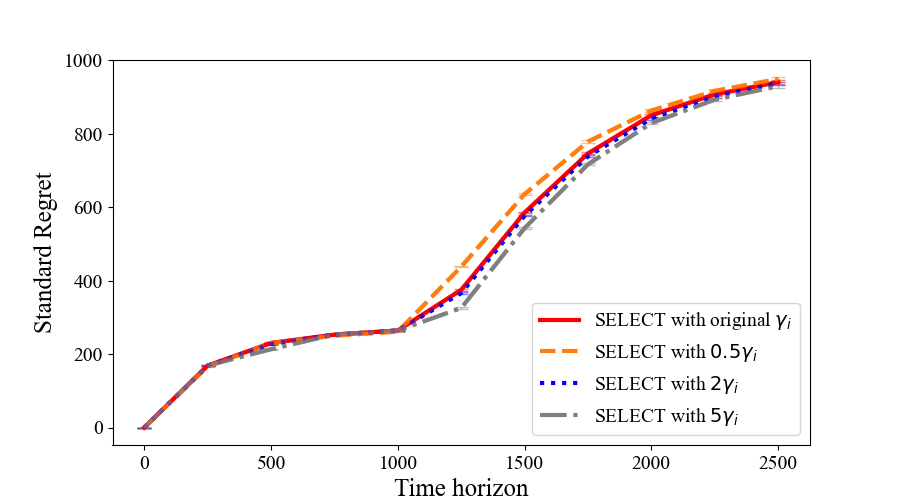}
    \label{fig:robustness_gamma_nonrealizable}}
    
    \caption{Performance of~\satex~with different choice of $\gamma_i$}
\end{figure}
In Algorithm~\ref{alg:sat_exploration}, we define $\gamma_i=2^{-i(1-\alpha)/\alpha}$ for all $i\in\mathbb{Z}_+$, but all our results still hold if all $\gamma_i$ are multiplied by a constant. In this section, we conduct numerical experiments on testing the robustness of~\satex~in the choice $\gamma_i$.

\vspace{2mm}
\noindent\textbf{Setup:} We conduct the experiment on the same instance of Lipschitz bandit in Section~\ref{sec:experiment_lipschitz}. We multiply the $\{\gamma_i\}$ in Algorithm~\ref{alg:sat_exploration}~with a constant $\lambda$, and vary $\lambda=0.5,1,2,5$.

\vspace{2mm}
\noindent\textbf{Results:} The result of the realizable case is provided in Figure~\ref{fig:robustness_gamma_realizable}, and the result of the non-realizable case is provided in Figure~\ref{fig:robustness_gamma_nonrealizable}. One can see from the results that different choices of $\gamma_i$ have limited impact on the empirical performance of~\satex, both in the realizable and non-realizable case. Therefore the empirical performance of \satex~is robust to the choice of $\gamma_i$.

{
\section{Ablation Test for Oracle Choice in Finite-Armed Bandits}
In this section, we conduct an ablation study to demonstrate that in finite-armed bandits, using oracle algorithm with sub-linear standard regret bound instead of a uniform sampling, the approach used in \cite{Michel23} and \cite{GarivierM19}, plays a vital role in obtaining a small constant satisficing regret in the realizable case and a sub-linear standard regret in the non-realizable case. 

Similar to Section~\ref{sec:numerical_KArm} and Section~\ref{sec:best_of_both_world}, we consider an instance of finite-armed bandits with 4 arms, and the expected rewards of all arms are $\{0.2,0.4,0.6,0.8 \}$. We vary the length of the time horizon from 500 to 7000 with a stepsize of 500. We conduct experiments for both the realizable case and the non-realizable case. For the realizable case, we set the satisficing level $S\in\{0.65,0.7,0.75\}$; for the non-realizable case, we set the satisficing level $S = 1.5$. In both cases, we test the performance of \texttt{SELECT-BoBW} with different oracle algorithms, including Thompson sampling, UCB algorithm, and uniform exploration (\ie, draw an arm uniformly at random at each time step). The experiment is repeated for 5000 times and we report the average satisficing regret in the realizable case and the average standard regret in the non-realizable case.

The results of the realizable case are presented in Figure~\ref{fig:oracle_ablation_0.65}, \ref{fig:oracle_ablation_0.7}, and \ref{fig:oracle_ablation_0.75}, and the results of the non-realizable case are presented in Figure~\ref{fig:oracle_ablation_non}. From Figure~\ref{fig:oracle_ablation_0.65} and \ref{fig:oracle_ablation_0.7}, one can observe that although \texttt{SELECT-BoBW} using uniform sampling as oracle algorithm is able to attain a constant expected satisficing regret, it exhibits a much slower convergence rate and incurs a much larger expected satisficing regret, compared to \texttt{SELECT-BoBW} using Thompson sampling or UCB as oracle algorithms. Furthermore, one can observe from Figure~\ref{fig:oracle_ablation_0.75} that if $S$ is large and $\Delta_S^*$ is small, \texttt{SELECT-BoBW} using uniform sampling as oracle algorithm might even fail to reach a constant satisficing regret within a time horizon of $7000$. Furthermore, in all these three realizable cases, \texttt{SELECT-BoBW} using uniform sampling as oracle algorithm has higher variance according to the error bar. This is because unlike an oracle algorithm with sub-linear standard regret bound, uniform exploration does not converge to the optimal arm, and it is more likely that a non-satisficing arm is selected as a candidate for Step 2 and 3 in each round. When comparing \texttt{SELECT-BoBW} using Thompson sampling with \texttt{SELECT-BoBW} using UCB algorithm, one can observe that the performance of \texttt{SELECT-BoBW} with Thompson sampling outperforms \texttt{SELECT-BoBW} with UCB. This observation indicates that while any learning oracle with sub-linear expected standard regret enables \texttt{SELECT-BoBW} to attain a constant expected satisficing regret, \texttt{SELECT-BoBW} can also attain better empirical performance if a learning oracle with better empirical performance is used in Step 1.

From Figure~\ref{fig:oracle_ablation_non}, one can see that in the non-realizable case, \texttt{SELECT-BoBW} using uniform exploration as oracle algorithm incurs linear expected standard regret in $T$, and the expected standard regret is much larger than that of \texttt{SELECT-BoBW} using sub-linear oracle algorithms. Therefore we conclude that using an oracle algorithm with sub-linear standard regret bound instead of a uniform exploration is necessary.

\begin{figure}[!ht]
    \centering
    \subfigure[Satisficing Regret in the Realizable Case $S=0.65$]{\includegraphics[height=4.5cm]{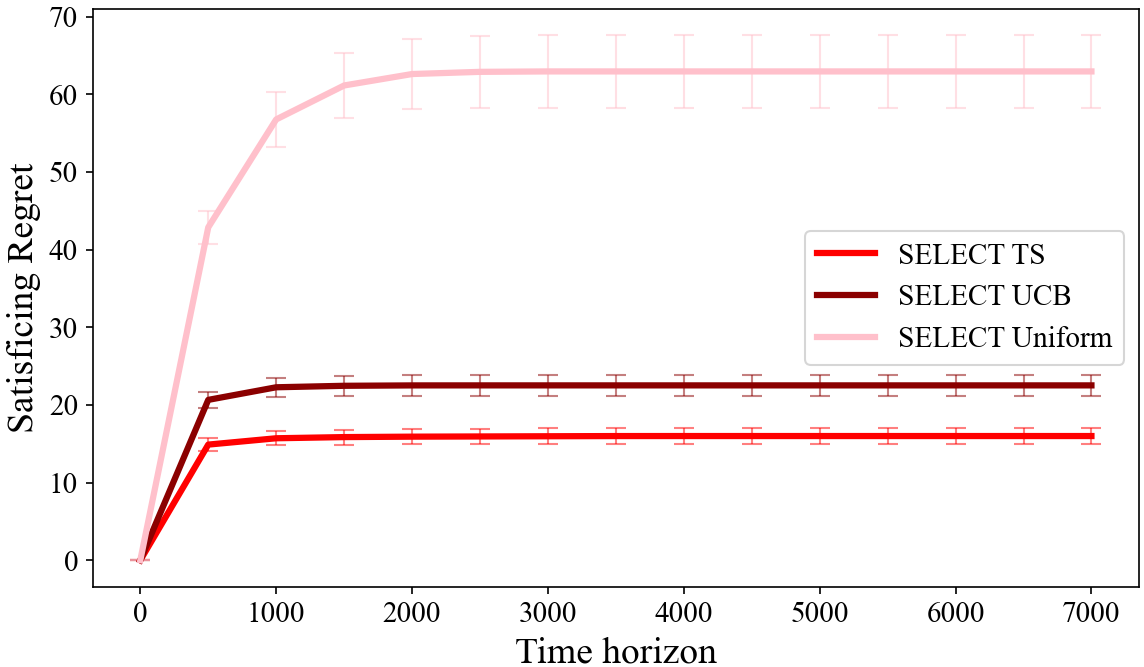}
    \label{fig:oracle_ablation_0.65}}
    \subfigure[Satisficing Regret in the Realizable Case $S=0.7$]{\includegraphics[height=4.5cm]{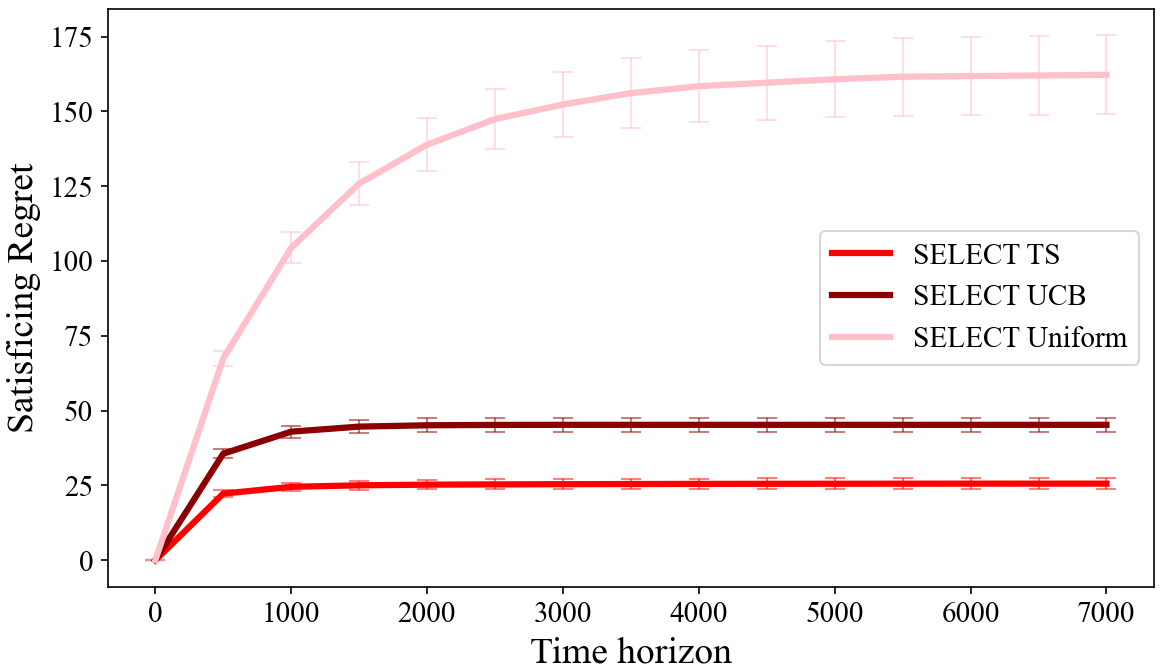}
    \label{fig:oracle_ablation_0.7}}
    \subfigure[Satisficing Regret in the Realizable Case $S=0.75$]{\includegraphics[height=4.5cm]{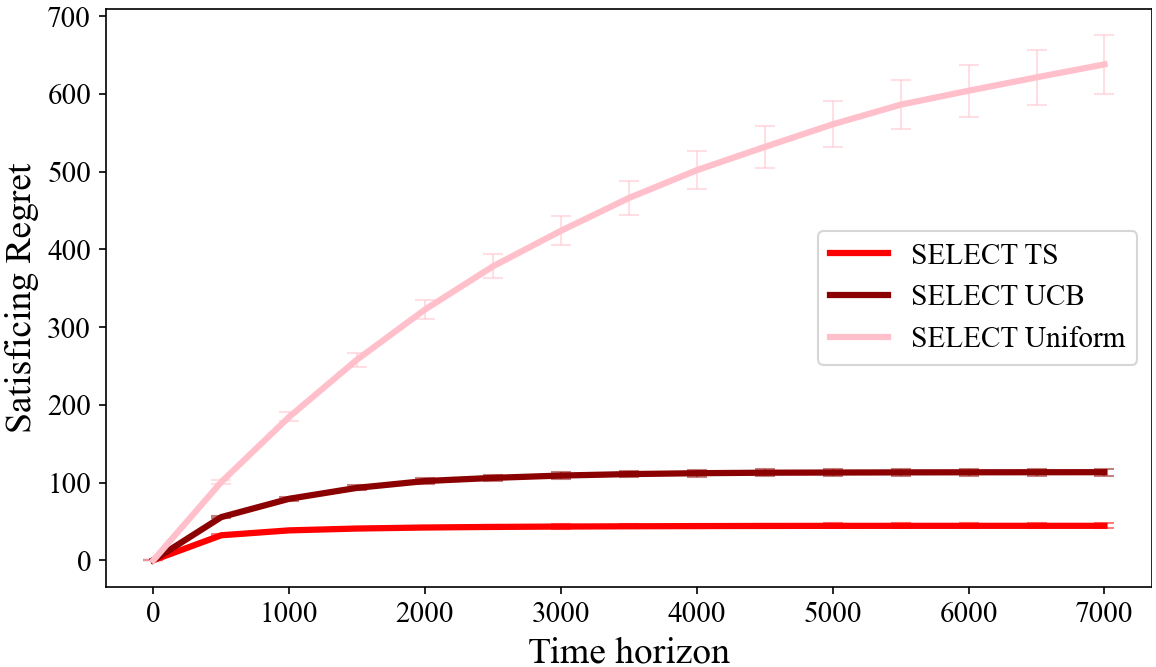}
    \label{fig:oracle_ablation_0.75}}
    \subfigure[Standard Regret in the Non-Realizable Case]{\includegraphics[height=4.5cm]{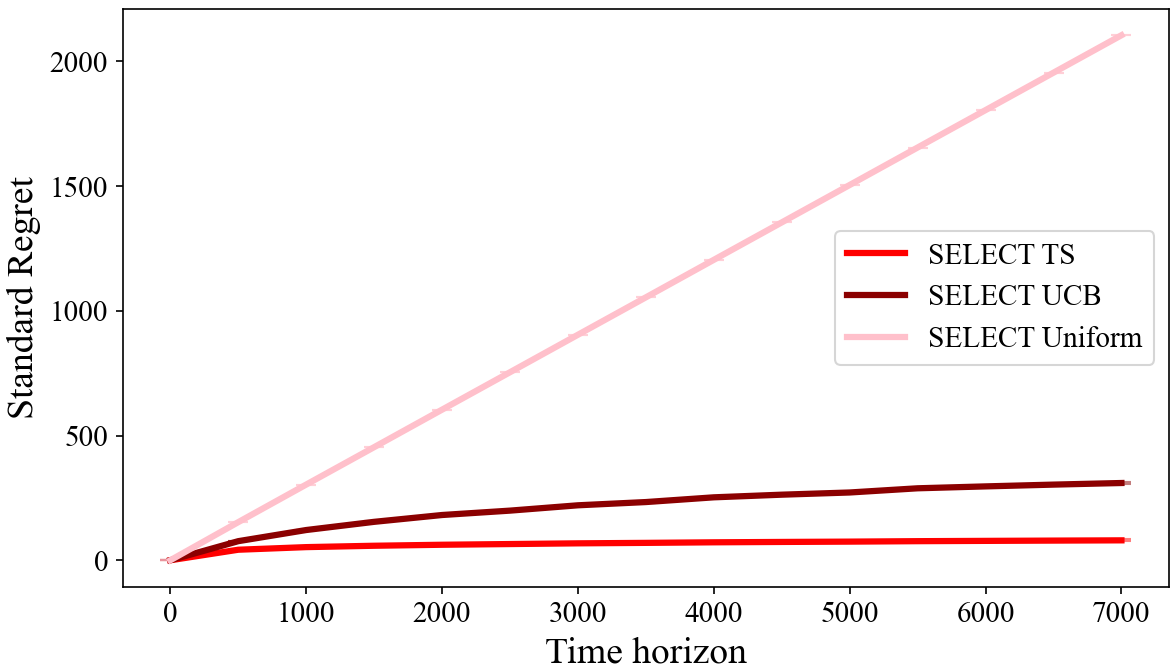}
    \label{fig:oracle_ablation_non}}
    
    \caption{Performance of \texttt{SELECT-BoBW} using Thompson sampling, UCB, and uniform sampling as oracle algorithm}
\end{figure}

\section{Ablation Test for Design of $\{\gamma_i\}$}

In order to highlight the importance of exponential decay of $\{\gamma_i\}$, in this section, We conduct an ablation test comparing different designs of $\{\gamma_i\}$ using an instance of Lipschitz bandit. Specifically, we use the instance of Lipschitz bandit used in Section~\ref{sec:experiment_lipschitz}, where the arm set is $\mathcal{A}=[0,1]^2$, and the reward function is $r(x,y) =e^{-100((x-0.5)^2 + (y-0.6)^2)}$. We vary the length of the time horizon from $500$ to $10000$ with a stepsize of $500$. We use the Uniform UCB as the learning oracle of \satex, and we compare \satex~equipped with the following three designs of $\{\gamma_i\}$: 1) exponential decay, $\gamma_i=2^{-i(1-\alpha)/\alpha}$, 2) linear decay $\gamma_i=(i/2)^{-1}$, and 3) constant $\gamma_i=1/4$.

The result of the realizable case is provided in Figure~\ref{fig:robustness_design_realizable}, and the result of the non-realizable case is provided in Figure~\ref{fig:robustness_design_nonrealizable}. One can see from \cref{fig:robustness_design_realizable} that in the realizable case, although both exponentially decaying $\{\gamma_i\}$ and linearly decaying $\{\gamma_i\}$ attain a constant expected satisficing regret, the regret incurred by the exponentially decaying $\{\gamma_i\}$ is smaller than that incurred by linearly decaying $\{\gamma_i\}$. On the other hand, the satisficing regret with constant $\{\gamma_i\}$ increases linearly, failing to reach a constant value. In the non-realizable case, the exponential design of $\gamma_i$ also incurs the least standard regret, followed by the linear design, while the constant design performs the worst. This shows that using a decreasing sequence of $\{\gamma_i\}$ is necessary for attaining a constant expected satisficing regret, and exponentially decaying $\{\gamma_i\}$ used in our algorithm is better than linearly decaying $\{\gamma_i\}$.

\begin{figure}[!ht]
    \centering
    \subfigure[Realizable case]{\includegraphics[height=4.3cm]{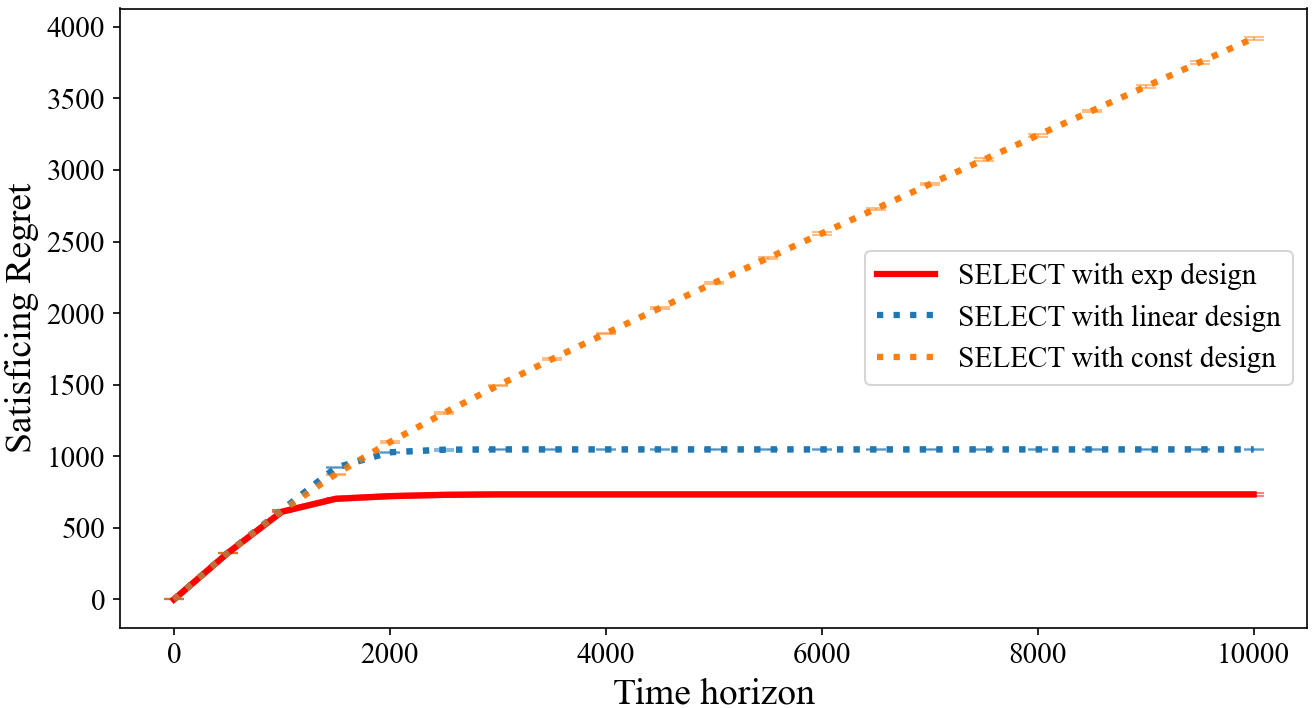}
    \label{fig:robustness_design_realizable}}
    \subfigure[Non-realizable case]{\includegraphics[height=4.3cm]{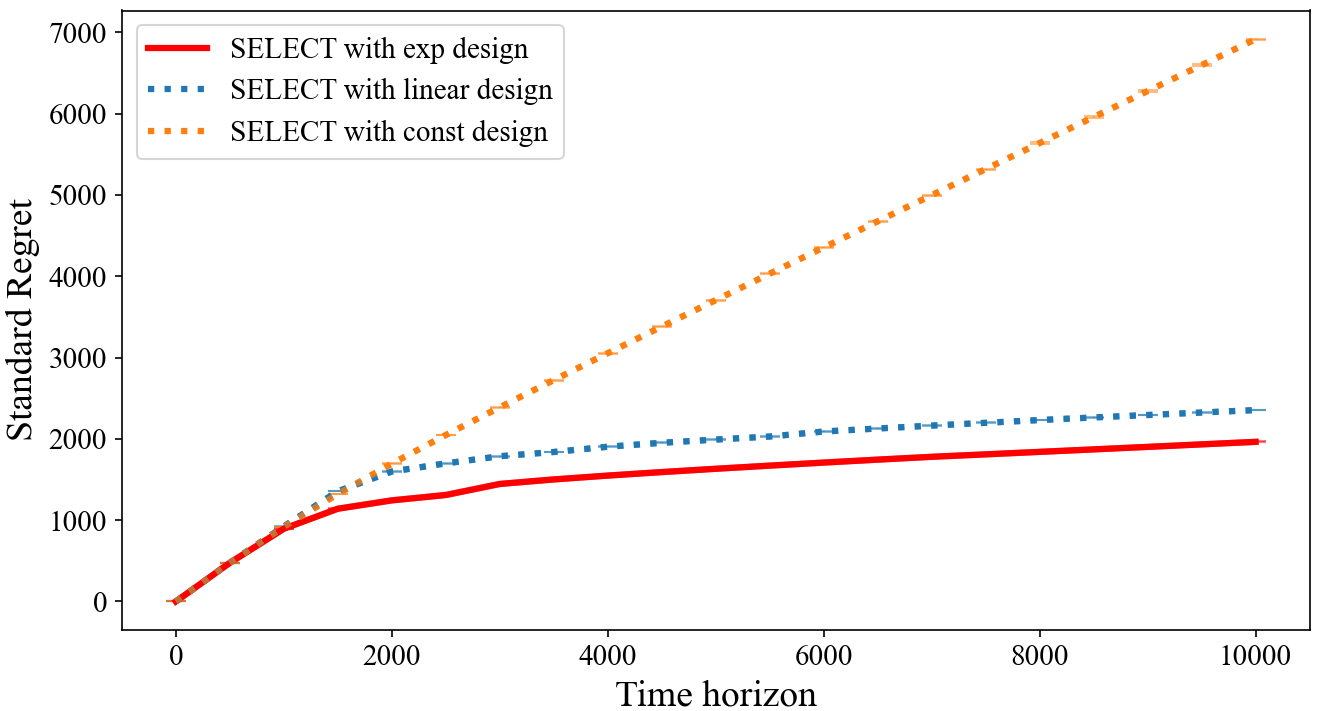}
    \label{fig:robustness_design_nonrealizable}}
    
    \caption{Performance of \texttt{SELECT} with different design of $\gamma_i$}
\end{figure}

\section{Best-of-Both-Worlds Regret Bound for Finite-Arm Bandits}\label{sec:best_of_both_world}

For satisficing regret minimization in finite-arm bandits, existing works such as \cite{GarivierM19} establishes a constant expected satisficing regret bound of $\tilde{O}(K/\Delta_S)$, where $\Delta_S$ is the satisficing gap given by $\Delta_S=\min\{S-r(A):A\in\mathcal{A},\,r(A)<S\}$ (see Theorem 9 therein). On the other hand, our algorithm \satex~is able to attain a $\tilde{O}(K/\Delta_S^*)$ expected regret bound, where $\Delta_S^*$ is the exceeding gap defined as $\Delta_S^*=r(A^*)-S$. When comparing our expected regret bound in \cref{cor:k_arm_bandit} with the one in \cite{GarivierM19}, one can see that our algorithm \satex~potentially has a better performance when $\Delta_S^*$ is large and $\Delta_S$ is small, while the algorithm in \cite{GarivierM19} potentially has a better performance when $\Delta_S^*$ is small and $\Delta_S$ is large. 

In order to establish an algorithm that attains a ``best of both worlds" expected satisficing regret bound and works well in both cases, in this section, we present \texttt{SELECT-BoBW}, an adapted version of \satex~for finite-armed bandits that incurs an expected satisficing regret bound $\tilde{O}(\min\{K/\Delta_S,K/\Delta_S^*\})$. Specifically, \texttt{SELECT-BoBW} achieves a ``best of both worlds" expected satisficing regret bound, dominating both the regret bound of \satex~and the one in \cite{GarivierM19}. The complete algorithm is provided in \cref{alg:sat_KArm_bbw}. Note that in \texttt{SELECT-BoBW}, we can use either the UCB algorithm (see, \eg, \cite{BubeckC12}) or Thompson sampling (see, \eg, \cite{AgrawalG17}) as the learning oracle $\texttt{ALG}$.

\begin{algorithm}[!ht]
\caption{Satisficing Regret Minimization for Finite-Armed Bandits (\texttt{SELECT-BoBW})}
\label{alg:sat_KArm_bbw}
\begin{algorithmic}
\SingleSpacedXI
\State \textbf{Input:} time horizon $T$, satisficing level $S$
\State Set $\gamma_i\leftarrow 2^{-i}$ for all $i=1,2,3\dots$
\For{round $i=1,2,\dots$}
\State Set $t_i\leftarrow\lceil 4^i\rceil$
\State Run $\texttt{ALG}(t_i)$, let $\{\bar{A}_s\}_{s\in[t_i]}$ be the trajectory of arm pulls
\State Sample $q$ uniformly at random from $\{1,2,\dots,t_i\}$ and set $\hat{A}_i\leftarrow \bar{A}_q$
\State Set $T_i\leftarrow 4^i$, $k\leftarrow 0$, $\hat{r}_i^{\text{tot}}\leftarrow0$
\For{$\tau = 1,2,\cdots,T_i$}
\State Pull arm $\hat{A}_i$ and observe $Y_t$, where $t$ is the current time step
\State Update $\hat{r}_i^{\text{tot}}\leftarrow \hat{r}_i^{\text{tot}}+Y_t$
\State Set
    \begin{equation*}
    UCB(\hat{A}_i)\leftarrow\dfrac{\hat{r}_i^{\text{tot}}}{\tau}+\sqrt{\dfrac{4\log(\tau+1)}{\tau}}
    \end{equation*}
\If{$UCB(\hat{A}_i) < S$}
\State Continue to next round $i+1$
\EndIf
\EndFor
\State Set $LCB(\hat{A}_i)\leftarrow\hat{r}_i^{\text{tot}}/T_i-\sqrt{4\log(T_i)/T_i}$
\While{$LCB(\hat{A}_i)\geq S$}
\State Set $k\leftarrow k+1$
\State Pull arm $\hat{A}_i$ and observe $Y_t$, where $t$ is the current time step
\State Update $\hat{r}_i^{\text{tot}}\leftarrow \hat{r}_i^{\text{tot}}+Y_t$
\State Set
    \begin{equation*}
    LCB(\hat{A}_i)\leftarrow\dfrac{\hat{r}_i^{\text{tot}}}{T_i+k}-\sqrt{\dfrac{4\log(T_i+k)}{T_i+k}}
    \end{equation*}
\EndWhile
\EndFor
\end{algorithmic}
\end{algorithm}

The key difference between \texttt{SELECT-BoBW} and \satex~is that when running forced sampling in Step 2 of \texttt{SELECT-BoBW}, we add a UCB test on the candidate arm's mean reward. Once the candidate arm's UCB falls below the threshold $S$, we determine that it is highly unlikely for the arm to be satisficing, thus the forced sampling in Step 2 is terminated early and the algorithm immediately start a new round. With the newly added UCB test, any non-satisficing candidate arm $\hat{A}$ with a satisficing gap $\Delta_S(\hat{A})=S-r(\hat{A})$ can be removed with at most $\tilde{O}(1/\Delta_S(\hat{A})^2)$, incurring a satisficing regret of at most $\Delta_S(\hat{A})\cdot \tilde{O}(1/\Delta_S(\hat{A})^2)=\tilde{O}(1/\Delta_S(\hat{A}))\leq \tilde{O}(1/\Delta_S)$.

We use the following theorem to establish an expected satisficing regret bound for \texttt{SELECT-BoBW}.

\begin{theorem}\label{thm:regret_bound_bbw}
If $r(A^*)\geq S$, then the expected satisficing regret of~\texttt{SELECT-BoBW}~is bounded by
\begin{equation}\label{eq:real_bbw}
\E[\texttt{Regret}_S]\leq\min\left\{ \dfrac{K}{\Delta_S^*}\cdot\textup{polylog}\left(\dfrac{K}{\Delta_S^*}\right) ,\, \frac{K}{\Delta_S}\cdot\textup{polylog}\left(\dfrac{K}{\Delta_S^*\Delta_S}\right) ,\, \sqrt{KT}\cdot\textup{polylog}(T)\right\}.
\end{equation}
\end{theorem}

Proof of Theorem~\ref{thm:regret_bound_bbw} is provided in Section~\ref{sec:thm:regret_bound_bbw}. 

We also show that in the non-realizable case, $\texttt{SELECT-BoBW}$ enjoys the same standard regret bound as the oracle shown in Theorem~\ref{thm:regret_nonsat_bbw}.

\begin{theorem}\label{thm:regret_nonsat_bbw}
If $r(A^*)<S$, the standard regret of \texttt{SELECT-BoBW}~is bounded by
\begin{equation*}
\E[\regret]\leq\sqrt{KT}\cdot\textup{polylog}(T).
\end{equation*}
\end{theorem}

Proof of \cref{thm:regret_nonsat_bbw} is provided in Section~\ref{sec:thm:regret_nonsat_bbw} of the appendix.

\begin{remark}\label{remark:regret_bound_bbw_wo_step2}
We would like to remark that forced sampling is not necessary in attaining a constant expected satisficing regret in finite-armed bandits. In fact, a simplified version of \texttt{SELECT-BoBW} is to directly use the arm that is pulled most frequently along the trajectory during $\texttt{ALG}(t_i)$ as the candidate arm, similar to Algorithm~\ref{alg:sat_lite_KArm}, and leverage the observed reward of this arm during oracle running. Using similar arguments in the proof of \cref{thm:regret_bound_bbw}, we can prove that the expected satisficing regret of \texttt{SELECT-Arm} without forced sampling in the realizable case is bounded by
\begin{equation}\label{eq:real_bbw_wo_step2}
\E[\texttt{Regret}_S]\leq\min\left\{ \min\left\{\dfrac{K}{\Delta_S^*}, \frac{K}{\Delta_S} \right\} \cdot \textup{polylog}\left(\dfrac{K}{\Delta_S^*}\right)  ,\, K\sqrt{T}\cdot\textup{polylog}(T)\right\}.
\end{equation}
In other words, after removing forced sampling, the algorithm is still able to attain a constant expected satisficing regret bound of $\tilde{O}(\min\{K/\Delta_S,K/\Delta_S^*\})$ similar to \texttt{SELECT-BoBW}. However, when $\Delta_S^*$ is small, the algorithm without forced sampling potentially incurs a larger expected satisficing regret of $\tilde{O}(K\sqrt{T})$ rather than the corresponding $\tilde{O}(\sqrt{KT})$ expected satisficing regret bound for \texttt{SELECT-BoBW}. This is because the most frequently pulled arm is pulled for at least $t_i/K$ times during $\texttt{ALG}(t_i)$, and its (satisficing) regret of a single pull is bounded by $\tilde{O}(t_i^{\alpha}/(t_i/K)) = \tilde{O}(K\gamma_i)$,  which only suffers from a difference in constant term $K$ compared to the original one with forced sampling. At the same time, its width of the confidence interval at the beginning of LCB test is of order $\tilde{O}((t_i/K)^{-1/2})=\tilde{O}(\gamma_i\sqrt{K})$ by leveraging its observed reward during $\texttt{ALG}(t_i)$, which still only suffers from a difference in constant term $\sqrt{K}$ compared to the original one. That is, both the expected satisficing regret incurred in round $i$ given in Proposition~\ref{prop:regret_bound_round_bbw}, and the round $i_0$ that we can start to bound the exit probability given by Proposition~\ref{prop:exit_prob_bbw}, only change at most by a constant term $K$. On the other hand, by skipping the forced sampling, we are able to get rid of the $1/\Delta_S\cdot\log\left(1/\Delta_S\right)$ in Proposition~\ref{prop:regret_bound_round_bbw} which is incurred in forced sampling. Finally, we are still able to bound the expected satisficing regret by $\tilde{O}\left(\min\{K/\Delta_S^*,K/\Delta_S\}\right)$ and a slightly worse sub-linear satisficing regret $\tilde{O}(K\sqrt{T})$.
\end{remark}

\subsection{Numerical Experiment}\label{sec:experiment_vary_S}
To test the performance of \texttt{SELECT-BoBW} under different pairs of satisficing gap and exceeding gap, we conduct a numerical experiment under a wide range of $(\Delta_S,\Delta_S^*)$ pairs on finite-armed bandits. We consider an instance of 4 arms, and the expected rewards of all arms are $\{0.2,0.4,0.6,0.8 \}$, the same as the setting in Section~\ref{sec:numerical_KArm}. We fix the time horizon to be $10000$, and vary the satisficing level $S$ from 0.62 to 0.78 with a stepsize of 0.01, resulting in the pair $(\Delta_S,\Delta_S^*)\in\{(0.02,0.18),(0.04,0.16),\cdots,(0.18,0.02)\}$. We compare \texttt{SELECT-BoBW} against SAT-UCB and SAT-UCB+ in~\cite{Michel23}, and the ``Phased Exploration algorithm" in \cite{GarivierM19} with $\mu^*$ changed to $S$. For \texttt{SELECT-BoBW}, we use its simplified version noted in Remark~\ref{remark:regret_bound_bbw_wo_step2}, and use UCB algorithm and Thompson sampling as the learning oracle. The experiment is repeated for 1000 times and we report the average satisficing regret.

\begin{figure}[!ht]
    \centering
    \includegraphics[width=0.5\linewidth]{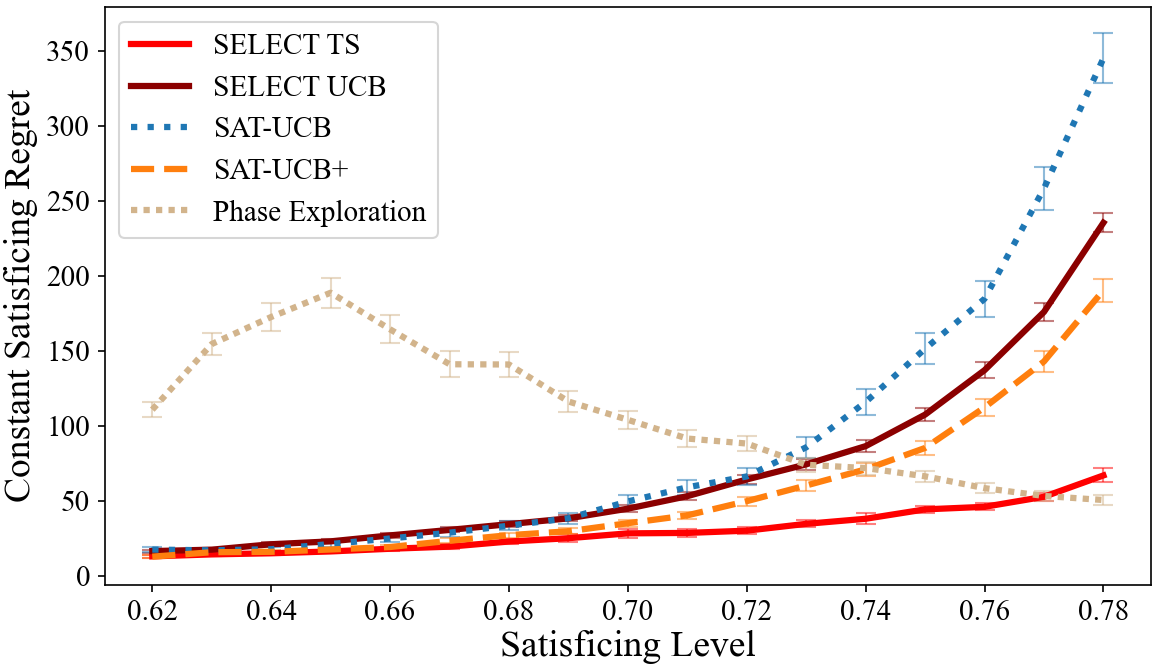}
    \caption{Constant satisficing regret changing with satisficing level $S$ at time horizon $T=10000$}
    \label{fig:const_regret_bbw}
\end{figure}

The results are presented in Figure~\ref{fig:const_regret_bbw}. Note that for the Phased Exploration algorithm, it fails to converge to a constant satisficing regret when $S<0.67$, even if we increase the time horizon to $T=20000$. Therefore the reported satisficing regret of the Phased Exploration algorithm when $S<0.67$ is smaller than its constant expected satisficing regret. One can observe from \cref{fig:const_regret_bbw} that \texttt{SELECT-BoBW} using Thompson sampling as oracle algorithm has the lowest constant expected satisficing regret under most of the $(\Delta_S,\Delta_S^*)$ pairs, except for $(\Delta_S,\Delta_S^*)=(0.18,0.02)$ (\ie, $S=0.78$). When $(\Delta_S,\Delta_S^*)=(0.18,0.02)$, the Phased Exploration algorithm has a lower satisficing regret than \texttt{SELECT-BoBW} due to an additional $\text{polylog}(1/\Delta_S^*)$ in the satisficing regret bound of \texttt{SELECT-BoBW}. The constant expected satisficing regret of \texttt{SELECT-BoBW} using UCB as oracle algorithm is comparable to that of the heuristic algorithm SAT-UCB+, and is lower than that of SAT-UCB.

\subsection{Proof of Theorem~\ref{thm:regret_bound_bbw}}\label{sec:thm:regret_bound_bbw}

For any sub-optimal arm $a\in\mathcal{A}$, we define $\Delta_a=r(A^*)-r(a)$, and $N_a(t)$ as the number of times arm $a$ is pulled if an algorithm (either the UCB algorithm or Thompson sampling) is run over a time horizon of $t$ time periods. By Theorem 2.1 of \cite{BubeckC12} and Theorem 1.1 of \cite{AgrawalG17}, there exist constants $c_1,c_2>0$ such that both the UCB algorithm and Thompson sampling satisfy the following two properties:

\begin{enumerate}
\item {\bf Worst-case expected standard regret bound:} the expected standard regret is upper bounded by $c_1\sqrt{KT\log(T)}$;
\item {\bf Instance-dependent bound on expected arm pulls:} The expected number of arm pulls of any sub-optimal arm $a$ is upper bounded by $\mathbb{E}[N_a(t)]\leq c_2\log(t)/\Delta_a^2$.
\end{enumerate}

Proof of the theorem relies on two critical results. First, we show an upper bound on the satisficing regret of round $i$.

\begin{proposition}\label{prop:regret_bound_round_bbw}
If $r(A^*)\geq S$, then the expected satisficing regret~incurred in round $i$ is bounded by
\begin{equation*}
\min\left\{4c_1\sqrt{K}\gamma_i^{-1}\log\left(\dfrac{1}{\gamma_i}\right)^{1/2} ,\; \frac{66}{\Delta_S}\log\left(\frac{5}{\Delta_S} \right) + \frac{K}{\Delta_S}\cdot 2c_2\log\left(\dfrac{1}{\gamma_i}\right)  \right\}.
\end{equation*}
\end{proposition}

Proof of \cref{prop:regret_bound_round_bbw} is provided in Section~\ref{sec:prop:regret_bound_round_bbw} of the appendix.

Next, we show that once $\texttt{SELECT-BoBW}$ runs for enough rounds, it is unlikely to start a new round.

\begin{proposition}\label{prop:exit_prob_bbw}
If $r(A^*)\geq S$, then for every $i$ that satisfies
\begin{equation*}
16\sqrt{2}c_1\sqrt{K}\gamma_i\log(1/\gamma_i)^{1/2}\leq \Delta_S^*,
\end{equation*}
the probability that round $i$ ends within the time horizon $T$ (conditioned on the event that round $i$ is started within the time horizon) is upper bounded by $1/4$.
\end{proposition}

Proof of \cref{prop:exit_prob_bbw} is provided in Section~\ref{sec:prop:exit_prob_bbw} of the appendix.

We define $i_0$ as
\begin{equation*}
i_0=\max\{i\in\mathbb{Z}_+: 16\sqrt{2}c_1\sqrt{K}\gamma_i\log(1/\gamma_i)^{1/2}>\Delta_S^*\},
\end{equation*}
and denote $\delta_0\in(0,1)$ the smallest solution to the following equation
\begin{equation*}
\delta_0\log(1/\delta_0)^{1/2}=\dfrac{\Delta_S^*}{16\sqrt{2}c_1\sqrt{K}}.
\end{equation*}
Then we have
\begin{equation*}
\dfrac{1}{\delta_0}=\dfrac{16\sqrt{2}c_1\sqrt{K}\log(1/\delta_0)^{1/2}}{\Delta_S^*}\leq \dfrac{16\sqrt{2}c_1\sqrt{K}\cdot 1/\delta_0^{1/2}}{\Delta_S^*}=\dfrac{16\sqrt{2}c_1\sqrt{K}}{\Delta_S^*}\cdot \dfrac{1}{\delta_0^{1/2}}.
\end{equation*}
Here the inequality is because $\log(x)<x$ for all $x>0$. Therefore we have
\begin{equation}
\dfrac{1}{\delta_0}\leq \left(\dfrac{16\sqrt{2}c_1\sqrt{K}}{\Delta_S^*}\right)^{2}.
\label{eq:1/delta0_upperbound_bbw}
\end{equation}
Therefore we have
\begin{equation*}
\delta_0=\dfrac{\Delta_S^*}{16\sqrt{2}c_1\sqrt{K}\cdot \log(1/\delta_0)^{1/2}}\geq \dfrac{\Delta_S^*}{32c_1\sqrt{K}\cdot \log(16\sqrt{2}c_1\sqrt{K}/\Delta_S^*)^{1/2}}.
\end{equation*}
Here the inequality follows from \eqref{eq:1/delta0_upperbound_bbw}. By definition of $i_0$ we have that $i_0$ is the largest integer that satisfies $\gamma_{i_0}\geq \delta_0$. Therefore we get
\begin{equation*}
\gamma_{i_0}\geq \dfrac{\Delta_S^*}{32c_1\sqrt{K}\cdot \log(16\sqrt{2}c_1\sqrt{K}/\Delta_S^*)^{1/2}}.
\end{equation*}
Recall that $\gamma_i=2^{-i}$, the above inequality implies that
\begin{equation}
i_0\leq \log_2\left(\dfrac{32c_1\sqrt{K}\cdot \log(16\sqrt{2}c_1\sqrt{K}/\Delta_S^*)^{1/2}}{\Delta_S^*}\right).
\label{eq:i0_upperbound}
\end{equation}

The satisficing regret bound depending only on $1/\Delta_S^*$ is exactly the same as the one in Theorem~\ref{thm:regret_bound_1}. In what follows, we focus on proving the expected satisficing regret bound that depends on $\Delta_S$.

Using the result of Proposition~\ref{prop:regret_bound_round_bbw}, we have that the satisficing regret by round $i_0$ is bounded by
\begin{equation}
\begin{aligned}
&\sum_{i=1}^{i_0} \left( \frac{66}{\Delta_S}\log\left(\frac{5}{\Delta_S} \right) + \frac{K}{\Delta_S}\cdot 2c_2\log\left(\dfrac{1}{\gamma_i}\right)\right) \\
\leq& i_0\cdot\left( \frac{66}{\Delta_S}\log\left(\frac{5}{\Delta_S} \right) + \frac{K}{\Delta_S}\cdot2c_2\log\left(\dfrac{1}{\delta_0}\right) \right)    \\
\leq& \frac{1}{\Delta_S} \log_2\left(\dfrac{64c_1\sqrt{K}\cdot \log(128\sqrt{2}c_1\sqrt{K}/\Delta_S^*)^{1/2}}{\Delta_S^*}\right)\cdot\left(66\log\left(\frac{5}{\Delta_S}\right) + 4c_2K\log\left(\frac{16\sqrt{2}c_1\sqrt{K}}{\Delta_S^*} \right) \right)\\
\leq&\dfrac{K}{\Delta_S}\cdot \text{polylog}\left(\dfrac{K}{\Delta_S\Delta_S^*}\right).
\label{eq:BBW_regret-before-i0}
\end{aligned}
\end{equation}
where the first inequality is by plugging into the definition of $\gamma_i$ and $\gamma_i\geq\gamma_{i_0}\geq\delta_0$ for any $i\leq i_0$, and the second inequality is due to \eqref{eq:1/delta0_upperbound_bbw} and \eqref{eq:i0_upperbound}.

By the results of Proposition~\ref{prop:exit_prob_bbw}, for every $k\geq 1$, the probability that round $i_0+k$ starts within the time horizon is at most $4^{-k+1}$. Therefore the regret incurred after round $i_0$ is bounded by
\begin{equation}
\begin{aligned}
&\sum_{k=1}^\infty \frac{1}{\Delta_S} \left( 66\log\left(\frac{5}{\Delta_S} \right) + 2Kc_2\log(1/\gamma_{i_0+k}) \right) \cdot 4^{-k+1} \\
\leq& \sum_{k=1}^\infty \frac{66}{\Delta_S}\log\left(\frac{5}{\Delta_S} \right)4^{-k+1} + \sum_{k=1}^\infty\left(\log(4/\gamma_{i_0}) + k\right) \frac{2Kc_2}{\Delta_S}\cdot 4^{-k+1}\\
\leq& \frac{88}{\Delta_S}\log\left(\frac{5}{\Delta_S} \right) + \sum_{k=1}^\infty 2\log(4/\gamma_{i_0}) k\cdot 4^{-k+1} \cdot\frac{2Kc_2}{\Delta_S}\\
\leq& \frac{88}{\Delta_S}\log\left(\frac{5}{\Delta_S} \right) + \log(4/\delta_0) \frac{4Kc_2}{\Delta_S} \sum_{k=1}^\infty k\cdot 4^{-k} \\
\leq& \frac{88}{\Delta_S}\log\left(\frac{5}{\Delta_S} \right) + \frac{16Kc_2}{9} \log\left(\dfrac{L_{\alpha,\beta}c_1\sqrt{K}}{\Delta_S^*} \right)\frac{1}{\Delta_S}\\
\leq&\dfrac{K}{\Delta_S}\cdot \text{polylog}\left(\dfrac{K}{\Delta_S\Delta_S^*}\right).
\label{eq:BBW_regret-after-i0}
\end{aligned}
\end{equation}
The first inequality follows directly from the definition of $\gamma_i=2^{-i}$ then $\log(1/\gamma_{i_0+k})=\log(1/2^{-(i_0+k)})=\log_2(2^{k})/\log_2(e)+\log(1/\gamma_{i_0})\leq\log(4/\gamma_{i_0})+k$. The second inequality is due to $\log(4/\gamma_{i_0})>1>1/2$ for $\gamma_{i_0}<1$, and for any $a,b\geq 1$, we have $2ab\geq ab+1\geq a+b$, and $\sum_{k\geq1}4^{-k+1}\leq4/3$. The third inequality is similar to the case in satisficing regret by round $i_0$. The fourth inequality is plugging into Inequality~\eqref{eq:1/delta0_upperbound_bbw} and $\sum_{k\geq1} k\cdot4^{-k}=4/9$. 

We note that $\sum1/\Delta_S^a\leq K/(\min\Delta_S^a)=K/\Delta_S$. Taking the sum of \eqref{eq:BBW_regret-before-i0} and \eqref{eq:BBW_regret-after-i0}, we have that the total expected satisficing regret is bounded by
\begin{equation}
\mathbb{E}[\texttt{Regret}_S]\leq \dfrac{K}{\Delta_S}\cdot \text{polylog}\left(\dfrac{K}{\Delta_S\Delta_S^*}\right)
\label{eq:total_regret_bbw_instance}
\end{equation}

Following similar steps in the proof of Theorem~\ref{thm:regret_bound_1}, we can prove that the expected satisficing regret is upper bounded by
\begin{equation}
\mathbb{E}[\texttt{Regret}_S] \leq \min\left\{\dfrac{K}{\Delta_S^*}\cdot \text{polylog}\left(\dfrac{K}{\Delta_S^*}\right),\sqrt{KT}\cdot \text{polylog}(T)\right\}.
\label{eq:total_regret_bbw_Delta*}
\end{equation}
Combining the regret bounds in \eqref{eq:total_regret_bbw_instance}, \eqref{eq:total_regret_bbw_Delta*}, we have completed the proof of Theorem~\ref{thm:regret_bound_bbw}.

\subsection{Proof of Proposition~\ref{prop:regret_bound_round_bbw}}\label{sec:prop:regret_bound_round_bbw}
Recall that $t_i=\gamma_i^{-2}$, then $\log(t_i)=\log(\gamma_i^{-2})\leq 2\log(1/\gamma_i)$ and $t_i^{1/2}=2^{i}=1/\gamma_i$. Therefore in the realizable case, by the worst-case expected standard regret bound of \texttt{ALG}, the satisficing regret incurred when running $\texttt{ALG}(t_i)$ can be bounded by
\begin{equation}
\begin{aligned}
&\E\left[\sum_{s=1}^{t_i}\max\{S-r(A_s),0\}\right]\\
\leq&\E\left[t_ir(A^*)-\sum_{s=1}^{t_i}r(A_s)\right]\leq c_1\sqrt{K}t_i^{1/2}\log(t_i)^{1/2}\\
=&\sqrt{2}c_1\sqrt{K}\gamma_i^{-1}\log(1/\gamma_i)^{1/2},
\label{eq:BBW_prop_oracle_time_bound}
\end{aligned}
\end{equation}
and the instance-dependent bound
\begin{equation}
\begin{aligned}
&\E\left[\sum_{s=1}^{t_i}\max\{S-r(A_s),0\}\right]\\
=&\E\left[\sum_{s=1}^{t_i}(S-r(A_s))\mathbf{1}(r(A_s)<S)\right]\\
\overset{\text{(i)}}{\leq}&\E\left[\sum_{a\in\mathcal{A},\,r(a)<S}N_a(t)(r(A^*)-r(a))\right]\\
\overset{\text{(ii)}}{\leq}&\sum_{a\in\mathcal{A},\,r(a)<S}\frac{c_2\log(t_i)}{\Delta_a} \overset{\text{(iii)}}{\leq} \sum_{a\in\mathcal{A},\,r(a)<S}\frac{c_2\log(t_i)}{\Delta_S^a}\\
=& \sum_{a\in\mathcal{A},\,r(a)<S}\frac{2c_2\log(1/\gamma_i)}{\Delta_S^a},
\label{eq:BBW_prop_oracle_instance-depend_bound}
\end{aligned}
\end{equation}
where step (i) is by decomposing satisficing regret by arm pulls, step (ii) follows from the instance-dependent bound on expected arm pulls for \texttt{ALG}, and inequality (iii) is because $\Delta_a=r(A^*)-r(a)>S-r(a)=\Delta_S^a$. Combining \eqref{eq:BBW_prop_oracle_time_bound} and \eqref{eq:BBW_prop_oracle_instance-depend_bound}, we conclude that the satisficing regret incurred when running $\texttt{ALG}(t_i)$ is bounded by the
\begin{equation}
\E\left[\sum_{s=1}^{t_i}\max\{S-r(A_s),0\}\right] \leq \min\left\{\sqrt{2}c_1\sqrt{K}\gamma_i^{-1}\log(1/\gamma_i)^{1/2},\, \sum_{a\in\mathcal{A},\,r(a)<S}\frac{2c_2\log(1/\gamma_i)}{\Delta_S^a}\right\}
\label{eq:Step1_Regret_bbw}
\end{equation}

Then, we consider the expected satisficing regret incurred in the forced sampling with UCB test. Using similar arguments as in~\eqref{eq:Step2_Regret_real}, the expected satisficing regret incurred in Step 2 is upper bounded by
\begin{equation}
T_i\cdot\dfrac{1}{t_i}\cdot  c_1\sqrt{Kt_i\log(t_i)}
=\sqrt{2}c_1\sqrt{K}\gamma_i^{-1}\log(1/\gamma_i)^{1/2}.
\label{eq:BBW_prop_UCB_time_bound}
\end{equation}
Then, we establish an expected satisficing regret bound in Step 2 with respect to $\Delta_S$. If $r(\hat{A}_i)<S$, for simplicity, we denote $\hat{A}_i$ as $a$, and define $\tau_0$ as
$$
\tau_0=\max\left\{\tau:\,\sqrt{4\log(\tau+1)/\tau}>\Delta_S^a/2  \right\}.
$$
We first provide an upper bound on $x$ given $x/\log(x+1)<y$. Because $\sqrt{x}>\log(x+1)$ for any $x\geq1$, we know $y>x/\log(x+1)>\sqrt{x}$. Then, for any $x\geq1$
$$
x<y\log(x+1)<y\log(2x)<y\log(2y^2).
$$
Plugging into the definition of $\tau_0$ as $\tau/\log(\tau+1)<16/{\Delta_S^a}^2$, we get
\begin{equation*}
\tau_0<\frac{16}{{\Delta_S^a}^2}\log\left(\frac{2\cdot16^2}{{\Delta_S^a}^4} \right) < \frac{64}{{\Delta_S^a}^2}\log\left(\frac{5}{\Delta_S^a} \right).
\end{equation*}
For $\tau\leq\tau_0$, the satisficing regret is bounded by
\begin{equation}
\tau_0\cdot\Delta_S^a \leq \frac{64}{\Delta_S^a}\log\left(\frac{5}{\Delta_S^a} \right).
\label{eq:BBW_prop_UCB_instance-depend_bound_before}
\end{equation}
For $\tau>\tau_0$, we know the confidence radius is less than $\Delta_S^a/2$, and the non-satisficing arm is pulled if its UCB is greater than $S$ as
$$
\dfrac{\hat{r}_i^{\text{tot}}}{\tau}+\sqrt{\dfrac{4\log(\tau)}{\tau}}\geq S=r(\hat{A}_i)+\Delta_S^a > r(\hat{A}_i)+2\sqrt{\dfrac{4\log(\tau)}{\tau}},
$$
this happens with probability
\begin{equation*}
\Pr\left(\dfrac{\hat{r}^{\text{tot}}_i}{\tau}+\sqrt{\dfrac{4\log(\tau+1)}{\tau}}>S\right) \leq \exp\left(-8\log(\tau+1) \right) =\frac{1}{(\tau+1)^8}.
\end{equation*}
Therefore the expected number of arm pulls of the non-satisficing $\hat{A}_i$ after first pulling it $\tau_0$ times is bounded by
\begin{equation}
\begin{aligned}
&\E\left[\text{Number of times }\hat{A}_i\text{ is pulled in forced sampling after the first }\tau_0 \text{ times}\right]\\
=&\sum_{s=\tau_0+1}^{\infty}\Pr(\tau\geq s)
= \sum_{s=\tau_0+1}^\infty \dfrac{1}{(s+1)^{8}}\leq 1.    
\label{eq:BBW_prop_UCB_instance-depend_bound_after}
\end{aligned}
\end{equation}
Combining the instance-dependent regret bound before $\tau_0$ \eqref{eq:BBW_prop_UCB_instance-depend_bound_before} and after $\tau_0$ \eqref{eq:BBW_prop_UCB_instance-depend_bound_after}, recall that $\Delta_S^a<1<1/\Delta_S^a$, we get
\begin{equation}
1\cdot\Delta_S^a + \frac{64}{\Delta_S^a}\log\left(\frac{5}{\Delta_S^a} \right) \leq \frac{1}{\Delta_S^a} + \frac{64}{\Delta_S^a}\log\left(\frac{5}{\Delta_S^a} \right) \leq \frac{65}{\Delta_S^a}\log\left(\frac{5}{\Delta_S^a} \right) \leq \frac{65}{\Delta_S}\log\left(\frac{5}{\Delta_S} \right).
\label{eq:BBW_prop_UCB_instance-depend_bound}
\end{equation}
Therefore, combining the instance-independent regret \eqref{eq:BBW_prop_UCB_time_bound} and the instance-dependent regret \eqref{eq:BBW_prop_UCB_instance-depend_bound} in forced sampling, the satisficing regret incurred in forced sampling with UCB test is bounded by
\begin{equation}
\min\left\{\sqrt{2}c_1\sqrt{K}\gamma_i^{-1}\log\left(\dfrac{1}{\gamma_i}\right)^{1/2} ,\; \frac{65}{\Delta_S}\log\left(\frac{5}{\Delta_S} \right)\right\}.
\label{eq:Step2_Regret_bbw}
\end{equation}

If $r(\hat{A}_i)<S$, then for every $k$, round $i$ is not terminated after pulling arm $\hat{A}_i$ for $k$ times (in addition to the number of pulls $T_i$ during $\texttt{ALG}(t_i)$) occurs with probability at most
\begin{align*}
\nonumber&\Pr\left(\dfrac{\hat{r}^{\text{tot}}_i}{T_i+k}-\sqrt{\dfrac{4\log(T_i+k)}{T_i+k}}>S\right)\\
\nonumber\leq &\Pr\left(\dfrac{\hat{r}^{\text{tot}}_i}{T_i+k}-\sqrt{\dfrac{4\log(T_i+k)}{T_i+k}}>r(\hat{A}_i)\right)\\
\leq& \exp\left(-2\cdot4\log(T_i+k)\right) \\
=& \dfrac{1}{(T_i+k)^{8}}\leq \dfrac{1}{(k+2)^{8}},
\end{align*}
where the second inequality is based on Hoeffding's inequality, and the third inequality based on $T_i\geq2$. Therefore expected number of arm pulls of $\hat{A}_i$ after first pulling it $T_i$ times during $\texttt{ALG}(t_i)$ is bounded by
\begin{align*}
\nonumber&\E\left[\text{Number of times }\hat{A}_i\text{ is pulled in round }i\text{ after the first }T_i \text{ times}\right]\\
=&\sum_{s=1}^{\infty}\Pr(k\geq s)
=\sum_{s=1}^\infty \dfrac{1}{(s+2)^{8}}\leq 1.
\end{align*}
So, the satisficing regret in LCB test can be bounded by $1$. 

Summing up the two parts of satisficing regret, \ie, the satisficing regret from $\texttt{ALG}(t_i)$ \eqref{eq:Step1_Regret_bbw}, forced sampling with UCB test \eqref{eq:Step2_Regret_bbw}, and LCB test, we have that the satisficing regret of round $i$ is bounded by
\begin{equation*}
\begin{aligned}
&\min\left\{2\sqrt{2}c_1\sqrt{K}\gamma_i^{-1}\log\left(\dfrac{1}{\gamma_i}\right)^{1/2}+1,\; \frac{65}{\Delta_S}\log\left(\frac{5}{\Delta_S} \right) + \sum_{a\in\mathcal{A},\,r(a)<S}\frac{2c_2\log(1/\gamma_i)}{\Delta_S^a}+1 \right\}\\
\leq & \min\left\{4c_1\sqrt{K}\gamma_i^{-1}\log\left(\dfrac{1}{\gamma_i}\right)^{1/2} ,\; \frac{66}{\Delta_S}\log\left(\frac{5}{\Delta_S} \right) + \frac{K}{\Delta_S}\cdot 2c_2\log\left(\dfrac{1}{\gamma_i}\right)  \right\}.
\end{aligned}
\end{equation*}

\subsection{Proof of Proposition~\ref{prop:exit_prob_bbw}}\label{sec:prop:exit_prob_bbw}
Using similar arguments as in~\eqref{eq:single_pull_regret_std} that
\begin{equation*}
\E[r(A^*)-r(\hat{A}_i)]\leq\dfrac{1}{t_i}\cdot c_1\sqrt{Kt_i\log(t_i)}=\sqrt{2}c_1\sqrt{K}\gamma_i\log(1/\gamma_i)^{1/2}.
\end{equation*}

According to the condition of this proposition, we have $i$ satisfies the condition 
\begin{equation*}
    16\sqrt{2}c_1\sqrt{K}\gamma_i\log(1/\gamma_i)^{1/2}\leq \Delta_S^*.
\end{equation*} 
By the Markov's inequality, we have
\begin{equation}
\Pr(r(A^*)-r(\hat{A}_i)\geq \Delta_S^*/2)\leq \Pr\left(r(A^*)-r(\hat{A}_i)\geq 8\sqrt{2}c_1\sqrt{K}\gamma_i\log(1/\gamma_i)^{1/2}\right)\leq\dfrac{1}{8}.
\label{eq:prob_bound_1_bbw}
\end{equation}

If $r(\hat{A}_i)\geq r(A^*)-\Delta_S^*/2$, which happens with probability at least 7/8 according to the above inequality, then $r(\hat{A}_i)\geq S+\Delta_S^*/2$ holds. We first analyze the exit probability in LCB test. If $i$ satisfies the condition stated in the proposition, we can recall Inequality~\eqref{eq:radius<Delta/4} that 
\begin{equation*}
\sqrt{\dfrac{{4\log(T_i+k)}}{T_i+k}} \leq \dfrac{\Delta_S^*}{4},
\end{equation*}
Suppose round $i$ is terminated for some $k$, then
\begin{equation*}
\dfrac{\hat{r}_i^{\text{tot}}}{T_i+k}-\sqrt{\dfrac{4\log(T_i+k)}{T_i+k}}<S=r(A^*)-\Delta_S^*\leq r(\hat{A}_i)-\dfrac{\Delta_S^*}{2}\leq r(\hat{A}_i)-2\sqrt{\dfrac{4\log(T_i+k)}{T_i+k}},
\end{equation*}
or equivalently
\begin{equation*}
\dfrac{\hat{r}_i^{\text{tot}}}{T_i+k}\leq r(\hat{A}_i)-\sqrt{\dfrac{4\log(T_i+k)}{T_i+k}}.
\end{equation*}
By the Hoeffding's inequality we have
\begin{equation*}
\Pr\left(\dfrac{\hat{r}_i^{\text{tot}}}{T_i+k}\leq r(\hat{A}_i)-\sqrt{\dfrac{4\log(T_i+k)}{T_i+k}}\right) \leq \dfrac{1}{(T_i+k)^8}\leq \dfrac{1}{(k+2)^8}.
\end{equation*}
Therefore conditioned on $r(\hat{A}_i)\geq r(A^*)-\Delta_S^*/2$, the probability that round $i$ is terminated during LCB test by the end of the horizon is bounded by
\begin{equation}
\sum_{k=0}^\infty \dfrac{1}{(k+2)^8} \leq \dfrac{1}{16}.
\label{eq:prob_bound_3_bbw}
\end{equation}

Then, we consider the probability of termination during forced sampling with UCB test given $r(\hat{A}_i)\geq S+\Delta_S^*/2$. For any $\tau\in[T_i]$, we start a new round if 
$$
\dfrac{\hat{r}_i^{\text{tot}}}{\tau} + \sqrt{\dfrac{4\log(\tau+1)}{\tau}} <S < r(\hat{A}_i) - \Delta_S^* < r(\hat{A}_i).
$$
By the Hoeffding's inequality we have
\begin{equation*}
\Pr\left(\dfrac{\hat{r}_i^{\text{tot}}}{\tau} + \sqrt{\dfrac{4\log(\tau+1)}{\tau}} < r(\hat{A}_i) \right) \leq \frac{1}{(\tau+1)^8}.
\end{equation*}
Therefore conditioned on $r(\hat{A}_i)\geq r(A^*)-\Delta_S^*/2$, the probability that round $i$ is terminated during forced sampling with UCB test by the end of the horizon is bounded by
\begin{equation}
\sum_{\tau=1}^{T_i} \dfrac{1}{(\tau+1)^8} \leq \sum_{\tau=1}^{\infty} \dfrac{1}{(\tau+1)^8} \leq \dfrac{1}{16}.
\label{eq:prob_bound_2_bbw}
\end{equation}

Combining \eqref{eq:prob_bound_1_bbw}, \eqref{eq:prob_bound_3_bbw}, and \eqref{eq:prob_bound_2_bbw}, we conclude that the probability that round $i$ terminates by the end of the time horizon is upper bounded by $1/4$.

\subsection{Proof of Theorem~\ref{thm:regret_nonsat_bbw}}\label{sec:thm:regret_nonsat_bbw}
Because the regret incurred during forced sampling with UCB test is bounded by the one incurred without UCB test, we know the standard regret incurred in round $i$ is still at most
\begin{equation*}
\dfrac{8c_1\sqrt{K}}{(1-\alpha)^\beta}\gamma_i^{-\alpha/(1-\alpha)}\log({4}/\gamma_i)^{\beta}.
\end{equation*}

Since the length of the time horizon is $T$ time steps, we have $t_i\leq T$, thus the maximum number of rounds is bounded by $i_{\max}=\lceil\alpha\log_2(T)\rceil$. Therefore, the same as the regret bound in Theorem~\ref{thm:regret_nonsat} (following the same chain of inequality in \eqref{eq:main_bound_sublinear}), the regret is bounded by
\begin{align*}
\mathbb{E}[\texttt{Regret}]\leq&\sum_{i=1}^{i_{\max}}\dfrac{8c_1\sqrt{K}}{(1-\alpha)^\beta}\gamma_i^{-\alpha/(1-\alpha)}\log(8T)^\beta\\
\leq &\sum_{i=1}^{i_{\max}} \dfrac{8c_1\sqrt{K}}{(1-\alpha)^\beta}2^{i+1}\log(8T)^\beta \\
\leq & \dfrac{32c_1\sqrt{K}}{(1-\alpha)^\beta}2^{i_{\max}}\log(8T)^\beta \\
\leq&   \dfrac{32c_1\sqrt{K}}{(1-\alpha)^\beta}2^{\alpha\log_2(T)+1}\log(8T)^\beta\\
\leq& \dfrac{64c_1\sqrt{K}}{(1-\alpha)^\beta}T^\alpha\log(8T)^\beta.
\end{align*}
Plugging into $\alpha=1/2$ and $c_1\sqrt{K}\propto\sqrt{K}$ in the UCB algorithm, the standard regret is bounded by $\sqrt{KT}\cdot\text{polylog}(T)$.

\section{Dependence in Dimension for Lipschitz Bandits}
In Corollary~\ref{cor:lipschitz_bandit}, the expected constant satisficing regret of \satex~in the realizable case is $\tilde{O}\left({\Delta_S^*}^{-(d+1)}\right)$, which grows exponentially in the dimension $d$. In this section, we test this exponential dependence on $d$. We consider a class of instances of Lipschitz bandits with concave expected reward in domain $(x_1,\cdots,x_d)\in[0,1]^d$. To fix the Lipschitz coefficient, the rewards are set as $r(x_1,\cdots,x_d)=1-25/(7\sqrt{d})\cdot\sum_{i=1}^d(x_i-0.7)^2$ so that $L=5$ for all $d\geq1$. We vary the dimension from 1 to 7 with a stepsize of 1. The time horizon is set to be 500000 and the satisficing level is set to be $S=0.5$. We use the Uniform UCB as the learning oracle. The experiment is repeated for 100 times and we report the average results.

\begin{figure}[!ht]
    \centering
    \includegraphics[height=4.5cm]{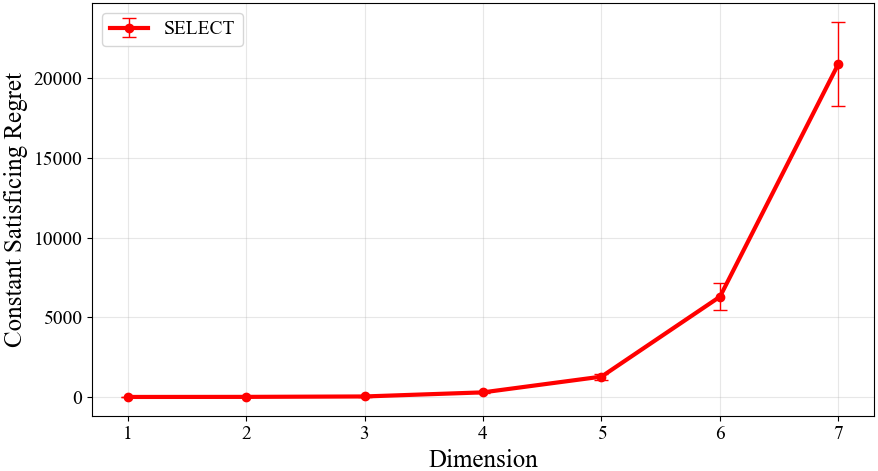}
    \caption{Constant satisficing regret changing with dimension $d$ in Lipschitz bandits}
    \label{fig:lip_dim}
\end{figure}

The result is presented in Figure~\ref{fig:lip_dim}, where one can see that the constant satisficing regret follows an exponential increase as the dimension $d$ gets large, which matches our theoretical result.

\section{Auxiliary Results}\label{sec:aux_results}

For completeness, we restate some known results from existing literature.

Consider the following $K$ instances of $K$-arm bandits. The reward of each arm follows a Bernoulli distribution, and under instance $k$, the mean rewards are defined as
\begin{equation}\label{eq:lb_instances}
r^{(k)}_k=\dfrac{1}{2}+\Delta,\ r^{(k)}_{k'}=\dfrac{1}{2},\ \forall k'\neq k.
\end{equation}

The following lemma establishes the information impossibility of distinguishing the instances defined in \eqref{eq:lb_instances} within $O(K/\Delta^2)$ time steps.

\begin{lemma}[Lemma 2.8 of \cite{Slivkins2019}]\label{lemma:aux}
Let $\pi$ be any non-anticipatory policy that commits one arm $i_T$ after adaptively making $T$ arm pulls. Then there exists a constant $0<c<1$ such that for all $T\leq cK/\Delta^2$, there exists at least $\lceil K/3\rceil$ arms $k\in[K]$ such that $\Pr^{(k)}(i_T=k)<3/4$.
\end{lemma}

}

\end{appendices}

\end{document}